\documentclass[a4paper,fleqn]{cas-dc}

\usepackage{natbib}
\AtBeginDocument{\DeclareRobustCommand{\cite}{\citep}}
\usepackage{grffile}
\usepackage{enumitem}
\usepackage{booktabs}
\usepackage{amsmath}
\usepackage{upgreek}
\usepackage{graphicx}
\usepackage{amsmath, amssymb}
\usepackage{booktabs} 
\usepackage{tabularx}
\usepackage[capitalize, nameinlink]{cleveref}
\usepackage{hyperref} 
\usepackage{cleveref} 
\usepackage{float} 
\usepackage[linesnumbered,ruled,vlined,algo2e]{algorithm2e} 
\usepackage{subcaption} 
\usepackage{makecell}
\usepackage{array}
\usepackage{float}

\def\tsc#1{\csdef{#1}{\textsc{\lowercase{#1}}\xspace}}
\tsc{WGM}
\tsc{QE}

\begin{document}
\let\WriteBookmarks\relax
\def\floatpagepagefraction{1}
\def\textpagefraction{.001}

\shorttitle{Random Walk on Bézier Curves for Global Optimization}    

\shortauthors{Jinpeng Wang et al.}  

\title [mode = title]{Random Walk on Bézier Curves for Global Optimization}  



%

\author[1]{Jinpeng Wang}
\fnmark[1]
\ead{2211060117@stu.lntu.edu.cn}
\credit{Conceptualization, Methodology, Software, Formal analysis, Writing - Original Draft}

\author[2]{Xingguo Xu}
\ead{xuxingguo@mail.dlut.edu.cn}
\fnmark[1]
\credit{Writing – review \& editing, Validation, Investigation, Visualization}

\author[3]{Yujing Sun}
\ead{2211060216@stu.lntu.edu.cn}
\credit{Validation, Software}

\author[4]{Jiguang Yu}
\ead{jyu678@bu.edu}
\credit{Writing – review \& editing}

\author[5]{Kaichen Ouyang}
\ead{oykc@mail.ustc.edu.cn}
\credit{Writing – review \& editing}

\author[6]{Yuansheng Gao}[orcid=0000-0003-3278-8835]
\cormark[1]
\ead{y.gao@zju.edu.cn}
\credit{Methodology, Supervision}

\affiliation[1]{organization={School of Computer Science, Northwestern Polytechnical University},
            city={Xi’an},
            postcode={710129}, 
            country={China}}
\affiliation[2]{organization={School of Mathematical Sciences, Dalian University of Technology},
            city={Dalian},
            postcode={116024}, 
            country={China}}
\affiliation[3]{organization={School of Mathematics, Hefei University of Technology},
            city={Hefei},
            postcode={230601}, 
            country={China}}
\affiliation[4]{organization={College of Engineering, Boston University},
            city={Boston},
            postcode={02215}, 
            country={USA}}
\affiliation[5]{organization={Department of Physics, University of Science and Technology of China},
            city={Hefei},
            postcode={230026}, 
            country={China}}
\affiliation[6]{organization={College of Computer Science and Technology, Zhejiang University},
            city={Hangzhou},
            postcode={310027}, 
            country={China}}

\fntext[1]{Equal contribution.}
\cortext[1]{Corresponding author.}
\begin{abstract}
Balancing exploration and exploitation remains a central challenge in metaheuristic optimization. To address this issue, this paper proposes Bézier Walk Evolution (BWE), a geometry-driven optimization framework that reformulates evolutionary search as adaptive trajectory construction in the decision space. BWE integrates Bézier curve modeling with a distance-aware random walk mechanism to generate topology-guided search trajectories. By adaptively varying the curve order during evolution, the proposed method enables a smooth transition from diversified global exploration to refined local exploitation. Higher-order Bézier curves leverage multiple population-derived control points to enhance search diversity, while lower-order curves generate near-linear trajectories to improve convergence efficiency. This adaptive geometric search mechanism provides an interpretable alternative to conventional nature-inspired designs. Extensive experiments on 41 benchmark functions from the CEC2017 and CEC2022 suites, spanning dimensions from 10 to 100, show that BWE achieves strong overall performance and favorable scalability compared with 7 classical and 6 state-of-the-art optimizers, including L-SHADE and CMA-ES. Additional evaluations on five constrained engineering design problems further demonstrate the practical applicability and robustness of BWE.
\end{abstract}




\maketitle

\section{Introduction}\label{Introduction}
With the continuous advancement of digitalization, optimization problems in fields such as intelligent manufacturing, smart cities, and quantum computing have grown increasingly complex \citep{singh2022towards}.
Traditional mathematical programming methods rely heavily on gradient information and convexity assumptions, making them less effective for high-dimensional non-convex problems \citep{yang2010engineering}.
Meanwhile, heuristic methods often generalize poorly due to their dependence on problem-specific structures \citep{wong2019review}.
Consequently, metaheuristic algorithms have emerged as effective tools because of their gradient-free nature, robustness, and adaptability to black-box models \citep{li2024kernel}.

Although existing metaheuristics have achieved significant success, two fundamental challenges remain.
First, balancing exploration and exploitation throughout the search process remains critical \citep{alba2005exploration}. Excessive exploration weakens solution refinement, whereas excessive exploitation increases the risk of premature convergence to suboptimal regions.
Second, many current algorithms still rely heavily on homogenized nature-inspired metaphors, limiting mechanism-level innovation and increasing the risk of stagnation in local optima.
Breaking beyond traditional nature-inspired paradigms has therefore become increasingly important for improving algorithmic performance \citep{wang2025logistic}.

According to the No Free Lunch theorem, no algorithm can outperform all others across every optimization problem \citep{wolpert2002no}. Therefore, despite the large number of existing methods, developing novel optimization strategies remains essential for increasingly complex and diverse real-world applications.

Trajectory planning is a cornerstone of computational geometry and has also attracted increasing attention in evolutionary computation, though the integration of these two fields remains limited.
Among geometric tools, Bézier curves are widely used in robotic motion planning \citep{simba2016real} and CAD modeling \citep{piegl2012nurbs} due to their smoothness, affine invariance, and convex hull property.
Their shape is fully determined by control points, whose arrangement governs trajectory curvature.

This geometric controllability naturally aligns with metaheuristic search dynamics.
Evolution can be viewed as dynamic curve fitting: global exploration uses higher-order curves with dispersed control points to traverse complex landscapes, whereas local exploitation uses low-order, near-linear trajectories for rapid convergence.
However, this requires autonomously constructing the curve’s control points in an unknown search space.
Unlike CAD systems where they are deterministically assigned, evolutionary optimization must strategically sample control points from the population to reflect fitness.
\emph{Existing trajectory- or geometry-driven search operators often rely on coarse control-point selection (e.g., purely random picks or ad-hoc heuristics), which makes the induced trajectories unstable and weakly aligned with the population’s promising structures.}
\emph{More importantly, there is still a lack of an interpretable mechanism that explicitly links control-point construction to population topology, and further to the exploration–exploitation trade-off throughout evolution.}
Without a robust selection principle, Bézier-guided movement may degenerate into blind stochastic drift, failing to capture meaningful search directions embedded in population topology.

To bridge this gap, random walk theory provides a principled framework for topology-aware sampling in unknown spaces \citep{lovasz1993random}.
In metaheuristic optimization, random walks balance local refinement through short-range steps and global exploration through occasional long-range transitions.
More importantly, random walk dynamics can be biased by structural information (e.g., distance and population topology), transforming purely random motion into informed stochastic navigation.
By embedding a distance-aware random walk into control-point selection, candidate solutions can be sampled to reflect both the spatial distribution of the population and the attraction toward high-quality regions.
Consequently, the control points of Bézier trajectories can emerge endogenously from population dynamics rather than being arbitrarily assigned.

Inspired by the above analysis, we propose Bézier Walk Evolution (BWE).
BWE reformulates evolutionary search as a path construction process governed by Bézier curves of varying orders.
Higher-order curves enhance exploration through diversified trajectories, whereas lower-order curves promote exploitation via near-linear paths.
By adaptively adjusting curve order during evolution, BWE achieves a dynamic balance between exploration and exploitation without problem-specific heuristics.
The main contributions of this work are summarized as follows
\begin{itemize}[align=left, leftmargin=*]
\item We propose Bézier Walk Evolution (BWE), a novel evolutionary algorithm.

\item We introduce a distance-aware random walk mechanism for selecting Bézier control points, enabling search trajectories to reflect population topology and directional guidance toward the best solution.

\item We develop an adaptive strategy scheme that dynamically adjusts the Bézier curve order to balance global exploration and local exploitation.

\item Extensive experiments on CEC2017 and CEC2022 benchmarks demonstrate that BWE achieves competitive performance, surpassing several SOTA metaheuristics.

\item Validate the method’s practical applicability and robustness on five real-world optimization problems.
\end{itemize}


\section{Related works}\label{sec:RelatedWorks}
\subsection{Overview of metaheuristic algorithms}
Metaheuristic algorithms have been extensively developed for complex optimization problems and are commonly categorized into single-solution-based and population-based approaches according to the number of maintained solutions.

Simulated annealing is one of the most representative single-solution-based metaheuristic algorithms. 
\citet{kirkpatrick1983optimization} formally introduced the SA framework in 1983 by drawing inspiration from the physical annealing process and applying it to combinatorial optimization.
SA emulates the gradual cooling of materials to reach a state of minimal energy, and its key feature is the probabilistic acceptance of inferior solutions, which allows the algorithm to escape from local optima.
Non-monopolize search is a recently proposed single-solution-based local search optimization algorithm introduced by \citet{abualigah2024non}, aiming to prevent search stagnation by avoiding the monopolization of the search process by a single incumbent solution.
In addition, other notable single-solution-based approaches include the tabu search \citep{glover2013tabu}, and the large neighborhood search \citep{pisinger2018large}.

In addition, population-based metaheuristic algorithms represent a substantial and actively studied branch within the field.
These algorithms are further categorized according to their search mechanisms into four main types: the evolution-based algorithms (EAs), the swarm intelligence-based algorithms (SIAs), the physics-based algorithms (PAs), and the mathematics-based algorithms (MAs).

Inspired by Darwinian evolutionary theory, EAs typically employ fundamental genetic operations, such as selection, crossover, mutation, and survivor replacement. 
Among them, the genetic algorithm (GA) developed by \citet{holland1975adaptation} is one of the earliest and most influential representatives, using probabilistic rules to guide a population toward superior solutions.
Chaotic evolution optimization \citep{dong2025chaotic} incorporates deterministic chaotic maps into the search process to enhance population diversity and ergodicity, improving global exploration and reducing premature convergence.
Other notable evolution-based algorithms include the evolution strategy \citep{beyer2002evolution}, differential evolution \citep{das2010differential}, and the love evolution algorithm \citep{gao2024love}.

Another major category of SIAs draws inspiration from the social behaviors of animals, such as foraging and reproduction. 
\citet{kennedy1995particle} introduced particle swarm optimization, inspired by the cooperative behavior of bird flocks, enabling individuals to adjust their positions via intra-population information sharing.
The \emph{Philoponella prominens} optimizer \citep{gao2025escape} is a novel SIA inspired by the unique post-mating behaviors of \emph{P. prominens} to guide the search process toward high-quality solutions.
Other representative metaheuristics inspired by swarm intelligence include the ant colony optimization \citep{dorigo2007ant}, grey wolf optimizer \citep{mirjalili2014grey}, crayfish optimization algorithm \citep{jia2023crayfish}, and traffic jam optimizer \citep{wang2025traffic}.

PAs primarily guide the search process by simulating physical or chemical phenomena.
A representative method is the gravitational search algorithm proposed by \citet{rashedi2009gsa}, where candidate solutions interact through gravitational forces, directing the population toward higher-quality regions.
The Schrödinger optimizer \citep{hussein2025schrodinger} models the search process based on concepts derived from Schrödinger’s equation, characterizing candidate solutions through probabilistic wave functions to enable stochastic exploration while collapsing toward high-quality solutions.
Other representative PAs include the big bang-big crunch \citep{erol2006new}, black hole algorithm \citep{hatamlou2013black}, chemical-reaction-inspired optimization \citep{lam2009chemical}, atomic orbital search \citep{azizi2021atomic}, Archimedes optimization algorithm \citep{hashim2021archimedes}, and langevin equation \citep{chen2025lee}.

MAs are inspired by mathematical functions, equations, or deterministic update rules.
A representative example is the sine cosine algorithm proposed by \citet{mirjalili2016sca}, which updates candidate solutions by generating oscillatory movements around the current best solution using sine and cosine functions.
The Logistic-Gauss Circle optimizer \citep{wang2025logistic} integrates multiple nonlinear dynamical systems, using the Logistic and Gauss maps to drive exploration and the Circle map for exploitation, effectively balancing search diversity and stable convergence.
Other representative MAs include the Runge Kutta optimizer \citep{ahmadianfar2021run}, arithmetic optimization algorithm \citep{abualigah2021arithmetic}, PID-based search algorithm \citep{gao2023pid}, and intelligent cross-entropy optimizer \citep{farahmand2024intelligent}.

\subsection{Bézier curve in optimization}
Bézier curves have been adopted in optimization from two distinct perspectives. In the earlier, Bézier curves serve as the \textit{target representation} to be optimized rather than the vehicle of search. Representative examples include robot and UAV path planning, where evolutionary algorithms such as genetic algorithms~\citep{elhoseny2018bezier} and chaotic particle swarm optimization~\citep{tharwat2019intelligent} are employed to determine control points that define collision-free, smooth trajectories. In these methods, the Bézier curve defines the phenotype of a candidate solution, while the evolutionary algorithm operates in the parameter space of control points; the curve itself is not used to model how individuals move within the search space.

A conceptually adjacent and independently developed concurrent 
work is the Bézier curve-based optimization (BCO) 
by~\citet{zhao2026effective}, which also employs 1st-, 2nd-, and 3rd-order Bézier curves as search operators, assigning linear curves to exploitation and cubic curves to global exploration. However, BCO relies on a fixed balance factor to schedule strategy switching and constructs intermediate control points without considering the geometric relationships among the current individual, the sampled candidate pool, and the global elite. In contrast, BWE introduces a Bernstein-polynomial-inspired time-evolving probability model for soft strategy transition, and a topology-aware random walk mechanism in which control point selection probabilities are governed by a distance-based softmax mapping to generate trajectories with higher path tension. Furthermore, BWE incorporates an approximate tortuosity perturbation scaled by the tortuosity ratio $\rho_i$, which adaptively prevents individuals from being strictly confined to the smooth Bézier path. These distinctions make BWE a more geometrically principled and dynamically adaptive framework.


\section{Preliminary}\label{sec:Preliminary}
\subsection{Bézier curve}
The Bézier curve is a classical parametric curve based on Bernstein polynomials, originally introduced by \citet{bezier1972numerical}.
Endowed with smoothness, continuity, and strong geometric interpretability, it is particularly suitable for modeling continuous trajectories in optimization.
Given a set of control points $\mathbf{P}_0, \mathbf{P}_1, \ldots, \mathbf{P}_n$, where $\mathbf{P}_i \in \mathbb{R}^D$, an $n$th-order Bézier curve is defined as
\begin{equation}
\mathbf{B}(o) = \sum_{j=0}^{n} \binom{n}{j}(1-o)^{n-j}o^j \mathbf{P}_j, \quad o \in [0,1]
\end{equation}
where $o \in [0,1]$ is a scalar parameter determining the position along the curve, transitioning continuously from the initial point $\mathbf{P}_0$ to the terminal point $\mathbf{P}_n$.

The curve's shape is entirely determined by its control points.
Although intermediate control points generally do not lie on the curve, they govern its curvature and direction.
Consequently, the curve degree reflects the geometric degrees of freedom for path shaping: a first-order Bézier curve forms a straight-line segment, whereas higher-order curves generate increasingly flexible trajectories.

\begin{figure}
    \centering
    \includegraphics[width=0.825\columnwidth]{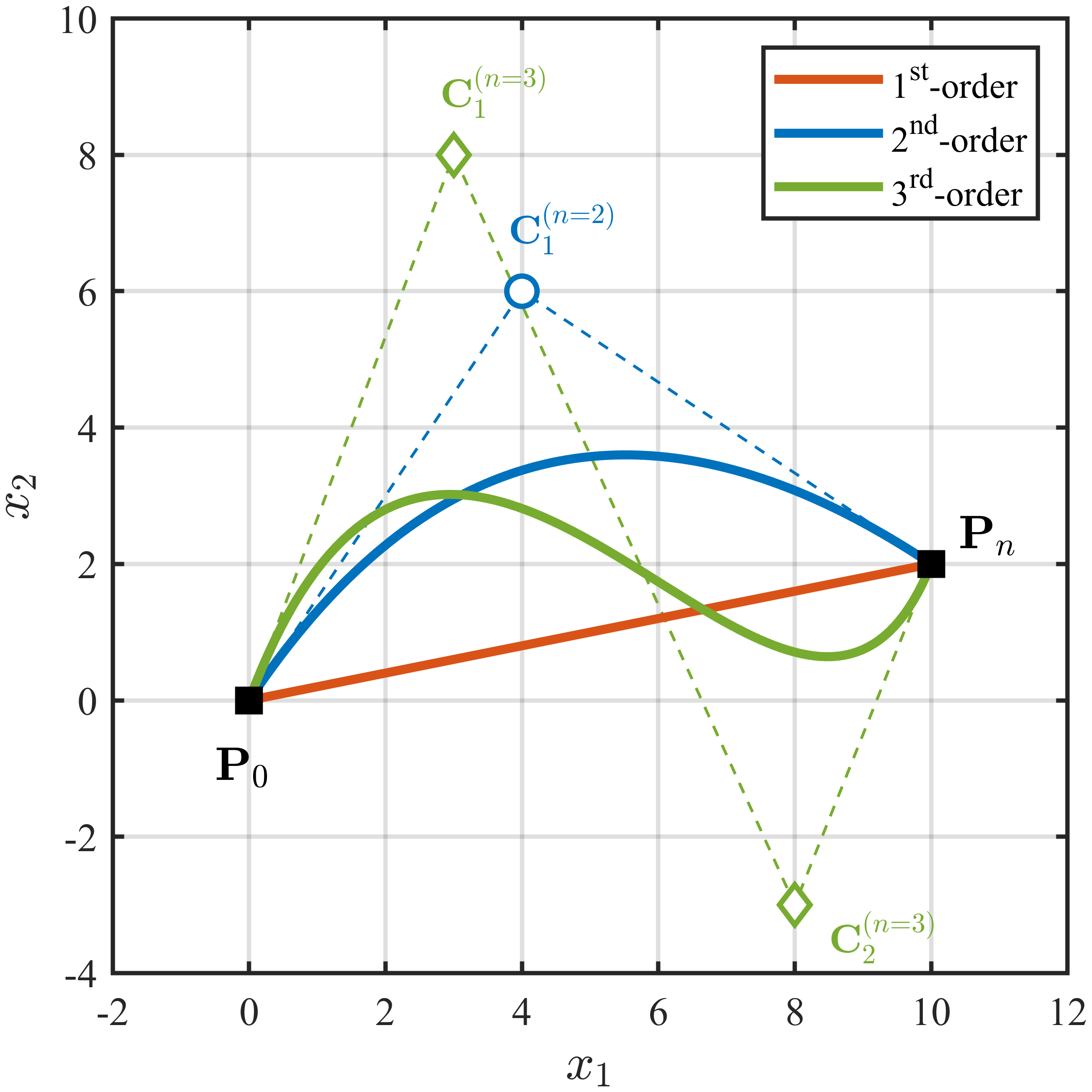}
    \caption{Examples of 1st, 2nd, and 3rd order Bézier curves.}
    \label{fig:Bezier_Trajectories}
\end{figure}

From an optimization perspective, Bézier curves provide a natural mechanism for constructing gradient-free search trajectories, where the initial and terminal control points represent the current and guiding solutions, while intermediate control points act as implicit waypoints that bend the trajectory toward promising regions.
This property enables adaptive control of the exploration–exploitation trade-off through control point arrangement and underpins the search-path design of BWE. 
To illustrate this behavior, \cref{fig:Bezier_Trajectories} compares trajectories generated by 1st-, 2nd-, and 3rd-order Bézier curves, highlighting the transition from linear exploitation to curvilinear exploration.

\subsection{Random walk}
A random walk is a fundamental stochastic process for modeling exploratory behaviors, extensively applied in evolutionary computation \citep{lovasz1993random}. In its classical form, it describes the successive movement of a particle whose next state is generated by a stochastic displacement from its current position. Formally, in a $D$-dimensional continuous space, it can be written as
\begin{equation}
\mathbf{x}^{(t+1)} \leftarrow \mathbf{x}^{(t)} + \boldsymbol{\Delta}^{(t)}
\label{x=x+delta}
\end{equation}
where $\boldsymbol{\Delta}^{(t)} \in \mathbb{R}^D$ is a random step vector. While purely random perturbations provide unbiased exploration, they often lead to inefficiency in complex landscapes. To improve sampling, structured random walks incorporate spatial relationships among candidate solutions into transition probabilities, such as the distance-aware formulation:
\begin{equation}
\mathbb{P}(\mathbf{x}^{(t)} \rightarrow \mathbf{y}) \propto \phi\big(d(\mathbf{x}^{(t)}, \mathbf{y})\big)
\end{equation}
where $d(\cdot,\cdot)$ is a distance metric and $\phi(\cdot)$ correlates transition probabilities with spatial topology.

In BWE, random walk is not used to update solution positions directly, but rather to stochastically generate intermediate guidance nodes for Bézier path construction. By viewing individuals as nodes in an implicit population graph, transition probabilities are designed to favor nodes that preserve geometric tension and avoid premature collapse into local regions. This process forms a structured stochastic walk, where transitions are guided by spatial relationships. Such a topology-guided walk embeds global population distribution into curve generation: stochastic transitions preserve geometric tension for exploration, while the destination-driven nature of Bézier curves ensures directional exploitation toward the elite. By integrating these mechanisms, BWE transforms discrete stochastic transitions into smooth evolutionary trajectories, unifying global exploration and local refinement.

\begin{figure}
    \centering
    \includegraphics[width=0.8\columnwidth]{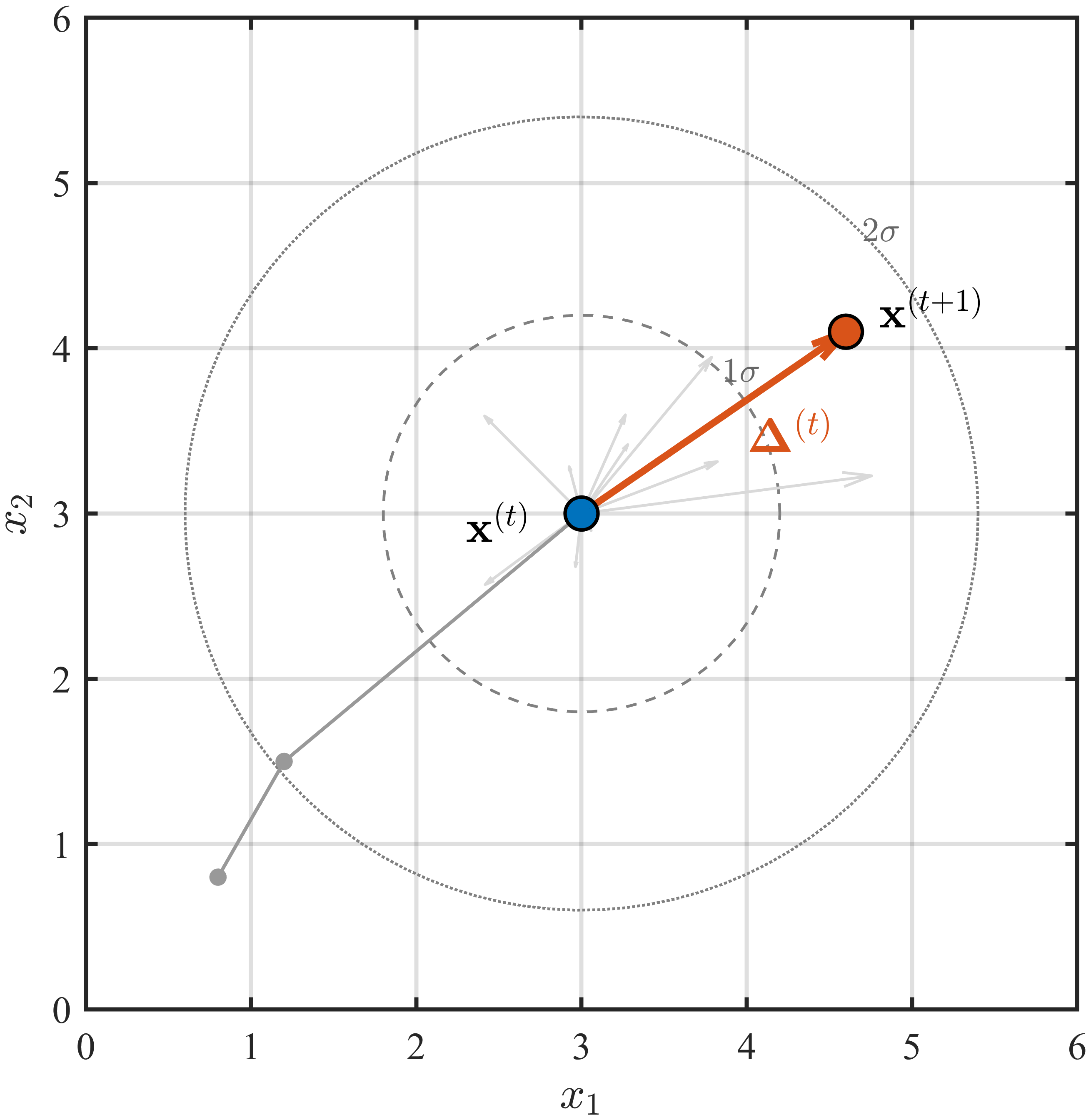}
    \caption{The stochastic displacement in random walk.}
    \label{fig:random_walk}
\end{figure}

The stochastic displacement mechanism is illustrated in \cref{fig:random_walk}. The faded gray line depicts the historical path, while concentric dashed circles represent the $1\sigma$ and $2\sigma$ confidence regions around $\mathbf{x}^{(t)}$. The displacement $\boldsymbol{\Delta}^{(t)}$ determines the transition to $\mathbf{x}^{(t+1)}$, forming the geometric basis for the distance-aware sampling strategy in BWE.

\section{Bézier walk evolution}
\label{sec:BezierWalkEvolutions}
\begin{figure*}
    \centering
    \includegraphics[trim=10pt 4pt 10pt 4pt, clip, width=0.6\linewidth]{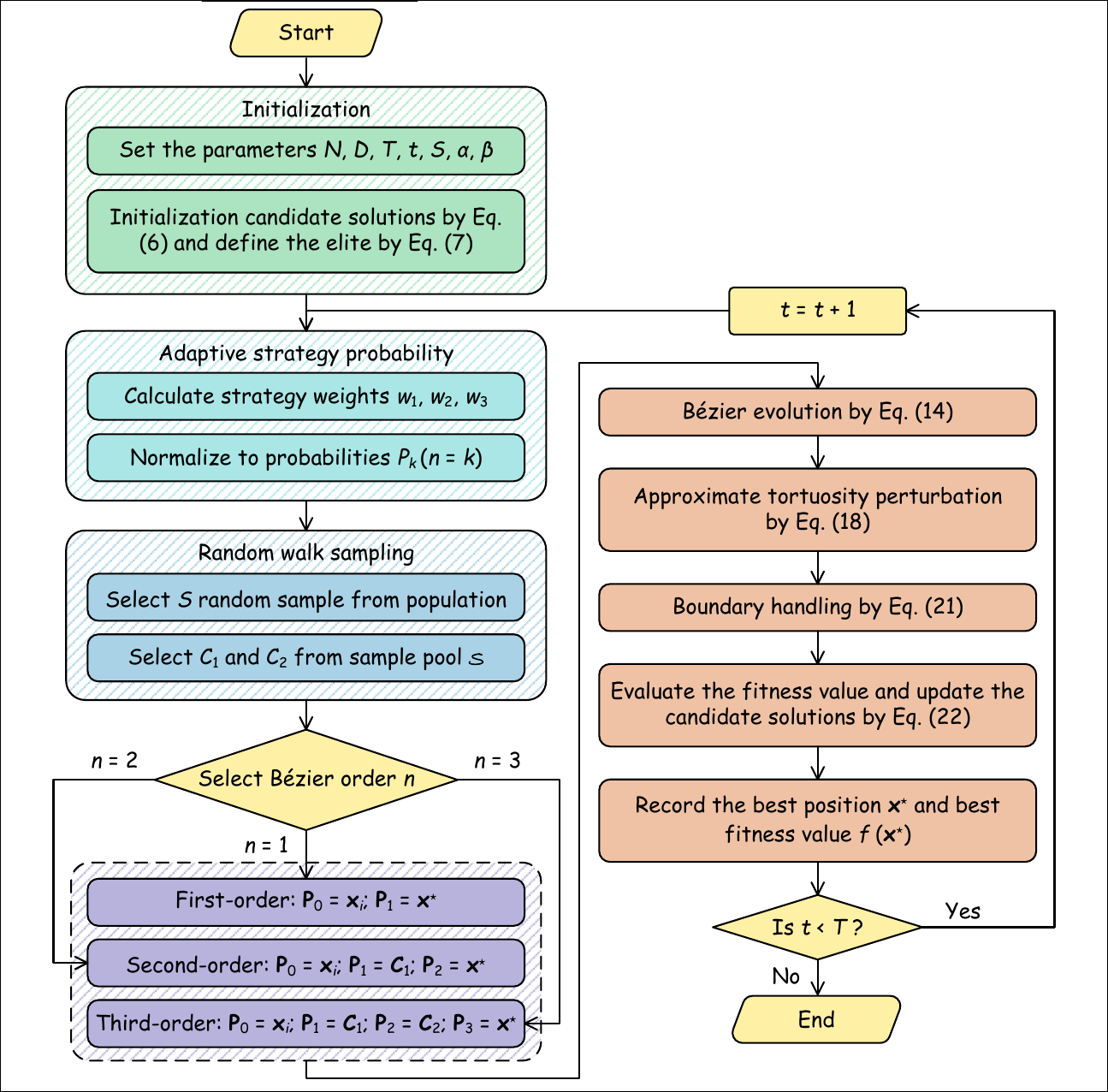}
    \caption{The flow chart of Bézier walk evolution.}
    \label{fig:FlowChart}
\end{figure*}

In this section, we present a rigorous mathematical formulation and detailed operational mechanism of BWE. The flowchart for BWE is shown in \cref{fig:FlowChart}. The source code of BWE is publicly available at 
\url{https://github.com/JinPengWang/bezier-walk-evolution}.

\subsection{Initialization}
To better highlight the evolutionary search mechanism of the algorithm under equivalent conditions, this paper adopts the most classical population initialization method. In the initial stage, each individual \(\mathbf{x}_{i}\in\mathbb{R}^{D},i=1,2,\cdots,N\) in the population is uniformly randomly distributed in the interval $[\boldsymbol{\ell},\mathbf{\textbf{u}}]$, defined as
\begin{equation}
\mathbf{x}_i=\boldsymbol{\ell}+(\boldsymbol{\mathbf{u}}-\boldsymbol{\ell})\odot\boldsymbol{\xi},\quad\boldsymbol{\xi}\sim U(0,1)^{1\times D}
\end{equation}
Here, $\odot$ denotes the element-wise product. At iteration $t$, the population is represented as $\mathcal{P}^{(t)}=\{\mathbf{x}_1^{(t)},\mathbf{x}_2^{(t)}, \ldots,\mathbf{x}_N^{(t)}\}$. We define the current global optimum (the elite) as
\begin{equation}
\mathbf{x}^\star=\arg\min_{\mathbf{x}_i^{(t)}\in\mathcal{P}^{(t)}}f(\mathbf{x}_i^{(t)})
\end{equation}

\subsection{Adaptive strategy probability}
The core concept of BWE is to leverage the trajectory generation capability of Bézier curves to plan individual evolutionary paths. To achieve both global exploration and precise local convergence, three search strategies based on 1st-, 2nd-, and 3rd-order Bézier curves are designed. Their selection is grounded in the intrinsic relationship between geometric degrees of freedom and search behavior, which is discussed further in the individual evolution phase. The search strategy is selected using a roulette wheel mechanism.

To smoothly switch among the three strategies, BWE employs a time-evolving probability model. Inspired by the Bernstein polynomial structure underlying Bézier curves, the selection weights $w_i$ define a nonlinear distribution that gradually shifts from exploration to exploitation. First, the iteration factor is defined as
\begin{equation}
    \vartheta=t/T
\end{equation}
where \(t\) denotes the current iteration number and \(T\) represents the maximum number of iterations. The selection probability weights \(w_i  (i=1,2,3)\) for three strategies are designed as
\begin{equation}\begin{cases}
w_3=(1-\vartheta) \\
w_2=3\vartheta(1-\vartheta) \\
w_1=\vartheta & 
\end{cases}\end{equation}

\begin{figure}
    \centering
    \includegraphics[width=0.8\columnwidth]{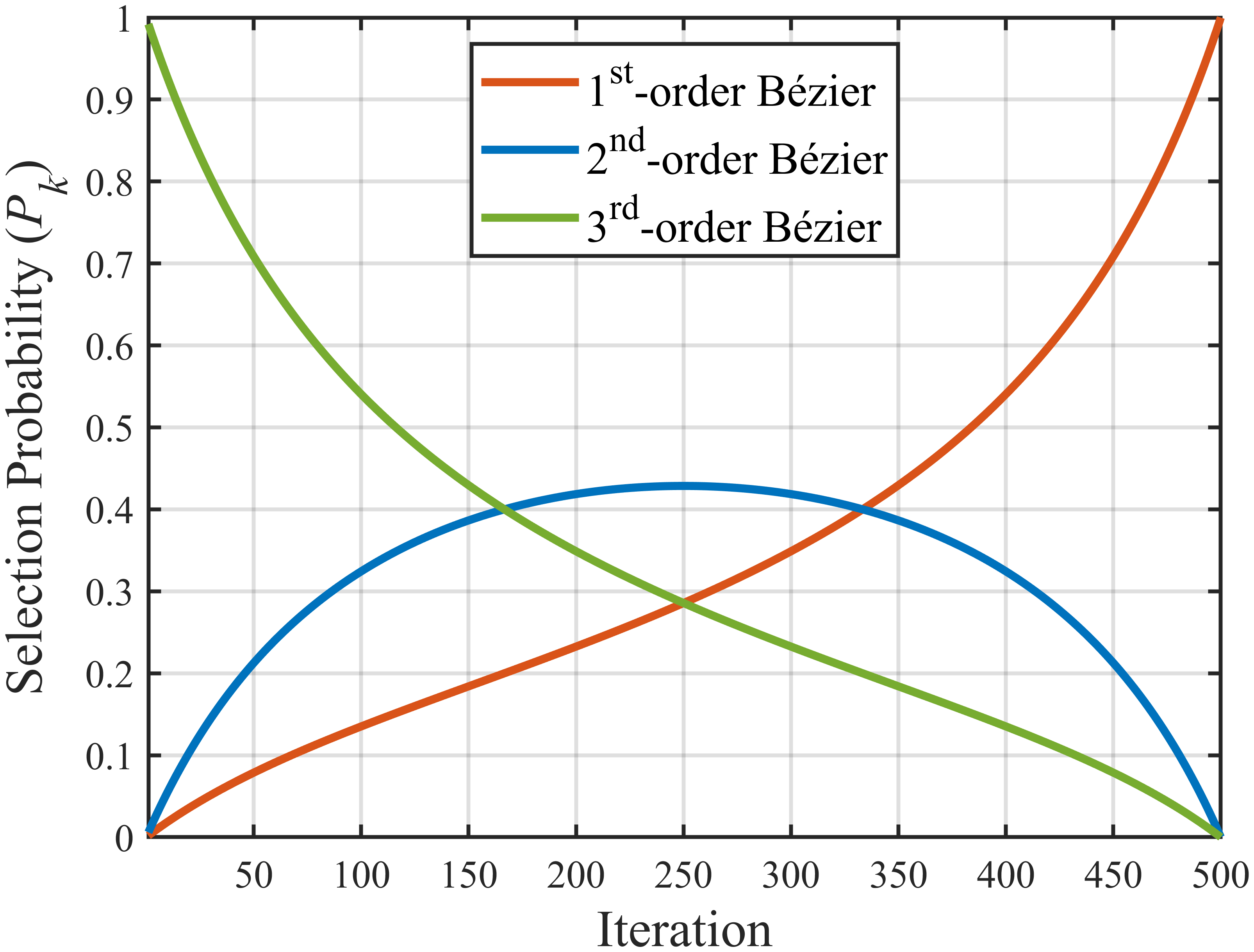}
    \caption{The selection probability for the orders.}
    \label{fig:probability}
\end{figure}

In the early phase, $w_3$ maintains a high profile to favor complex cubic paths for global exploration, while in the later phase, $w_1$ dominates to facilitate direct linear paths toward the optimal solution. The weight $w_2$ follows a unimodal distribution that increases initially and decreases later, peaking at the middle of the evolutionary process ($\vartheta = 0.5$). At this stage, $w_2$ possesses the highest selection probability among the three strategies, acting as a buffer that balances exploration and exploitation.
In each iteration, the probability $P_k$ that individual $\mathbf{x}_i$ executes the $k$-th search strategy is obtained through a linear normalization:
\begin{equation}P_k(n=k) = \frac{w_k}{\sum_{j=1}^3 w_j}\end{equation}
where $n$ denotes the order of the subsequent Bézier evolution. This probabilistic mechanism does not impose a rigid phase division but rather implements a soft switching strategy. Even in the later stages, the algorithm retains a small probability of executing higher-order searches, which endows the algorithm with an intrinsic capability to escape local optima when stagnation occurs. The selection probability curves for Bézier evolution orders are illustrated in \cref{fig:probability}.

\subsection{Random walk sampling}
To determine the control points of Bézier curves, BWE does not employ pure random selection but instead constructs a random walk mechanism based on topological distances between individuals.
In each iteration, we randomly sample a subset of $S$ individuals (where $S = \text{round}(\eta N)$) from the current population $\mathcal{P}^{(t)}$ to form the candidate pool $\mathcal{S} = \{\mathbf{s}_1, \mathbf{s}_2, \ldots, \mathbf{s}_S\}$. A randomly selected sample pool is used to characterize the local spatial structure of the current population. For the current individual \(\mathbf{x}_i\), we need to select guidance nodes \(\mathbf{C}_1,\mathbf{C}_2\in\mathcal{S}\). The selection probabilities are constructed based on the following three distance vectors:
\begin{itemize}[align=left, leftmargin=*]
    \item \(p_1\): The distance from \(\textbf{x}_i\) to each point in the sample pool.
    \item \(p_2\): The distance between points within the sample pool.
    \item \(p_3\): The distances from each point in the sample pool to the elite \(\mathbf{x}^\star\).
\end{itemize}

The distance between any two points is defined as the Euclidean norm \(d_{wv}=\|\mathbf{w}-\mathbf{v}\|_2\). To ensure numerical stability and balanced influence, raw distances are normalized to \(\tilde{d}\in[0,1]\) within the candidate set. These normalized distances are then mapped to a probability distribution via a softmax transformation. Specifically, given a reference point \(\mathbf{w}\) and a candidate set \(\mathcal{V}\) (e.g., the sampled pool \(\mathcal{S}\)), the probability of selecting \(\mathbf{v}\in\mathcal{V}\) is defined as
\begin{equation}
 p_{w\rightarrow v}=\frac{\exp(\tilde{d}_{wv})}{\sum\limits_{\mathbf{u}\in\mathcal{V}}\exp(\tilde{d}_{wu})}
\end{equation}
This mapping assigns higher probabilities to candidates that are farther away in the normalized distance sense. Since the endpoint of the Bézier curve is fixed at the global elite $\mathbf{x}^\star$ (ensuring ultimate convergence), selecting intermediate control points too close to $\mathbf{x}^\star$ or $\mathbf{x}_i^{(t)}$ would cause the trajectory to degenerate into a straight line, thereby reducing the search capability. 

By favoring samples that are distant from both the current individual and the elite within the sample pool, the algorithm generates curve paths with higher \textit{path tension} and wider spatial coverage. This repulsive probability mechanism effectively prevents the population from clustering too rapidly, ensuring that the path explores the potential basin of attraction before converging to the elite.

\subsubsection{First-order ({\normalfont$n=1$})}
When employing first-order Bézier evolution, the evolutionary path is controlled solely by the endpoints. This scenario can be regarded as a special case of the control point selection mechanism with minimal degrees of freedom, exhibiting higher exploitation performance. Therefore, all control points in the first-order Bézier evolution are defined as
\begin{itemize}
    \item[-] $\mathbf{P}_0=\mathbf{x}_i^{(t)}$; $\mathbf{P}_1=\mathbf{x}^\star$
\end{itemize}

\subsubsection{Second-order ({\normalfont$n=2$})}
For second-order Bézier evolution, only one intermediate control point $\mathbf{C}_1$ is required. This control point introduces limited curvature between the current solution and the elite solution, thereby achieving an effective balance between local exploration and convergence. The selection of $\mathbf{C}_1$ depends on the distance relationship between individual $\mathbf{x}_i^{(t)}$ and sample pool $\mathcal{S}$. The selection probability is defined as
\begin{equation}\mathbb{P}(\mathbf{C}_1)\propto p_1(\mathbf{x}_i^{(t)},\mathbf{C}_1)\cdot p_3(\mathbf{C}_1,\mathbf{x}^\star)\end{equation}
where $\quad\mathbf{C}_1\in\mathcal{S}$. This probability model simultaneously considers the local neighborhood structure of the current individual and the position of the control point relative to the global elite, making the generated trajectory both directional and maintaining certain randomness. Consequently, the control points for the second-order Bézier curve are defined as
\begin{itemize}
    \item[-] $\mathbf{P}_0=\mathbf{x}_i^{(t)}$; $\mathbf{P}_1=\mathbf{C}_1$; $\mathbf{P}_2=\mathbf{x}^\star$
\end{itemize}

\subsubsection{Third-order ({\normalfont$n=3$})}
When adopting the third-order Bézier evolution strategy, two intermediate control points $\mathbf{C}_1$ and $\mathbf{C}_2$ need to be selected to construct a curve path with higher degrees of freedom. The joint selection probability of the control points is defined as
\begin{equation}\mathbb{P}(\mathbf{C}_1,\mathbf{C}_2)\propto p_1(\mathbf{x}_i^{(t)},\mathbf{C}_1)\cdot p_2(\mathbf{C}_1,\mathbf{C}_2)\cdot p_3(\mathbf{C}_2,\mathbf{x}^\star) \end{equation}

Individuals at greater distances have higher probabilities of being selected as intermediate control points \(\mathbf{C}_1,\mathbf{C}_2\in\mathcal{S}\). This joint probability mechanism can be viewed as a stochastic walk path generation process constrained by the population topological structure. The control points for the third-order Bézier curve are defined as
\begin{itemize}
    \item[-] $\mathbf{P}_0=\mathbf{x}_i^{(t)}$; $\mathbf{P}_1=\mathbf{C}_1$; $\mathbf{P}_2=\mathbf{C}_2$; $\mathbf{P}_3=\mathbf{x}^\star$
\end{itemize}

\subsection{Bézier evolution}
For any individual $\mathbf{x}_i^{(t)}(\mathbf{P}_0)$, its update is given by the following Bézier path:
\begin{equation}\tilde{\mathbf{x}}_i\leftarrow\sum_{j=0}^n\binom{n}{j}(1-\tau)^{n-j}\tau^j\mathbf{P}_j\end{equation}

The curve parameter $\tau$ determines the step size and movement aggressiveness along the Bézier curve. To avoid using fixed or manually tuned parameters, BWE defines $\tau$ as an adaptive stochastic variable:
\begin{equation}\tau=\big|\gamma+\alpha\cdot\xi\big|,\quad\xi\sim U(0,1)\end{equation}
where $\gamma$ is a time-dependent decay factor and $\alpha$ controls the stochastic fluctuation. In this implementation, the shift factor $\gamma$ is formulated as
\begin{equation}\gamma=\beta-\zeta\end{equation}
where $\beta$ is a constant base and $\zeta$ follows a quadratic decay defined by
\begin{equation}\zeta=\big(1-\vartheta\big)^2\end{equation}

\begin{figure}
    \centering
    \includegraphics[width=0.8\columnwidth]{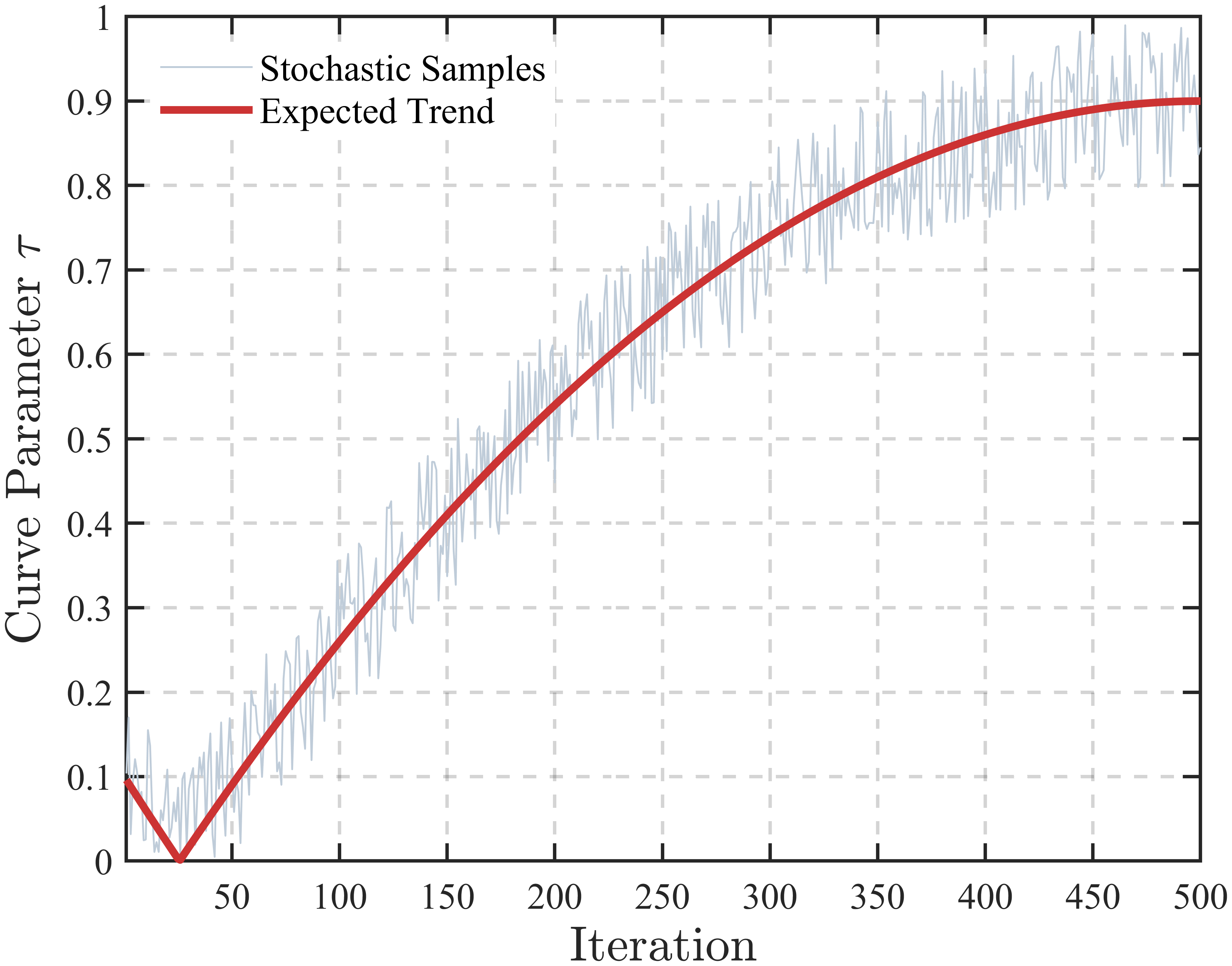}
    \caption{The curve parameter $\tau$ across the iteration process.}
    \label{fig:tau}
\end{figure}

This mechanism ensures that in the early stages of the search, the individuals are positioned with lower $\tau$ values to remain longer in the curvilinear exploration phase of the B\'{e}zier path. As the iteration progresses, the increasing $\gamma$ effectively pushes $\tau$ toward unity, thereby forcing the population to gravitate toward the global elite $\mathbf{x}^\star$ with higher precision. This transition from trajectory-based exploration to destination-oriented exploitation facilitates a robust convergence behavior. The stochastic behavior and the overall path of $\tau$ are visualized in \cref{fig:tau}.

\subsection{Approximate tortuosity perturbation}
The final position update incorporates a stochastic perturbation defined as
\begin{equation}
\tilde{\mathbf{x}}_i \leftarrow \tilde{\mathbf{x}}_i + \rho_i \zeta \tan(\boldsymbol{\varphi}), \quad \boldsymbol{\varphi} \sim \mathcal{N}(\mathbf{0}, \mathbf{I}_D)
\end{equation}
where $\boldsymbol\varphi \in \mathbb{R}^{1 \times D}$ is a multidimensional standard normal vector. The element-wise tangent mapping $\tan(\cdot)$ transforms the Gaussian distribution into a composite heavy-tailed distribution, which effectively facilitates long-range jumps to escape local optima.

To prevent individuals from being strictly confined to the smooth Bézier curve, BWE introduces a perturbation mechanism based on the geometric complexity of the path. This approach ensures that trajectories with higher degrees of freedom receive larger exploratory impulses. First, we calculate the total length of the control polyline as an approximation of the path coverage:
\begin{equation} \mathcal{L}_{i} = \sum_{k=0}^{n-1} \left\| \mathbf{P}_{k+1} - \mathbf{P}_{k} \right\|_2 \end{equation}

To obtain a normalized indicator of trajectory complexity, the polyline length is scaled by the Euclidean distance between the starting point and the global elite:
\begin{equation}
\rho_i = \frac{\mathcal{L}_i}{\left| \mathbf{x}^\star - \mathbf{x}_i \right| + \varepsilon}
\end{equation}
where $\rho_i$ denotes the tortuosity ratio of the path. A larger value of $\rho_i$ implies a more winding trajectory with higher path tension, thereby necessitating a proportional increase in perturbation intensity to balance exploitation and exploration.

\subsection{Boundary handling and population update}
After generating the candidate solution $\tilde{\mathbf{x}}_i$, boundary constraints are enforced via projection:
\begin{equation}\tilde{\mathbf{x}}_i \leftarrow \min\big(\max(\tilde{\mathbf{x}}_i,\boldsymbol{\ell}),\mathbf{u}\big)\end{equation}

Subsequently, a greedy selection mechanism is applied:
\begin{equation}
\mathbf{x}_i^{(t+1)}=
\begin{cases}
\tilde{\mathbf{x}}_i, & f(\tilde{\mathbf{x}}_i) < f(\mathbf{x}_i^{(t)}) \\
\mathbf{x}_i^{(t)}, & \mathrm{otherwise}
\end{cases}
\end{equation}

Meanwhile, the global elite $\mathbf{x}^\star$ is updated whenever a better solution is found. $\mathbf{x}^{\star, T}$ is the optimal solution after $T$ iterations, and also the acceptable solution found by the algorithm upon completing the iterative search.

The pseudocode for BWE is shown in Algorithm~\ref{algorithm1}.

\begin{algorithm2e}[!htbp]
\SetAlgoLined
\caption{Bézier Walk Evolution (BWE)}
\label{algorithm1}
\KwIn{Parameters $\alpha, \beta$, Population size $N$, Bounds $[\boldsymbol{\ell}, \mathbf{u}]$, Maximum number of iterations $T$}

\KwOut{Global elite solution $\mathbf{x}^{\star, T}$}

 \textbf{Initialization:}\\
 Initialize population $\mathcal{P}^{(0)} = \{\mathbf{x}_1, \dots, \mathbf{x}_N\}$ uniformly in $[\boldsymbol{\ell}, \mathbf{u}]$ and current iteration count $t$\;
 Evaluate fitness $f(\mathbf{x}_i)$ and $\mathbf{x}^\star = \arg\min f(\mathbf{x}_i)$\;
 
 \BlankLine
 \textbf{Main Loop:}\\
 \For{$t = 1$ \KwTo $T$}{
  Calculate iteration factor $\vartheta = t/T$\;
  Calculate decay factor $\zeta = (1-\vartheta)^2$\;
  Update strategy weights: $w_3=(1-\vartheta), w_2=3\vartheta(1-\vartheta), w_1=\vartheta$\;
  Calculate weights probabilities $P_k(n=k)$\;
  
  \tcp{Random Walk Sampling}
  Construct sample pool $\mathcal{S} \subset \mathcal{P}^{(t)}$ of size $S$\;
  Calculate distance-based probabilities 
    $p(d) \propto \exp(\tilde{d}_{wv})$\;

  \For{$i = 1$ \KwTo $N$}{
   Select order $n \in \{1, 2, 3\}$ using roulette wheel selection on $P_k$\;
   Define start $\mathbf{P}_0 = \mathbf{x}_i^{(t)}$ and end $\mathbf{P}_n = \mathbf{x}^\star$\;

   \uIf{$n==1$ (First-order)}{
     Set $\mathbf{P}_1 = \mathbf{x}^\star$\;
   }
   \uElseIf{$n==2$ (Second-order)}{
     Select $\mathbf{C}_1 \in \mathcal{S}$ via prob $p_1 \cdot p_3$\;
     Set $\mathbf{P}_1 = \mathbf{C}_1, \mathbf{P}_2 = \mathbf{x}^\star$\;
   }
   \Else{
     \tcp*[h]{n==3 (Third-order)}
     
     Select $\mathbf{C}_1, \mathbf{C}_2 \in \mathcal{S}$ via prob $p_1 \cdot p_2 \cdot p_3$\;
     Set $\mathbf{P}_1 = \mathbf{C}_1, \mathbf{P}_2 = \mathbf{C}_2, \mathbf{P}_3 = \mathbf{x}^\star$\;
   }
   
    \tcp{Bézier evolution} 
    Compute $\tau \leftarrow |\gamma + \alpha \xi|$\;
    $\tilde{\mathbf{x}}_i \leftarrow \sum_{j=0}^{n} \binom{n}{j}(1-\tau)^{n-j}\tau^j \mathbf{P}_j$\;

    \tcp{Tortuosity perturbation}
    Compute polyline length $\mathcal{L}_i \leftarrow \sum_{k=0}^{n-1} \|\mathbf{P}_{k+1}-\mathbf{P}_k\|_2$\;
    Compute tortuosity ratio $\rho_i \leftarrow \mathcal{L}_i / (\|\mathbf{x}^\star-\mathbf{x}_i\|+\varepsilon)$\;
    $\tilde{\mathbf{x}}_i \leftarrow \tilde{\mathbf{x}}_i + \rho_i \zeta \tan(\boldsymbol{\varphi})$\;

   \tcp{Boundary handling \& Update}
    $\tilde{\mathbf{x}}_i \leftarrow \min(\mathbf{u}, \max(\boldsymbol{\ell}, \tilde{\mathbf{x}}_i))$\;
    
   \If{$f(\tilde{\mathbf{x}}_i) < f(\mathbf{x}_i^{(t)})$}{
    $\mathbf{x}_i^{(t+1)} \leftarrow \tilde{\mathbf{x}}_i$\;
    \If{$f(\tilde{\mathbf{x}}_i) < f(\mathbf{x}^\star)$}{
     $\mathbf{x}^\star \leftarrow \tilde{\mathbf{x}}_i$\;
    }
   }
  }
 }
 \KwRet{$\mathbf{x}^{\star, T}$}\;
\end{algorithm2e}

\subsection{Time and space complexity analysis}
The computational complexity of the proposed BWE is analyzed based on the population size $N$, the problem dimension $D$, the maximum number of iterations $T$, and the size of the random walk sample pool $S$.

\subsubsection{Time complexity}
The time complexity is determined by initialization and the iterative search mechanisms involving Bézier curve generation and distance-based sampling.
\begin{enumerate}
    \renewcommand{\labelenumi}{\Roman{enumi}.}
    \item \textbf{Initialization:} Population initialization requires $O(ND)$.
    
    \item \textbf{Iterative Process:} The computational costs per iteration are:
    \begin{itemize}
        \item \textit{Distance Calculation:} Computing topological distances between $N$ individuals and $S$ samples in $D$ dimensions requires $O(NSD)$.
        
        \item \textit{Bézier Update:} Generating Bézier trajectories, applying curvature perturbations, and updating positions requires $O(ND)$.
        
        \item \textit{Boundary Handling:} Boundary checking and correction require $O(ND)$.
    \end{itemize}
\end{enumerate}

Thus, the total time complexity of BWE is $O(\text{BWE}) \approx O(TNSD)$, scaling linearly with the maximum number of iterations, population size, sample size, and dimensionality. Although distance calculation introduces an additional factor of $S$, the sample size is typically kept small and remains controllable in practice, resulting in only a modest computational overhead relative to standard metaheuristics.

\subsubsection{Space complexity}
The space complexity is determined by storing the population, candidate solutions, and auxiliary distance matrices. Specifically, population and candidate storage requires $O(ND)$, while the distance matrices for probability calculation require $O(NS + S^2)$. Therefore, the total space complexity is $O(ND + NS + S^2)$. Assuming $S \ll N$ and $S \ll D$, the dominant term simplifies to $O(ND)$.

\section{Experimental results and analysis}
\label{sec:experimental}
This section evaluates BWE on 41 benchmark functions from the CEC2017 and CEC2022 suites through comparative analysis with competing algorithms.

\subsection{Experimental setup}
All experiments are conducted on a 64-bit Windows 11 system with an Intel Core i9-12900H processor using MATLAB R2023b.
The CEC2017 \citep{wu2017problem} and CEC2022 \citep{biedrzycki2022version} are used for evaluation.

The CEC2017 benchmark suite includes unimodal functions (F1, F3), multimodal functions (F4$\sim$F10), hybrid functions (F11$\sim$F20), and composition functions (F21$\sim$F30). For all algorithms, the population size is set to 50 and the maximum number of function evaluations ($MaxFEs$) is set to $D \times 10^5$. In BWE, each individual generates one trial solution per iteration, resulting in $N$ evaluations per generation; thus, the maximum number of iterations is determined by $T=(MaxFEs-N)/N$. Problems with $D=[10,30,50,100]$ are considered under a search range of $[-100,100]^D$. BWE is compared with the genetic algorithm (GA) \citep{holland1975adaptation}, the love evolution algorithm (LEA) \citep{gao2024love}, the human evolutionary optimization algorithm (HEOA) \citep{lian2024human}, the arithmetic optimization algorithm (AOA) \citep{abualigah2021arithmetic}, the sine cosine algorithm (SCA) \citep{mirjalili2016sca}, the crayfish optimization algorithm (COA) \citep{jia2023crayfish}, and the grey wolf optimizer (GWO) \citep{mirjalili2014grey}.

To further evaluate robustness, BWE is also tested on CEC2022. The benchmark includes unimodal (C1), basic (C2$\sim$C5), hybrid (C6$\sim$C8), and composition (C9$\sim$C12) functions. For all algorithms, the population size is set to 50, the dimension to 10 and 20, and the MaxFEs to $10^6$. The compared SOTAs include CMA-ES \citep{hansen2003reducing}, JADE \citep{zhang2009jade}, L-SHADE \citep{tanabe2014improving}, AL-SHADE \citep{li2022novel}, LSHADE-cnEpSin \citep{awad2016ensemble}, and LSHADE-SPACMA \citep{mohamed2017lshade}.
The parameter settings of BWE and competitive algorithms are shown in \cref{tab:parameter_settings}.

\newcolumntype{A}{>{\raggedright\arraybackslash}p{2.6cm}} 
\newcolumntype{L}{>{\raggedright\arraybackslash}X}       

\begin{table}[!htbp]
\centering
\caption{Parameter settings of BWE and its compared algorithms.}
\label{tab:parameter_settings}
\sffamily
\footnotesize
\setlength{\tabcolsep}{1pt}
\begin{tabularx}{\columnwidth}{@{}A L@{}}
\toprule
\textbf{Algorithm} & \textbf{Parameter Settings} \\
\midrule

BWE &
$\alpha=0.2,\ \beta=0.8,\ \eta=0.2.$ \\

GA &
$P_c=0.8,\ P_m=0.3,\ \mu=0.3.$ \\

LEA &
$h_{\max}=0.7,\ h_{\min}=0,\ \lambda_c=0.5,\ \lambda_p=0.5.$ \\

HEOA &
$A=0.6,\ LN=0.4,\ EN=0.4,\ FN=0.1.$ \\

AOA &
$\alpha=5,\ \mu=0.5.$ \\

SCA &
$\alpha=2.$ \\

COA &
$C_1=0.2,\ C_2=3,\ \sigma=3,\ \mu=25.$ \\

GWO &
$a_{\max}=2,\ a_{\min}=0.$ \\

CMA-ES &
$\sigma=2.$ \\

JADE &
$c=0.1,\ p=0.05,\ CR_m=0.5,\ F_m=0.5,\ A=1.$ \\

L-SHADE &
\makecell[l]{
$H_m=5,\ p=0.11,\ A=1.4,\ N_{\max}=50,$\\
$N_{\min}=4.$
} \\

AL-SHADE &
\makecell[l]{
$M_{CR}=0.5,\ M_F=0.5,\ H_m=6,\ p=0.11,$\\
$A=2.6, N_{\max}=50,\ N_{\min}=4.$
} \\

LSHADE-cnEpSin &
\makecell[l]{
$pd=0.4,\ ps=0.5,\ cf=0.5,\ H_m=5,$\\
$p=0.11,\ A=1.4, N_{\max}=50,\ N_{\min}=4.$
} \\

LSHADE-SPACMA &
\makecell[l]{
$L=0.8,\ H_m=5,\ p=0.11,\ A=1.4,\ F_{cp}=0.5$\\
$N_{\max}=50,\ N_{\min}=4.$
} \\

\bottomrule
\end{tabularx}
\end{table}

\subsection{Performance analysis}
We conduct a comparative analysis between BWE and well-known competing algorithms on the CEC2017 benchmark functions. The experimental results demonstrate BWE's superior performance in solution accuracy, convergence rate, and robustness. \cref{tablecec2017-10D,tablecec2017-30D,tablecec2017-50D,tablecec2017-100D} present the optimization results for $D=10, 30, 50, 100$ (denoted as $D=10\sim100$ for brevity). Best results are marked in bold. "Mean", "Min", and "Std" represent the average, minimum, and standard deviation of 51 independent runs, respectively. 

To verify BWE's superiority, two statistical methods are employed. The Wilcoxon rank-sum test is implemented at a significance level of $p = 0.05$, where symbols "$<$", "$\approx$", and "$>$" denote the competitor achieved significantly better, similar, or worse performance than BWE, respectively. Additionally, the Friedman rank test is used to provide an overall assessment across various optimization landscapes.

\begin{table*}[!htbp]
\centering
\caption{The results of BWE and its competitors in CEC2017 10D using Wilcoxon rank-sum test and Friedman rank test.}
\label{tablecec2017-10D}
\sffamily
\scriptsize 
\setlength{\tabcolsep}{2.5pt} 
\begin{tabularx}{\textwidth}{@{}llXXXXXXXX@{}}
\toprule
No. & & BWE & GA & LEA & HEOA & AOA & SCA & COA & GWO \\ \midrule
F1 & Mean & \textbf{5.62E+02} & 1.56E+07 ($>$) & 5.25E+03 ($>$) & 7.77E+08 ($>$) & 2.35E+09 ($>$) & 6.17E+08 ($>$) & 3.40E+03 ($>$) & 4.48E+07 ($>$) \\
   & Std  & \textbf{5.12E+02} & 5.84E+07       & 3.83E+03       & 8.86E+08       & 2.16E+09       & 2.25E+08       & 3.42E+03       & 1.47E+16       \\
F3 & Mean & \textbf{3.00E+02} & 4.86E+04 ($>$) & 3.00E+02 ($>$) & 2.12E+04 ($>$) & 3.68E+03 ($>$) & 1.21E+03 ($>$) & 3.00E+02 ($>$) & 1.65E+03 ($>$) \\
   & Std  & \textbf{2.16E-08} & 2.03E+04       & 1.68E-07       & 1.44E+04       & 1.45E+03       & 5.87E+02       & 9.47E-04       & 4.90E+06       \\
F4 & Mean & \textbf{4.02E+02} & 4.67E+02 ($>$) & 4.04E+02 ($>$) & 4.47E+02 ($>$) & 5.52E+02 ($>$) & 4.37E+02 ($>$) & 4.04E+02 ($>$) & 4.19E+02 ($>$) \\
   & Std  & \textbf{4.41E-01} & 5.16E+01       & 1.30E+01       & 5.01E+01       & 1.39E+02       & 1.25E+01       & 2.58E+00       & 5.44E+02       \\
F5 & Mean & \textbf{5.13E+02} & 5.41E+02 ($>$) & 5.21E+02 ($>$) & 5.66E+02 ($>$) & 5.51E+02 ($>$) & 5.46E+02 ($>$) & 5.13E+02 ($\approx$) & 5.15E+02 ($\approx$) \\
   & Std  & \textbf{4.26E+00} & 1.43E+01       & 8.36E+00       & 2.00E+01       & 1.74E+01       & 6.26E+00       & 5.53E+00       & 6.25E+01       \\
F6 & Mean & \textbf{6.00E+02} & 6.31E+02 ($>$) & 6.00E+02 ($>$) & 6.44E+02 ($>$) & 6.37E+02 ($>$) & 6.16E+02 ($>$) & 6.03E+02 ($>$) & 6.01E+02 ($>$) \\
   & Std  & \textbf{2.41E-04} & 1.07E+01       & 1.45E-01       & 1.22E+01       & 7.34E+00       & 3.40E+00       & 6.91E+00       & 6.83E-01       \\
F7 & Mean & \textbf{7.16E+02} & 7.64E+02 ($>$) & 7.30E+02 ($>$) & 7.99E+02 ($>$) & 7.99E+02 ($>$) & 7.71E+02 ($>$) & 7.56E+02 ($>$) & 7.25E+02 ($>$) \\
   & Std  & \textbf{2.67E+00} & 1.49E+01       & 1.05E+01       & 2.62E+01       & 1.35E+01       & 7.93E+00       & 2.73E+01       & 5.74E+01       \\
F8 & Mean & \textbf{8.12E+02} & 8.50E+02 ($>$) & 8.23E+02 ($>$) & 8.43E+02 ($>$) & 8.31E+02 ($>$) & 8.37E+02 ($>$) & 8.26E+02 ($>$) & 8.12E+02 ($\approx$) \\
   & Std  & \textbf{4.04E+00} & 1.56E+01       & 9.72E+00       & 1.37E+01       & 6.54E+00       & 5.90E+00       & 9.02E+00       & 3.16E+01       \\
F9 & Mean & \textbf{9.00E+02} & 1.06E+03 ($>$) & 9.00E+02 ($\approx$) & 1.63E+03 ($>$) & 1.35E+03 ($>$) & 9.74E+02 ($>$) & 9.36E+02 ($>$) & 9.10E+02 ($>$) \\
   & Std  & \textbf{1.10E-08} & 1.88E+02       & 9.58E-01       & 2.88E+02       & 1.87E+02       & 3.54E+01       & 8.92E+01       & 7.73E+02       \\
F10& Mean & 1.64E+03          & 1.78E+03 ($>$) & 1.62E+03 ($\approx$) & 2.19E+03 ($>$) & 2.01E+03 ($>$) & 2.14E+03 ($>$) & 1.82E+03 ($>$) & \textbf{1.47E+03 ($<$)} \\
   & Std  & \textbf{2.01E+02} & 3.16E+02       & 2.33E+02       & 3.12E+02       & 2.98E+02       & 2.20E+02       & 3.27E+02       & 6.85E+04       \\
F11& Mean & \textbf{1.11E+03} & 3.89E+03 ($>$) & 1.13E+03 ($>$) & 1.90E+03 ($>$) & 1.16E+03 ($>$) & 1.18E+03 ($>$) & 1.12E+03 ($>$) & 1.13E+03 ($>$) \\
   & Std  & \textbf{3.01E+00} & 4.67E+03       & 1.93E+01       & 2.24E+03       & 3.99E+01       & 2.21E+01       & 2.57E+01       & 6.12E+02       \\
F12& Mean & \textbf{9.81E+03} & 5.12E+06 ($>$) & 4.57E+04 ($>$) & 6.51E+06 ($>$) & 1.04E+06 ($>$) & 1.07E+07 ($>$) & 4.43E+04 ($\approx$) & 6.51E+05 ($>$) \\
   & Std  & \textbf{3.69E+03} & 6.52E+06       & 2.98E+04       & 4.47E+06       & 1.79E+06       & 8.73E+06       & 1.87E+05       & 5.30E+11       \\
F13& Mean & 8.44E+03          & 1.36E+04 ($\approx$) & 1.03E+04 ($\approx$) & 1.60E+04 ($>$) & 1.17E+04 ($\approx$) & 2.43E+04 ($>$) & \textbf{3.51E+03 ($<$)} & 1.03E+04 ($\approx$) \\
   & Std  & \textbf{2.71E+03} & 1.19E+04       & 9.64E+03       & 9.66E+03       & 8.24E+03       & 1.57E+04       & 4.40E+03       & 4.41E+07       \\
F14& Mean & 1.64E+03          & 8.88E+03 ($>$) & \textbf{1.47E+03 ($<$)} & 3.65E+03 ($>$) & 8.37E+03 ($>$) & 1.57E+03 ($\approx$) & 1.50E+03 ($\approx$) & 2.99E+03 ($>$) \\
   & Std  & 2.48E+02          & 7.45E+03       & \textbf{3.23E+01}       & 3.65E+03       & 8.12E+03       & 6.13E+01       & 3.90E+01       & 3.17E+06       \\
F15& Mean & 1.68E+03          & 1.06E+04 ($>$) & \textbf{1.63E+03 ($\approx$)} & 1.26E+04 ($>$) & 1.15E+04 ($>$) & 1.91E+03 ($>$) & 1.72E+03 ($>$) & 2.78E+03 ($>$) \\
   & Std  & 1.94E+02          & 9.15E+03       & \textbf{5.65E+01}       & 6.49E+03       & 5.23E+03       & 3.38E+02       & 1.38E+02       & 1.88E+06       \\
F16& Mean & \textbf{1.65E+03} & 1.81E+03 ($>$) & 1.67E+03 ($>$) & 2.09E+03 ($>$) & 1.97E+03 ($>$) & 1.71E+03 ($>$) & 1.65E+03 ($\approx$) & 1.72E+03 ($>$) \\
   & Std  & 5.49E+01          & 1.30E+02       & 7.67E+01       & 1.73E+02       & 1.43E+02       & \textbf{4.25E+01} & 9.85E+01       & 1.11E+04       \\
F17& Mean & 1.73E+03          & 1.80E+03 ($>$) & 1.75E+03 ($>$) & 1.93E+03 ($>$) & 1.84E+03 ($>$) & 1.77E+03 ($>$) & \textbf{1.73E+03 ($<$)} & 1.76E+03 ($>$) \\
   & Std  & 1.11E+01          & 5.40E+01       & 2.85E+01       & 1.01E+02       & 8.24E+01       & \textbf{8.56E+00} & 2.54E+01       & 1.37E+03       \\
F18& Mean & \textbf{4.25E+03} & 1.47E+04 ($>$) & 2.07E+04 ($>$) & 1.47E+06 ($>$) & 1.45E+04 ($>$) & 8.03E+04 ($>$) & 1.13E+04 ($>$) & 2.50E+04 ($>$) \\
   & Std  & \textbf{2.26E+03} & 1.19E+04       & 1.37E+04       & 4.93E+06       & 9.43E+03       & 6.92E+04       & 8.55E+03       & 2.66E+08       \\
F19& Mean & 2.34E+03          & 1.04E+04 ($>$) & 2.00E+03 ($<$) & 8.37E+04 ($>$) & 2.15E+04 ($>$) & 3.16E+03 ($>$) & \textbf{2.00E+03 ($<$)} & 4.81E+03 ($\approx$) \\
   & Std  & 5.05E+02          & 8.68E+03       & 2.34E+02       & 3.72E+05       & 2.14E+04       & 2.74E+03       & \textbf{5.73E+01} & 2.45E+07       \\
F20& Mean & 2.03E+03          & 2.13E+03 ($>$) & 2.03E+03 ($>$) & 2.23E+03 ($>$) & 2.13E+03 ($>$) & 2.08E+03 ($>$) & \textbf{2.02E+03 ($<$)} & 2.06E+03 ($>$) \\
   & Std  & \textbf{1.21E+01} & 6.14E+01       & 1.59E+01       & 9.78E+01       & 5.25E+01       & 1.95E+01       & 1.94E+01       & 1.70E+03       \\
F21& Mean & \textbf{2.20E+03} & 2.37E+03 ($>$) & 2.20E+03 ($>$) & 2.36E+03 ($>$) & 2.31E+03 ($>$) & 2.22E+03 ($>$) & 2.30E+03 ($>$) & 2.30E+03 ($>$) \\
   & Std  & 1.41E+01          & 2.74E+01       & \textbf{1.50E+00} & 1.82E+01       & 4.08E+01       & 3.08E+01       & 3.99E+01       & 9.11E+02       \\
F22& Mean & 2.30E+03          & 2.42E+03 ($>$) & \textbf{2.30E+03 ($<$)} & 2.56E+03 ($>$) & 2.61E+03 ($>$) & 2.35E+03 ($>$) & 2.30E+03 ($\approx$) & 2.33E+03 ($>$) \\
   & Std  & \textbf{8.44E+00} & 2.05E+02       & 2.29E+01       & 5.43E+02       & 1.91E+02       & 2.59E+01       & 1.37E+01       & 1.82E+04       \\
F23& Mean & \textbf{2.61E+03} & 2.68E+03 ($>$) & 2.61E+03 ($>$) & 2.72E+03 ($>$) & 2.70E+03 ($>$) & 2.65E+03 ($>$) & 2.61E+03 ($\approx$) & 2.62E+03 ($\approx$) \\
   & Std  & 4.41E+01          & 2.45E+01       & 6.35E+01       & 4.43E+01       & 3.66E+01       & 7.42E+00       & \textbf{6.01E+00} & 1.00E+02       \\
F24& Mean & 2.57E+03          & 2.81E+03 ($>$) & \textbf{2.52E+03 ($<$)} & 2.80E+03 ($>$) & 2.80E+03 ($>$) & 2.75E+03 ($>$) & 2.75E+03 ($>$) & 2.74E+03 ($>$) \\
   & Std  & 1.11E+02          & 4.69E+01       & 8.26E+01       & 8.84E+01       & 6.09E+01       & 7.52E+01       & \textbf{7.02E+00} & 1.23E+02       \\
F25& Mean & 2.93E+03          & 2.99E+03 ($>$) & \textbf{2.93E+03 ($<$)} & 2.98E+03 ($>$) & 3.05E+03 ($>$) & 2.95E+03 ($>$) & 2.93E+03 ($>$) & 2.93E+03 ($>$) \\
   & Std  & 2.11E+01          & 3.50E+01       & 3.04E+01       & 4.68E+01       & 8.75E+01       & \textbf{1.64E+01} & 3.08E+01       & 3.48E+02       \\
F26& Mean & \textbf{2.83E+03} & 3.51E+03 ($>$) & 2.95E+03 ($>$) & 3.56E+03 ($>$) & 3.73E+03 ($>$) & 3.05E+03 ($>$) & 3.08E+03 ($>$) & 3.07E+03 ($>$) \\
   & Std  & 9.22E+01          & 4.03E+02       & 4.86E+01       & 3.84E+02       & 3.41E+02       & \textbf{3.60E+01} & 3.19E+02       & 1.17E+05       \\
F27& Mean & 3.10E+03          & 3.12E+03 ($>$) & \textbf{3.09E+03 ($<$)} & 3.19E+03 ($>$) & 3.21E+03 ($>$) & 3.10E+03 ($>$) & 3.10E+03 ($\approx$) & 3.10E+03 ($\approx$) \\
   & Std  & 5.59E+00          & 2.65E+01       & 3.33E+00       & 4.90E+01       & 4.09E+01       & \textbf{1.70E+00} & 1.68E+01       & 2.35E+02       \\
F28& Mean & 3.31E+03          & 3.33E+03 ($>$) & \textbf{3.19E+03 ($<$)} & 3.54E+03 ($>$) & 3.56E+03 ($>$) & 3.25E+03 ($<$) & 3.29E+03 ($\approx$) & 3.37E+03 ($>$) \\
   & Std  & 1.34E+02          & 8.76E+01       & 1.20E+02       & 9.07E+01       & 1.65E+02       & \textbf{5.33E+01} & 1.68E+02       & 6.54E+03       \\
F29& Mean & \textbf{3.18E+03} & 3.25E+03 ($>$) & 3.18E+03 ($\approx$) & 3.35E+03 ($>$) & 3.33E+03 ($>$) & 3.22E+03 ($>$) & 3.18E+03 ($\approx$) & 3.18E+03 ($\approx$) \\
   & Std  & \textbf{1.87E+01} & 5.98E+01       & 3.75E+01       & 9.90E+01       & 1.02E+02       & 3.01E+01       & 3.47E+01       & 1.72E+03       \\
F30& Mean & \textbf{8.08E+03} & 5.91E+05 ($>$) & 2.36E+05 ($>$) & 2.48E+06 ($>$) & 6.08E+06 ($>$) & 5.65E+05 ($>$) & 3.61E+05 ($>$) & 5.96E+05 ($>$) \\
   & Std  & \textbf{1.60E+03} & 8.63E+05       & 4.15E+05       & 2.29E+06       & 8.91E+06       & 5.27E+05       & 4.89E+05       & 9.00E+11       \\ \midrule
Wilcoxon & $+/\approx/-$ & 0/29/0 & 28/1/0 & 16/7/6 & 29/0/0 & 28/1/0 & 27/1/1 & 18/7/4 & 21/7/1 \\ \midrule
FM-rank & & \textbf{1.6552} & 6.1379 & 2.3448 & 7.3103 & 6.7586 & 5.0345 & 2.8276 & 3.9310 \\
M-rank & & \textbf{1} & 6 & 2 & 8 & 7 & 5 & 3 & 4 \\ \bottomrule
\end{tabularx}
\end{table*}

\begin{table*}[!htbp]
\centering
\caption{The results of BWE and its competitors in CEC2017 30D using Wilcoxon rank-sum test and Friedman rank test.}
\label{tablecec2017-30D}
\sffamily
\scriptsize
\setlength{\tabcolsep}{2pt}
\begin{tabularx}{\textwidth}{@{}llXXXXXXXX@{}}
\toprule
No. & & BWE & GA & LEA & HEOA & AOA & SCA & COA & GWO \\ \midrule
F1  & Mean & \textbf{2.51E+03} & 2.99E+09 ($>$) & 9.49E+03 ($>$) & 1.02E+10 ($>$) & 4.19E+10 ($>$) & 1.22E+10 ($>$) & 9.54E+04 ($>$) & 1.41E+09 ($>$) \\
    & Std  & \textbf{2.07E+03} & 2.80E+09 & 7.27E+03 & 4.50E+09 & 5.68E+09 & 2.09E+09 & 4.33E+05 & 9.53E+17 \\
F3  & Mean & \textbf{3.00E+02} & 2.54E+05 ($>$) & 3.00E+02 ($>$) & 1.37E+05 ($>$) & 7.34E+04 ($>$) & 3.74E+04 ($>$) & 5.77E+03 ($>$) & 3.29E+04 ($>$) \\
    & Std  & 5.99E-05 & 6.62E+04 & \textbf{5.97E-05} & 5.25E+04 & 8.89E+03 & 5.85E+03 & 3.47E+03 & 1.04E+08 \\
F4  & Mean & 4.96E+02 & 1.22E+03 ($>$) & 4.90E+02 ($\approx$) & 9.57E+02 ($>$) & 9.65E+03 ($>$) & 1.45E+03 ($>$) & \textbf{4.90E+02} ($\approx$) & 5.89E+02 ($>$) \\
    & Std  & 1.75E+01 & 4.13E+02 & \textbf{1.45E+01} & 4.04E+02 & 2.73E+03 & 2.62E+02 & 2.91E+01 & 2.80E+04 \\
F5  & Mean & \textbf{5.54E+02} & 8.17E+02 ($>$) & 6.31E+02 ($>$) & 8.18E+02 ($>$) & 8.12E+02 ($>$) & 7.79E+02 ($>$) & 7.16E+02 ($>$) & 5.90E+02 ($>$) \\
    & Std  & \textbf{1.05E+01} & 6.11E+01 & 3.44E+01 & 4.22E+01 & 3.33E+01 & 1.77E+01 & 7.50E+01 & 5.65E+02 \\
F6  & Mean & \textbf{6.00E+02} & 7.05E+02 ($>$) & 6.07E+02 ($>$) & 6.66E+02 ($>$) & 6.66E+02 ($>$) & 6.50E+02 ($>$) & 6.30E+02 ($>$) & 6.06E+02 ($>$) \\
    & Std  & \textbf{1.84E-02} & 1.39E+01 & 6.79E+00 & 7.89E+00 & 5.89E+00 & 5.79E+00 & 1.63E+01 & 1.03E+01 \\
F7  & Mean & \textbf{7.56E+02} & 1.29E+03 ($>$) & 8.70E+02 ($>$) & 1.45E+03 ($>$) & 1.29E+03 ($>$) & 1.13E+03 ($>$) & 1.12E+03 ($>$) & 8.58E+02 ($>$) \\
    & Std  & \textbf{7.11E+00} & 1.33E+02 & 3.74E+01 & 1.31E+02 & 6.53E+01 & 4.07E+01 & 1.37E+02 & 2.67E+03 \\
F8  & Mean & \textbf{8.57E+02} & 1.13E+03 ($>$) & 9.43E+02 ($>$) & 1.06E+03 ($>$) & 1.04E+03 ($>$) & 1.05E+03 ($>$) & 9.68E+02 ($>$) & 8.84E+02 ($>$) \\
    & Std  & \textbf{1.34E+01} & 5.38E+01 & 3.27E+01 & 3.80E+01 & 3.12E+01 & 1.58E+01 & 2.70E+01 & 5.69E+02 \\
F9  & Mean & \textbf{9.00E+02} & 6.37E+03 ($>$) & 4.14E+03 ($>$) & 8.91E+03 ($>$) & 5.62E+03 ($>$) & 5.34E+03 ($>$) & 4.35E+03 ($>$) & 1.57E+03 ($>$) \\
    & Std  & \textbf{6.74E-01} & 2.02E+03 & 2.14E+03 & 2.02E+03 & 7.68E+02 & 9.91E+02 & 1.62E+03 & 1.41E+05 \\
F10 & Mean & \textbf{3.77E+03} & 5.95E+03 ($>$) & 4.29E+03 ($>$) & 6.34E+03 ($>$) & 6.38E+03 ($>$) & 8.11E+03 ($>$) & 5.23E+03 ($>$) & 3.91E+03 ($\approx$) \\
    & Std  & 4.90E+02 & 8.13E+02 & 5.01E+02 & 6.88E+02 & 5.76E+02 & \textbf{3.61E+02} & 5.51E+02 & 3.17E+05 \\
F11 & Mean & \textbf{1.16E+03} & 1.34E+04 ($>$) & 1.30E+03 ($>$) & 5.05E+03 ($>$) & 3.77E+03 ($>$) & 2.08E+03 ($>$) & 1.32E+03 ($>$) & 1.79E+03 ($>$) \\
    & Std  & \textbf{2.21E+01} & 8.74E+03 & 7.22E+01 & 2.60E+03 & 1.34E+03 & 3.15E+02 & 5.98E+01 & 6.03E+05 \\
F12 & Mean & \textbf{7.01E+05} & 1.15E+08 ($>$) & 2.20E+06 ($>$) & 2.55E+08 ($>$) & 8.12E+09 ($>$) & 1.21E+09 ($>$) & 9.90E+05 ($\approx$) & 4.89E+07 ($>$) \\
    & Std  & \textbf{2.81E+05} & 1.03E+08 & 1.29E+06 & 2.32E+08 & 2.13E+09 & 2.25E+08 & 8.44E+05 & 3.97E+15 \\
F13 & Mean & \textbf{2.05E+04} & 1.10E+08 ($>$) & 9.45E+04 ($>$) & 2.96E+07 ($>$) & 4.25E+04 ($>$) & 3.79E+08 ($>$) & 2.80E+04 ($\approx$) & 5.25E+06 ($>$) \\
    & Std  & \textbf{7.86E+03} & 2.33E+08 & 5.07E+04 & 1.66E+08 & 1.51E+04 & 1.49E+08 & 1.88E+04 & 5.04E+14 \\
F14 & Mean & \textbf{3.49E+03} & 8.22E+06 ($>$) & 1.04E+04 ($>$) & 2.26E+06 ($>$) & 5.20E+04 ($>$) & 1.26E+05 ($>$) & 2.38E+04 ($>$) & 2.14E+05 ($>$) \\
    & Std  & \textbf{1.86E+03} & 1.33E+07 & 4.65E+03 & 1.75E+06 & 4.87E+04 & 8.84E+04 & 2.59E+04 & 1.54E+11 \\
F15 & Mean & \textbf{4.21E+03} & 1.83E+07 ($>$) & 3.61E+04 ($>$) & 1.45E+05 ($>$) & 2.44E+04 ($>$) & 1.48E+07 ($>$) & 1.06E+04 ($\approx$) & 1.50E+06 ($>$) \\
    & Std  & \textbf{1.57E+03} & 9.07E+07 & 1.90E+04 & 6.15E+05 & 1.20E+04 & 1.36E+07 & 1.14E+04 & 4.52E+13 \\
F16 & Mean & \textbf{2.24E+03} & 3.61E+03 ($>$) & 2.53E+03 ($>$) & 3.47E+03 ($>$) & 3.96E+03 ($>$) & 3.59E+03 ($>$) & 2.50E+03 ($>$) & 2.34E+03 ($\approx$) \\
    & Std  & \textbf{1.68E+02} & 3.93E+02 & 2.85E+02 & 5.98E+02 & 6.48E+02 & 2.33E+02 & 2.83E+02 & 6.27E+04 \\
F17 & Mean & \textbf{1.87E+03} & 2.79E+03 ($>$) & 2.16E+03 ($>$) & 2.56E+03 ($>$) & 2.76E+03 ($>$) & 2.43E+03 ($>$) & 2.03E+03 ($>$) & 1.96E+03 ($>$) \\
    & Std  & \textbf{8.87E+01} & 3.04E+02 & 1.86E+02 & 3.16E+02 & 3.19E+02 & 1.79E+02 & 1.51E+02 & 2.03E+04 \\
F18 & Mean & \textbf{8.15E+04} & 1.35E+07 ($>$) & 2.50E+05 ($>$) & 1.18E+07 ($>$) & 8.81E+05 ($>$) & 2.81E+06 ($>$) & 3.49E+05 ($>$) & 9.48E+05 ($>$) \\
    & Std  & \textbf{2.89E+04} & 1.37E+07 & 1.80E+05 & 1.36E+07 & 1.16E+06 & 1.45E+06 & 3.30E+05 & 1.51E+12 \\
F19 & Mean & \textbf{4.90E+03} & 3.50E+06 ($>$) & 2.28E+04 ($>$) & 2.66E+06 ($>$) & 1.09E+06 ($>$) & 2.35E+07 ($>$) & 1.09E+04 ($>$) & 7.31E+05 ($>$) \\
    & Std  & \textbf{2.05E+03} & 4.09E+06 & 1.76E+04 & 2.94E+06 & 1.38E+05 & 1.16E+07 & 1.22E+04 & 2.39E+12 \\
F20 & Mean & \textbf{2.22E+03} & 2.99E+03 ($>$) & 2.43E+03 ($>$) & 2.88E+03 ($>$) & 2.73E+03 ($>$) & 2.63E+03 ($>$) & 2.40E+03 ($>$) & 2.35E+03 ($>$) \\
    & Std  & \textbf{4.64E+01} & 2.57E+02 & 1.41E+02 & 2.27E+02 & 1.77E+02 & 1.22E+02 & 1.79E+02 & 1.31E+04 \\
F21 & Mean & \textbf{2.35E+03} & 2.75E+03 ($>$) & 2.43E+03 ($>$) & 2.57E+03 ($>$) & 2.60E+03 ($>$) & 2.55E+03 ($>$) & 2.43E+03 ($>$) & 2.38E+03 ($>$) \\
    & Std  & 2.51E+01 & 7.62E+01 & 3.42E+01 & 4.92E+01 & 5.16E+01 & \textbf{2.00E+01} & 3.89E+01 & 4.19E+02 \\
F22 & Mean & \textbf{2.30E+03} & 7.23E+03 ($>$) & 3.74E+03 ($>$) & 7.61E+03 ($>$) & 8.00E+03 ($>$) & 8.65E+03 ($>$) & 2.88E+03 ($>$) & 4.00E+03 ($>$) \\
    & Std  & \textbf{5.83E-01} & 1.33E+03 & 1.71E+03 & 1.68E+03 & 9.23E+02 & 2.07E+03 & 1.60E+03 & 2.55E+06 \\
F23 & Mean & \textbf{2.70E+03} & 3.21E+03 ($>$) & 2.77E+03 ($>$) & 3.25E+03 ($>$) & 3.43E+03 ($>$) & 2.98E+03 ($>$) & 2.82E+03 ($>$) & 2.76E+03 ($>$) \\
    & Std  & \textbf{1.37E+01} & 1.12E+02 & 3.57E+01 & 1.56E+02 & 1.49E+02 & 2.97E+01 & 5.90E+01 & 2.14E+03 \\
F24 & Mean & \textbf{2.86E+03} & 3.38E+03 ($>$) & 2.95E+03 ($>$) & 3.37E+03 ($>$) & 3.75E+03 ($>$) & 3.16E+03 ($>$) & 2.95E+03 ($>$) & 2.92E+03 ($>$) \\
    & Std  & \textbf{1.40E+01} & 8.32E+01 & 3.92E+01 & 1.35E+02 & 1.59E+02 & 3.19E+01 & 4.07E+01 & 2.26E+03 \\
F25 & Mean & \textbf{2.89E+03} & 3.81E+03 ($>$) & 2.89E+03 ($\approx$) & 3.11E+03 ($>$) & 4.29E+03 ($>$) & 3.20E+03 ($>$) & 2.90E+03 ($>$) & 2.96E+03 ($>$) \\
    & Std  & 2.71E+00 & 2.91E+02 & \textbf{1.65E+00} & 8.17E+01 & 4.15E+02 & 5.07E+01 & 2.17E+01 & 1.01E+03 \\
F26 & Mean & \textbf{3.21E+03} & 7.27E+03 ($>$) & 5.04E+03 ($>$) & 7.30E+03 ($>$) & 9.79E+03 ($>$) & 6.94E+03 ($>$) & 5.38E+03 ($>$) & 4.50E+03 ($>$) \\
    & Std  & 7.76E+02 & 6.82E+02 & 4.01E+02 & 1.62E+03 & 9.62E+02 & \textbf{3.03E+02} & 1.56E+03 & 1.32E+05 \\
F27 & Mean & \textbf{3.22E+03} & 3.39E+03 ($>$) & 3.23E+03 ($>$) & 3.45E+03 ($>$) & 4.41E+03 ($>$) & 3.40E+03 ($>$) & 3.26E+03 ($>$) & 3.24E+03 ($>$) \\
    & Std  & \textbf{5.52E+00} & 1.34E+02 & 1.65E+01 & 1.25E+02 & 3.05E+02 & 4.54E+01 & 3.19E+01 & 3.02E+02 \\
F28 & Mean & \textbf{3.11E+03} & 4.66E+03 ($>$) & 3.20E+03 ($>$) & 3.85E+03 ($>$) & 5.95E+03 ($>$) & 3.83E+03 ($>$) & 3.21E+03 ($>$) & 3.38E+03 ($>$) \\
    & Std  & 3.28E+01 & 7.79E+02 & 6.53E+01 & 2.97E+02 & 6.46E+02 & 1.24E+02 & \textbf{3.20E+01} & 3.91E+03 \\
F29 & Mean & \textbf{3.48E+03} & 4.71E+03 ($>$) & 3.71E+03 ($>$) & 5.00E+03 ($>$) & 5.81E+03 ($>$) & 4.66E+03 ($>$) & 3.79E+03 ($>$) & 3.71E+03 ($>$) \\
    & Std  & \textbf{1.15E+02} & 4.70E+02 & 1.90E+02 & 5.14E+02 & 8.99E+02 & 2.27E+02 & 2.24E+02 & 2.84E+04 \\
F30 & Mean & \textbf{1.17E+04} & 9.92E+06 ($>$) & 4.14E+04 ($>$) & 1.92E+07 ($>$) & 3.13E+07 ($>$) & 6.63E+07 ($>$) & 2.38E+04 ($>$) & 5.53E+06 ($>$) \\
    & Std  & \textbf{2.38E+03} & 1.28E+07 & 2.00E+04 & 3.28E+07 & 9.36E+07 & 2.59E+07 & 2.01E+04 & 2.15E+13 \\ \midrule
Wilcoxon & $+/\approx/-$ & 0/29/0 & 29/0/0 & 27/2/0 & 29/0/0 & 29/0/0 & 27/1/1 & 25/4/0 & 27/2/0 \\ \midrule
FM-rank & & \textbf{1.1034} & 6.7586 & 2.8276 & 6.4138 & 6.5862 & 5.9655 & 3.1034 & 3.2414 \\
M-rank  & & \textbf{1} & 8 & 2 & 6 & 7 & 5 & 3 & 4 \\ \bottomrule
\end{tabularx}
\end{table*}

\begin{table*}[!htbp]
\centering
\caption{The results of BWE and its competitors in CEC2017 50D using Wilcoxon rank-sum test and Friedman rank test.}
\label{tablecec2017-50D}
\sffamily
\scriptsize
\setlength{\tabcolsep}{2pt}
\begin{tabularx}{\textwidth}{@{}llXXXXXXXX@{}}
\toprule
No. & & BWE & GA & LEA & HEOA & AOA & SCA & COA & GWO \\ \midrule
F1  & Mean & \textbf{8.31E+02} & 6.78E+08 ($>$) & 1.41E+04 ($>$) & 2.42E+10 ($>$) & 1.03E+11 ($>$) & 3.94E+10 ($>$) & 2.45E+06 ($>$) & 6.55E+09 ($>$) \\
    & Std  & \textbf{9.06E+02} & 7.53E+17 & 1.15E+04 & 9.97E+19 & 6.10E+19 & 2.29E+19 & 4.23E+13 & 8.20E+18 \\
F3  & Mean & 3.00E+02 & 4.58E+05 ($>$) & \textbf{3.00E+02 ($<$)} & 2.28E+05 ($>$) & 1.66E+05 ($>$) & 9.62E+04 ($>$) & 6.30E+04 ($>$) & 7.78E+04 ($>$) \\
    & Std  & 4.26E-02 & 1.02E+10 & \textbf{3.48E-04} & 1.35E+09 & 3.94E+08 & 1.19E+08 & 4.45E+08 & 2.28E+08 \\
F4  & Mean & \textbf{4.90E+02} & 1.12E+03 ($>$) & 5.33E+02 ($>$) & 3.12E+03 ($>$) & 2.64E+04 ($>$) & 6.07E+03 ($>$) & 5.35E+02 ($>$) & 1.02E+03 ($>$) \\
    & Std  & \textbf{4.76E+01} & 6.45E+04 & 5.02E+01 & 1.78E+06 & 1.91E+07 & 1.20E+06 & 3.77E+03 & 1.03E+05 \\
F5  & Mean & \textbf{6.00E+02} & 1.14E+03 ($>$) & 7.72E+02 ($>$) & 1.01E+03 ($>$) & 1.06E+03 ($>$) & 1.05E+03 ($>$) & 8.63E+02 ($>$) & 6.87E+02 ($>$) \\
    & Std  & \textbf{1.75E+01} & 5.73E+03 & 5.62E+01 & 2.31E+03 & 1.61E+03 & 7.36E+02 & 1.36E+03 & 1.14E+03 \\
F6  & Mean & \textbf{6.00E+02} & 7.23E+02 ($>$) & 6.28E+02 ($>$) & 6.74E+02 ($>$) & 6.83E+02 ($>$) & 6.69E+02 ($>$) & 6.52E+02 ($>$) & 6.12E+02 ($>$) \\
    & Std  & \textbf{2.54E-02} & 1.01E+02 & 1.04E+01 & 3.87E+01 & 3.24E+01 & 2.79E+01 & 2.04E+02 & 1.83E+01 \\
F7  & Mean & \textbf{8.02E+02} & 1.61E+03 ($>$) & 1.11E+03 ($>$) & 2.19E+03 ($>$) & 1.85E+03 ($>$) & 1.63E+03 ($>$) & 1.63E+03 ($>$) & 1.03E+03 ($>$) \\
    & Std  & \textbf{9.43E+00} & 1.99E+04 & 6.61E+01 & 5.07E+04 & 4.75E+03 & 6.37E+03 & 2.30E+04 & 3.03E+03 \\
F8  & Mean & \textbf{8.94E+02} & 1.44E+03 ($>$) & 1.08E+03 ($>$) & 1.40E+03 ($>$) & 1.42E+03 ($>$) & 1.36E+03 ($>$) & 1.20E+03 ($>$) & 1.00E+03 ($>$) \\
    & Std  & \textbf{1.56E+01} & 4.12E+03 & 5.86E+01 & 4.90E+03 & 1.49E+03 & 7.35E+02 & 5.45E+02 & 1.43E+03 \\
F9  & Mean & \textbf{9.04E+02} & 1.79E+04 ($>$) & 1.90E+04 ($>$) & 2.23E+04 ($>$) & 2.28E+04 ($>$) & 2.25E+04 ($>$) & 1.62E+04 ($>$) & 5.86E+03 ($>$) \\
    & Std  & \textbf{3.51E+00} & 2.22E+07 & 7.57E+03 & 1.37E+07 & 8.68E+06 & 1.55E+07 & 6.80E+06 & 8.88E+06 \\
F10 & Mean & \textbf{6.02E+03} & 9.77E+03 ($>$) & 7.25E+03 ($>$) & 1.04E+04 ($>$) & 1.25E+04 ($>$) & 1.44E+04 ($>$) & 9.72E+03 ($>$) & 6.50E+03 ($\approx$) \\
    & Std  & \textbf{6.53E+02} & 1.44E+06 & 8.53E+02 & 9.10E+05 & 5.96E+05 & 9.91E+04 & 9.85E+05 & 6.02E+05 \\
F11 & Mean & \textbf{1.20E+03} & 4.04E+04 ($>$) & 1.45E+03 ($>$) & 7.90E+03 ($>$) & 1.57E+04 ($>$) & 5.87E+03 ($>$) & 1.38E+03 ($>$) & 3.82E+03 ($>$) \\
    & Std  & \textbf{1.53E+01} & 3.24E+08 & 8.38E+01 & 1.25E+07 & 5.73E+06 & 1.00E+06 & 8.35E+03 & 3.10E+06 \\
F12 & Mean & \textbf{1.72E+06} & 1.89E+08 ($>$) & 1.21E+07 ($>$) & 4.66E+09 ($>$) & 5.76E+10 ($>$) & 1.04E+10 ($>$) & 8.00E+06 ($>$) & 6.65E+08 ($>$) \\
    & Std  & \textbf{7.03E+05} & 2.27E+16 & 5.98E+06 & 1.95E+19 & 1.06E+20 & 6.00E+18 & 2.30E+13 & 9.23E+17 \\
F13 & Mean & \textbf{1.49E+04} & 7.25E+06 ($>$) & 1.38E+05 ($>$) & 3.86E+08 ($>$) & 4.75E+09 ($>$) & 2.52E+09 ($>$) & 5.36E+04 ($>$) & 1.12E+08 ($>$) \\
    & Std  & \textbf{3.71E+03} & 2.08E+14 & 6.30E+04 & 3.87E+17 & 1.91E+19 & 5.58E+17 & 4.32E+08 & 2.38E+16 \\
F14 & Mean & \textbf{2.24E+04} & 1.93E+07 ($>$) & 8.12E+04 ($>$) & 4.21E+06 ($>$) & 6.46E+05 ($>$) & 2.08E+06 ($>$) & 9.81E+04 ($>$) & 6.29E+05 ($>$) \\
    & Std  & \textbf{1.04E+04} & 2.29E+14 & 5.71E+04 & 1.08E+13 & 4.93E+11 & 1.05E+12 & 4.14E+09 & 7.05E+11 \\
F15 & Mean & \textbf{1.66E+04} & 4.10E+06 ($>$) & 6.90E+04 ($>$) & 7.30E+07 ($>$) & 3.25E+04 ($>$) & 3.04E+08 ($>$) & 1.84E+04 ($>$) & 5.25E+06 ($>$) \\
    & Std  & \textbf{3.54E+03} & 8.64E+13 & 2.58E+04 & 5.29E+16 & 7.84E+07 & 2.03E+16 & 5.43E+07 & 1.42E+14 \\
F16 & Mean & \textbf{2.57E+03} & 4.69E+03 ($>$) & 3.54E+03 ($>$) & 4.63E+03 ($>$) & 6.28E+03 ($>$) & 5.43E+03 ($>$) & 3.13E+03 ($>$) & 2.92E+03 ($>$) \\
    & Std  & \textbf{2.47E+02} & 2.97E+05 & 4.25E+02 & 5.91E+05 & 1.30E+06 & 1.15E+05 & 2.25E+05 & 1.23E+05 \\
F17 & Mean & \textbf{2.59E+03} & 3.86E+03 ($>$) & 3.22E+03 ($>$) & 3.99E+03 ($>$) & 4.07E+03 ($>$) & 4.27E+03 ($>$) & 3.10E+03 ($>$) & 2.70E+03 ($>$) \\
    & Std  & \textbf{2.02E+02} & 2.10E+05 & 2.98E+02 & 6.49E+05 & 1.89E+05 & 9.11E+04 & 1.27E+05 & 5.42E+04 \\
F18 & Mean & \textbf{1.42E+05} & 2.99E+07 ($>$) & 4.53E+05 ($>$) & 2.77E+07 ($>$) & 1.45E+07 ($>$) & 1.31E+07 ($>$) & 1.19E+06 ($>$) & 2.92E+06 ($>$) \\
    & Std  & \textbf{2.65E+04} & 4.03E+14 & 1.99E+05 & 7.00E+14 & 1.51E+14 & 4.53E+13 & 7.12E+11 & 1.51E+13 \\
F19 & Mean & \textbf{1.84E+04} & 5.60E+06 ($>$) & 2.93E+04 ($>$) & 6.69E+06 ($>$) & 4.69E+05 ($>$) & 2.43E+08 ($>$) & 1.96E+04 ($\approx$) & 5.99E+06 ($>$) \\
    & Std  & \textbf{4.45E+03} & 1.43E+13 & 1.62E+04 & 7.86E+14 & 1.54E+08 & 8.49E+15 & 1.90E+08 & 5.13E+14 \\
F20 & Mean & \textbf{2.22E+03} & 2.99E+03 ($>$) & 2.43E+03 ($>$) & 2.88E+03 ($>$) & 2.73E+03 ($>$) & 2.63E+03 ($>$) & 2.40E+03 ($>$) & 2.35E+03 ($>$) \\
    & Std  & \textbf{4.64E+01} & 2.57E+02 & 1.41E+02 & 2.27E+02 & 1.77E+02 & 1.22E+02 & 1.79E+02 & 1.31E+04 \\
F21 & Mean & \textbf{2.39E+03} & 3.15E+03 ($>$) & 2.60E+03 ($>$) & 2.86E+03 ($>$) & 3.03E+03 ($>$) & 2.86E+03 ($>$) & 2.65E+03 ($>$) & 2.49E+03 ($>$) \\
    & Std  & \textbf{2.07E+01} & 7.55E+03 & 5.10E+01 & 6.61E+03 & 5.42E+03 & 1.47E+03 & 6.81E+03 & 1.03E+03 \\
F22 & Mean & \textbf{3.36E+03} & 1.24E+04 ($>$) & 8.72E+03 ($>$) & 1.20E+04 ($>$) & 1.52E+04 ($>$) & 1.59E+04 ($>$) & 1.06E+04 ($>$) & 8.38E+03 ($>$) \\
    & Std  & 2.33E+03 & 2.29E+06 & \textbf{9.58E+02} & 1.27E+06 & 5.44E+05 & 1.61E+05 & 6.28E+06 & 8.64E+05 \\
F23 & Mean & \textbf{2.82E+03} & 3.77E+03 ($>$) & 3.02E+03 ($>$) & 3.90E+03 ($>$) & 4.32E+03 ($>$) & 3.51E+03 ($>$) & 3.16E+03 ($>$) & 2.93E+03 ($>$) \\
    & Std  & \textbf{2.49E+01} & 1.73E+04 & 5.92E+01 & 5.44E+04 & 4.26E+04 & 3.29E+03 & 1.10E+04 & 1.46E+03 \\
F24 & Mean & \textbf{2.98E+03} & 4.08E+03 ($>$) & 3.22E+03 ($>$) & 3.99E+03 ($>$) & 4.89E+03 ($>$) & 3.66E+03 ($>$) & 3.25E+03 ($>$) & 3.11E+03 ($>$) \\
    & Std  & \textbf{1.87E+01} & 2.55E+04 & 9.41E+01 & 3.47E+04 & 5.36E+04 & 2.13E+03 & 9.70E+03 & 5.92E+03 \\
F25 & Mean & 3.06E+03 & 4.44E+03 ($>$) & \textbf{3.03E+03 ($<$)} & 4.35E+03 ($>$) & 1.39E+04 ($>$) & 5.93E+03 ($>$) & 3.08E+03 ($>$) & 3.49E+03 ($>$) \\
    & Std  & \textbf{2.10E+01} & 3.89E+05 & 3.84E+01 & 2.41E+05 & 1.71E+06 & 3.70E+05 & 1.30E+03 & 6.14E+04 \\
F26 & Mean & \textbf{3.52E+03} & 1.00E+04 ($>$) & 6.68E+03 ($>$) & 1.27E+04 ($>$) & 1.58E+04 ($>$) & 1.16E+04 ($>$) & 9.97E+03 ($>$) & 6.11E+03 ($>$) \\
    & Std  & 1.27E+03 & 1.19E+06 & \textbf{6.61E+02} & 4.18E+06 & 1.02E+06 & 3.85E+05 & 6.88E+06 & 3.64E+05 \\
F27 & Mean & \textbf{3.31E+03} & 4.67E+03 ($>$) & 3.44E+03 ($>$) & 4.46E+03 ($>$) & 6.61E+03 ($>$) & 4.38E+03 ($>$) & 3.67E+03 ($>$) & 3.54E+03 ($>$) \\
    & Std  & \textbf{3.26E+01} & 6.36E+04 & 8.78E+01 & 1.37E+05 & 3.29E+05 & 1.41E+04 & 3.14E+04 & 6.81E+03 \\
F28 & Mean & 3.30E+03 & 5.90E+03 ($>$) & \textbf{3.29E+03 ($<$)} & 5.53E+03 ($>$) & 1.09E+04 ($>$) & 6.40E+03 ($>$) & 3.33E+03 ($>$) & 4.12E+03 ($>$) \\
    & Std  & \textbf{1.06E+01} & 7.53E+05 & 2.37E+01 & 4.59E+05 & 8.03E+05 & 2.34E+05 & 1.10E+03 & 1.18E+05 \\
F29 & Mean & \textbf{3.70E+03} & 5.77E+03 ($>$) & 4.24E+03 ($>$) & 6.36E+03 ($>$) & 1.83E+04 ($>$) & 7.00E+03 ($>$) & 4.84E+03 ($>$) & 4.26E+03 ($>$) \\
    & Std  & \textbf{2.14E+02} & 2.15E+05 & 2.95E+02 & 6.69E+05 & 3.34E+07 & 2.71E+05 & 2.43E+05 & 7.28E+04 \\
F30 & Mean & \textbf{9.67E+05} & 1.87E+08 ($>$) & 1.68E+06 ($>$) & 1.05E+08 ($>$) & 5.88E+08 ($>$) & 5.62E+08 ($>$) & 2.21E+06 ($>$) & 7.62E+07 ($>$) \\
    & Std  & \textbf{7.94E+04} & 8.83E+15 & 5.43E+05 & 5.12E+15 & 1.40E+18 & 3.22E+16 & 8.91E+11 & 6.89E+14 \\ \midrule
Wilcoxon & $+/\approx/-$ & 0/29/0 & 29/0/0 & 26/0/3 & 29/0/0 & 29/0/0 & 29/0/0 & 28/1/0 & 29/0/0 \\ \midrule
FM-rank & & \textbf{1.1034} & 6.0690 & 2.7586 & 6.1379 & 7.0345 & 6.3793 & 3.3103 & 3.2069 \\
M-rank  & & \textbf{1} & 5 & 2 & 6 & 8 & 7 & 4 & 3 \\ \bottomrule
\end{tabularx}
\end{table*}

\begin{table*}[!htbp]
\centering
\caption{The results of BWE and its competitors in CEC2017 100D using Wilcoxon rank-sum test and Friedman rank test.}
\label{tablecec2017-100D}
\sffamily
\scriptsize
\setlength{\tabcolsep}{2pt}
\begin{tabularx}{\textwidth}{@{}llXXXXXXXX@{}}
\toprule
No. & & BWE & GA & LEA & HEOA & AOA & SCA & COA & GWO \\ \midrule
F1  & Mean & \textbf{3.54E+03} & 3.12E+11 ($>$) & 3.43E+04 ($>$) & 6.46E+10 ($>$) & 2.62E+11 ($>$) & 1.54E+11 ($>$) & 3.16E+08 ($>$) & 3.80E+10 ($>$) \\
    & Std  & \textbf{3.46E+03} & 5.63E+10       & 2.37E+04       & 1.32E+10       & 1.07E+10       & 9.18E+09       & 8.37E+08       & 6.28E+19       \\
F3  & Mean & 4.81E+03          & 9.29E+05 ($>$) & \textbf{3.00E+02 ($<$)} & 5.96E+05 ($>$) & 3.28E+05 ($>$) & 2.82E+05 ($>$) & 3.23E+05 ($>$) & 2.07E+05 ($>$) \\
    & Std  & 1.41E+03          & 1.32E+05       & \textbf{2.18E-02}       & 1.47E+05       & 1.81E+04       & 2.26E+04       & 5.90E+04       & 5.74E+08       \\
F4  & Mean & \textbf{6.24E+02} & 3.81E+04 ($>$) & 6.52E+02 ($>$) & 7.00E+03 ($>$) & 7.60E+04 ($>$) & 2.53E+04 ($>$) & 9.10E+02 ($>$) & 3.82E+03 ($>$) \\
    & Std  & \textbf{3.83E+01} & 1.34E+04       & 3.29E+01       & 2.55E+03       & 1.09E+04       & 3.65E+03       & 1.31E+02       & 9.09E+05       \\
F5  & Mean & \textbf{7.02E+02} & 2.57E+03 ($>$) & 1.36E+03 ($>$) & 1.69E+03 ($>$) & 1.97E+03 ($>$) & 1.86E+03 ($>$) & 1.35E+03 ($>$) & 1.09E+03 ($>$) \\
    & Std  & \textbf{3.24E+01} & 2.11E+02       & 1.31E+02       & 8.68E+01       & 5.31E+01       & 4.49E+01       & 3.41E+01       & 4.46E+03       \\
F6  & Mean & \textbf{6.00E+02} & 7.55E+02 ($>$) & 6.66E+02 ($>$) & 6.79E+02 ($>$) & 7.01E+02 ($>$) & 6.89E+02 ($>$) & 6.61E+02 ($>$) & 6.31E+02 ($>$) \\
    & Std  & \textbf{1.83E-01} & 6.30E+00       & 8.78E+00       & 4.49E+00       & 3.69E+00       & 3.79E+00       & 4.89E+00       & 1.51E+01       \\
F7  & Mean & \textbf{9.55E+02} & 8.88E+03 ($>$) & 2.03E+03 ($>$) & 4.91E+03 ($>$) & 3.75E+03 ($>$) & 3.39E+03 ($>$) & 3.13E+03 ($>$) & 1.83E+03 ($>$) \\
    & Std  & \textbf{2.90E+01} & 1.01E+03       & 1.82E+02       & 4.64E+02       & 8.61E+01       & 1.30E+02       & 2.24E+02       & 1.35E+04       \\
F8  & Mean & \textbf{9.99E+02} & 2.91E+03 ($>$) & 1.61E+03 ($>$) & 2.30E+03 ($>$) & 2.43E+03 ($>$) & 2.21E+03 ($>$) & 1.84E+03 ($>$) & 1.39E+03 ($>$) \\
    & Std  & \textbf{3.38E+01} & 2.16E+02       & 1.43E+02       & 1.12E+02       & 5.98E+01       & 5.76E+01       & 6.47E+01       & 5.50E+03       \\
F9  & Mean & \textbf{1.35E+03} & 8.55E+04 ($>$) & 7.51E+04 ($>$) & 4.32E+04 ($>$) & 5.41E+04 ($>$) & 6.80E+04 ($>$) & 2.74E+04 ($>$) & 2.37E+04 ($>$) \\
    & Std  & \textbf{3.48E+02} & 1.46E+04       & 1.69E+04       & 5.55E+03       & 4.87E+03       & 6.33E+03       & 5.76E+03       & 9.77E+07       \\
F10 & Mean & \textbf{1.30E+04} & 2.76E+04 ($>$) & 1.56E+04 ($>$) & 2.09E+04 ($>$) & 2.80E+04 ($>$) & 3.12E+04 ($>$) & 1.73E+04 ($>$) & 1.40E+04 ($>$) \\
    & Std  & 1.12E+03          & 1.60E+03       & 1.36E+03       & 1.83E+03       & 1.13E+03       & \textbf{6.12E+02} & 1.64E+03       & 2.47E+06       \\
F11 & Mean & \textbf{1.75E+03} & 2.63E+05 ($>$) & 2.58E+03 ($>$) & 1.16E+05 ($>$) & 1.65E+05 ($>$) & 7.00E+04 ($>$) & 3.05E+03 ($>$) & 4.32E+04 ($>$) \\
    & Std  & \textbf{8.53E+01} & 7.55E+04       & 2.39E+02       & 2.47E+04       & 2.37E+04       & 1.16E+04       & 3.33E+02       & 1.47E+08       \\
F12 & Mean & \textbf{6.40E+06} & 6.83E+10 ($>$) & 4.06E+07 ($>$) & 1.39E+10 ($>$) & 1.77E+11 ($>$) & 5.52E+10 ($>$) & 7.40E+07 ($>$) & 6.46E+09 ($>$) \\
    & Std  & 2.16E+06          & 2.83E+10       & \textbf{1.52E+07} & 8.01E+09       & 1.80E+10       & 6.68E+09       & 7.40E+07       & 2.16E+19       \\
F13 & Mean & \textbf{1.76E+04} & 8.79E+09 ($>$) & 1.19E+05 ($>$) & 1.49E+09 ($>$) & 3.71E+10 ($>$) & 8.09E+09 ($>$) & 9.65E+04 ($>$) & 5.59E+08 ($>$) \\
    & Std  & \textbf{3.30E+03} & 6.70E+09       & 4.00E+04       & 1.48E+09       & 5.21E+09       & 1.40E+09       & 5.35E+04       & 2.45E+17       \\
F14 & Mean & \textbf{1.57E+05} & 9.68E+07 ($>$) & 4.71E+05 ($>$) & 1.61E+07 ($>$) & 1.97E+07 ($>$) & 1.84E+07 ($>$) & 6.30E+05 ($>$) & 5.05E+06 ($>$) \\
    & Std  & \textbf{3.93E+04} & 4.42E+07       & 1.99E+05       & 6.41E+06       & 1.17E+07       & 6.88E+06       & 3.93E+05       & 8.78E+12       \\
F15 & Mean & \textbf{8.13E+03} & 4.61E+08 ($>$) & 1.03E+05 ($>$) & 1.56E+08 ($>$) & 5.29E+09 ($>$) & 2.60E+09 ($>$) & 4.02E+04 ($>$) & 1.10E+08 ($>$) \\
    & Std  & \textbf{1.72E+03} & 5.54E+08       & 2.85E+04       & 2.18E+08       & 2.53E+09       & 6.81E+08       & 2.13E+04       & 4.62E+16       \\
F16 & Mean & \textbf{4.53E+03} & 1.22E+04 ($>$) & 5.76E+03 ($>$) & 9.16E+03 ($>$) & 1.92E+04 ($>$) & 1.24E+04 ($>$) & 5.73E+03 ($>$) & 5.75E+03 ($>$) \\
    & Std  & \textbf{4.55E+02} & 1.79E+03       & 7.60E+02       & 1.16E+03       & 2.60E+03       & 5.66E+02       & 8.05E+02       & 5.90E+05       \\
F17 & Mean & \textbf{3.97E+03} & 1.09E+04 ($>$) & 5.39E+03 ($>$) & 1.78E+04 ($>$) & 3.73E+05 ($>$) & 1.10E+04 ($>$) & 5.75E+03 ($>$) & 4.91E+03 ($>$) \\
    & Std  & \textbf{4.35E+02} & 5.93E+03       & 5.98E+02       & 3.09E+04       & 5.12E+05       & 1.39E+03       & 6.18E+02       & 5.96E+06       \\
F18 & Mean & \textbf{2.91E+05} & 7.59E+07 ($>$) & 9.38E+05 ($>$) & 1.44E+07 ($>$) & 3.92E+07 ($>$) & 3.15E+07 ($>$) & 1.02E+06 ($>$) & 4.09E+06 ($>$) \\
    & Std  & \textbf{5.64E+04} & 5.70E+07       & 3.61E+05       & 7.91E+06       & 3.74E+07       & 1.04E+07       & 6.55E+05       & 7.44E+12       \\
F19 & Mean & \textbf{4.35E+03} & 1.28E+09 ($>$) & 1.02E+05 ($>$) & 3.59E+08 ($>$) & 4.94E+09 ($>$) & 2.08E+09 ($>$) & 2.71E+04 ($>$) & 7.36E+07 ($>$) \\
    & Std  & \textbf{1.15E+03} & 2.08E+09       & 3.78E+04       & 8.17E+08       & 2.54E+09       & 6.86E+08       & 1.84E+04       & 7.94E+15       \\
F20 & Mean & \textbf{4.28E+03} & 7.61E+03 ($>$) & 4.99E+03 ($>$) & 6.06E+03 ($>$) & 5.71E+03 ($>$) & 7.14E+03 ($>$) & 5.66E+03 ($>$) & 4.35E+03 ($\approx$) \\
    & Std  & 3.79E+02          & 6.84E+02       & 5.24E+02       & 5.13E+02       & 5.30E+02       & \textbf{2.32E+02} & 3.64E+02       & 4.72E+05       \\
F21 & Mean & \textbf{2.52E+03} & 4.86E+03 ($>$) & 3.21E+03 ($>$) & 3.78E+03 ($>$) & 4.55E+03 ($>$) & 3.88E+03 ($>$) & 3.41E+03 ($>$) & 2.89E+03 ($>$) \\
    & Std  & \textbf{3.59E+01} & 2.24E+02       & 1.16E+02       & 1.38E+02       & 2.00E+02       & 5.75E+01       & 1.69E+02       & 4.42E+03       \\
F22 & Mean & \textbf{8.63E+03} & 3.09E+04 ($>$) & 1.83E+04 ($>$) & 2.40E+04 ($>$) & 3.12E+04 ($>$) & 3.35E+04 ($>$) & 2.13E+04 ($>$) & 1.71E+04 ($>$) \\
    & Std  & 6.83E+03          & 1.66E+03       & 1.60E+03       & 1.67E+03       & 1.03E+03       & \textbf{6.25E+02} & 4.85E+03       & 2.32E+06       \\
F23 & Mean & \textbf{3.04E+03} & 5.53E+03 ($>$) & 3.54E+03 ($>$) & 4.75E+03 ($>$) & 6.91E+03 ($>$) & 4.73E+03 ($>$) & 3.99E+03 ($>$) & 3.47E+03 ($>$) \\
    & Std  & \textbf{3.64E+01} & 3.00E+02       & 8.65E+01       & 2.35E+02       & 5.04E+02       & 8.49E+01       & 2.29E+02       & 5.25E+03       \\
F24 & Mean & \textbf{3.43E+03} & 7.52E+03 ($>$) & 4.15E+03 ($>$) & 5.81E+03 ($>$) & 1.14E+04 ($>$) & 6.29E+03 ($>$) & 4.78E+03 ($>$) & 4.03E+03 ($>$) \\
    & Std  & \textbf{3.69E+01} & 7.35E+02       & 1.48E+02       & 3.63E+02       & 8.24E+02       & 1.73E+02       & 3.32E+02       & 1.28E+04       \\
F25 & Mean & \textbf{3.28E+03} & 2.88E+04 ($>$) & 3.26E+03 ($<$) & 7.53E+03 ($>$) & 2.34E+04 ($>$) & 1.40E+04 ($>$) & 3.60E+03 ($>$) & 5.74E+03 ($>$) \\
    & Std  & \textbf{4.41E+01} & 6.37E+03       & 6.96E+01       & 1.05E+03       & 1.69E+03       & 1.21E+03       & 1.06E+02       & 3.98E+05       \\
F26 & Mean & \textbf{6.46E+03} & 3.67E+04 ($>$) & 1.41E+04 ($>$) & 3.34E+04 ($>$) & 5.10E+04 ($>$) & 3.16E+04 ($>$) & 2.54E+04 ($>$) & 1.36E+04 ($>$) \\
    & Std  & 3.86E+03          & 3.83E+03       & 1.45E+03       & 4.77E+03       & 3.06E+03       & \textbf{1.42E+03} & 4.85E+03       & 9.83E+05       \\
F27 & Mean & \textbf{3.41E+03} & 6.43E+03 ($>$) & 3.52E+03 ($>$) & 5.25E+03 ($>$) & 1.27E+04 ($>$) & 6.71E+03 ($>$) & 3.98E+03 ($>$) & 3.91E+03 ($>$) \\
    & Std  & \textbf{2.37E+01} & 8.72E+02       & 7.81E+01       & 6.70E+02       & 1.23E+03       & 3.11E+02       & 2.35E+02       & 1.77E+04       \\
F28 & Mean & \textbf{3.38E+03} & 2.78E+04 ($>$) & 3.38E+03 ($\approx$) & 1.04E+04 ($>$) & 3.00E+04 ($>$) & 1.81E+04 ($>$) & 3.69E+03 ($>$) & 7.21E+03 ($>$) \\
    & Std  & \textbf{3.22E+01} & 8.37E+03       & 4.49E+01       & 1.80E+03       & 1.76E+03       & 1.34E+03       & 1.39E+02       & 9.21E+05       \\
F29 & Mean & \textbf{5.70E+03} & 1.87E+04 ($>$) & 6.59E+03 ($>$) & 1.17E+04 ($>$) & 9.98E+04 ($>$) & 1.56E+04 ($>$) & 8.17E+03 ($>$) & 7.74E+03 ($>$) \\
    & Std  & \textbf{4.23E+02} & 9.12E+03       & 5.30E+02       & 1.61E+03       & 6.07E+04       & 1.30E+03       & 6.87E+02       & 3.19E+05       \\
F30 & Mean & \textbf{4.68E+04} & 5.34E+09 ($>$) & 7.50E+05 ($>$) & 1.26E+09 ($>$) & 3.36E+10 ($>$) & 5.80E+09 ($>$) & 1.96E+06 ($>$) & 6.56E+08 ($>$) \\
    & Std  & \textbf{1.05E+04} & 3.55E+09       & 1.99E+05       & 1.81E+09       & 6.91E+09       & 8.69E+08       & 1.86E+06       & 4.12E+17       \\ \midrule
Wilcoxon & $+/\approx/-$ & 0/29/0 & 29/0/0 & 26/1/2 & 29/0/0 & 29/0/0 & 29/0/0 & 29/0/0 & 28/1/0 \\ \midrule
FM-rank & & \textbf{1.1034} & 7.1379 & 2.7241 & 5.3448 & 7.2759 & 6.1034 & 3.3448 & 2.9655 \\
M-rank  & & \textbf{1} & 7 & 2 & 5 & 8 & 6 & 4 & 3 \\ \bottomrule
\end{tabularx}
\end{table*}

\subsubsection{Capabilities of exploration and exploitation}
Unimodal functions possess a single global optimum, making them ideal for evaluating an algorithm’s exploitation capability.
In contrast, multimodal functions (F4$\sim$F10) contain numerous local optima, primarily reflecting an algorithm’s exploration capability.
To provide a rigorous quantitative assessment, \cref{tablecec2017-10D,tablecec2017-30D,tablecec2017-50D,tablecec2017-100D} present Mean and Std values for BWE and its competitors.

On unimodal functions, BWE exhibits a commanding advantage.
For F1, it outperforms the second-best algorithm by several orders of magnitude across all dimensions.
While optimization accuracy naturally degrades with increasing dimensionality, BWE scales gracefully compared to peers like COA and GA, maintaining a significantly lower error margin.
On F3, BWE achieves near-optimal values (300) from $10D\sim50D$, remaining the top-ranked algorithm.

For multimodal, BWE demonstrates exceptional exploration capabilities and local optima avoidance, achieving the best Mean values across almost all dimensions, with only minor exceptions (e.g., F10 at $D=10$ and F4 at $D=30$).
Its competitive edge becomes more pronounced as dimensionality increases to $50D$ and $100D$, suggesting the Bézier search effectively manages the increased complexity where algorithms like SCA and AOA fail.
BWE also consistently ranks first in Std for the majority of functions, indicating robustness against stochasticity. Overall, it demonstrates an excellent balance between exploration and exploitation.

\subsubsection{Capability of avoiding local optima}
Hybrid and composition functions construct complex, multimodal landscapes by integrating or superimposing multiple basic functions, thereby providing a rigorous testbed for evaluating an algorithm's global search capability.

As detailed in \cref{tablecec2017-10D}, BWE exhibits slightly subpar performance on hybrid functions at $D=10$, particularly underperforming on F14 and F19.
Nevertheless, its ranking steadily improves with increasing dimensionality.
As shown in \cref{tablecec2017-30D} and \cref{tablecec2017-50D}, BWE ranks first in both Mean and Std across all hybrid functions at $D=30$ and $D=50$.
By $D=100$ (\cref{tablecec2017-100D}), BWE dominates the Mean for all hybrid functions, yielding top Std rankings except for F12 and F20.

Similarly, for composition functions, BWE secures the best Mean performance across all test cases from $D=30$ to $D=100$ (\cref{tablecec2017-30D,tablecec2017-50D,tablecec2017-100D}), with the sole exception of F25 at $D=50$.
While competitors such as SCA, COA, and AOA occasionally achieve marginally better Std values in lower dimensions (\cref{tablecec2017-10D}), BWE consistently enhances its ranking as dimensionality scales up, underscoring its superior suitability for high-dimensional problem spaces.

In summary, the experimental results demonstrate BWE's outstanding capability to escape local optima and sustain high convergence accuracy across complex landscapes.
Although there is minor room for improving stability in specific low-dimensional structures, BWE's exceptional scalability and global search precision render it a highly effective optimizer for complex, high-dimensional problems.

\subsubsection{Statistical tests}
Since the optimization process of metaheuristic algorithms is stochastic, merely comparing data and charts is insufficient to convincingly demonstrate an algorithm's superiority.
Therefore, we conducted statistical analyses on all algorithms.
Firstly, we performed Wilcoxon rank-sum tests \citep{gao2025freedom} at a significance level of 0.05.
\cref{tablecec2017-10D,tablecec2017-30D,tablecec2017-50D,tablecec2017-100D} illustrate that BWE significantly outperforms popular algorithms in almost all results.
Among these, LEA, COA, and SCA demonstrate competitive performance on certain functions across $10D$, but their overall performance falls short of the proposed method.

Additionally, \cref{tablecec2017-10D,tablecec2017-30D,tablecec2017-50D,tablecec2017-100D} also present the Friedman rank test \citep{ouyang2024escape} results for all algorithms. In \cref{tablecec2017-10D,tablecec2017-30D,tablecec2017-50D,tablecec2017-100D}, we use "FM-rank" to denote Friedman's mean rank and "M-rank" to denote the mean rank.
From the results, BWE ranks 1st in all dimensions.
The FM-rank result is 1.6552 on $10D$. However, as the number of dimensions increases, the FM-rank results for BWE clearly increase.
On the $30D\sim100D$, BWE's FM-rank is 1.1034 across the board, nearly identical to 1.
This indicates that BWE ranks first in virtually all functions.

In summary, both statistical tests confirm that BWE maintains significant superiority, indicating that the algorithm's excellent optimization performance is not random.

\begin{figure*}[!htbp]
    \centering

    \includegraphics[width=0.7\textwidth]{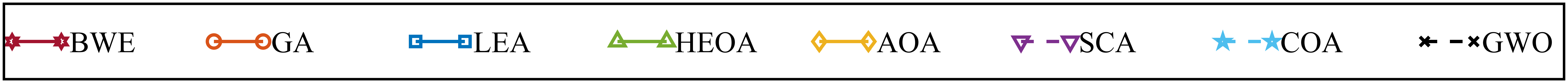}
    \vspace{0.1cm} 
    
    \includegraphics[width=0.25\textwidth]{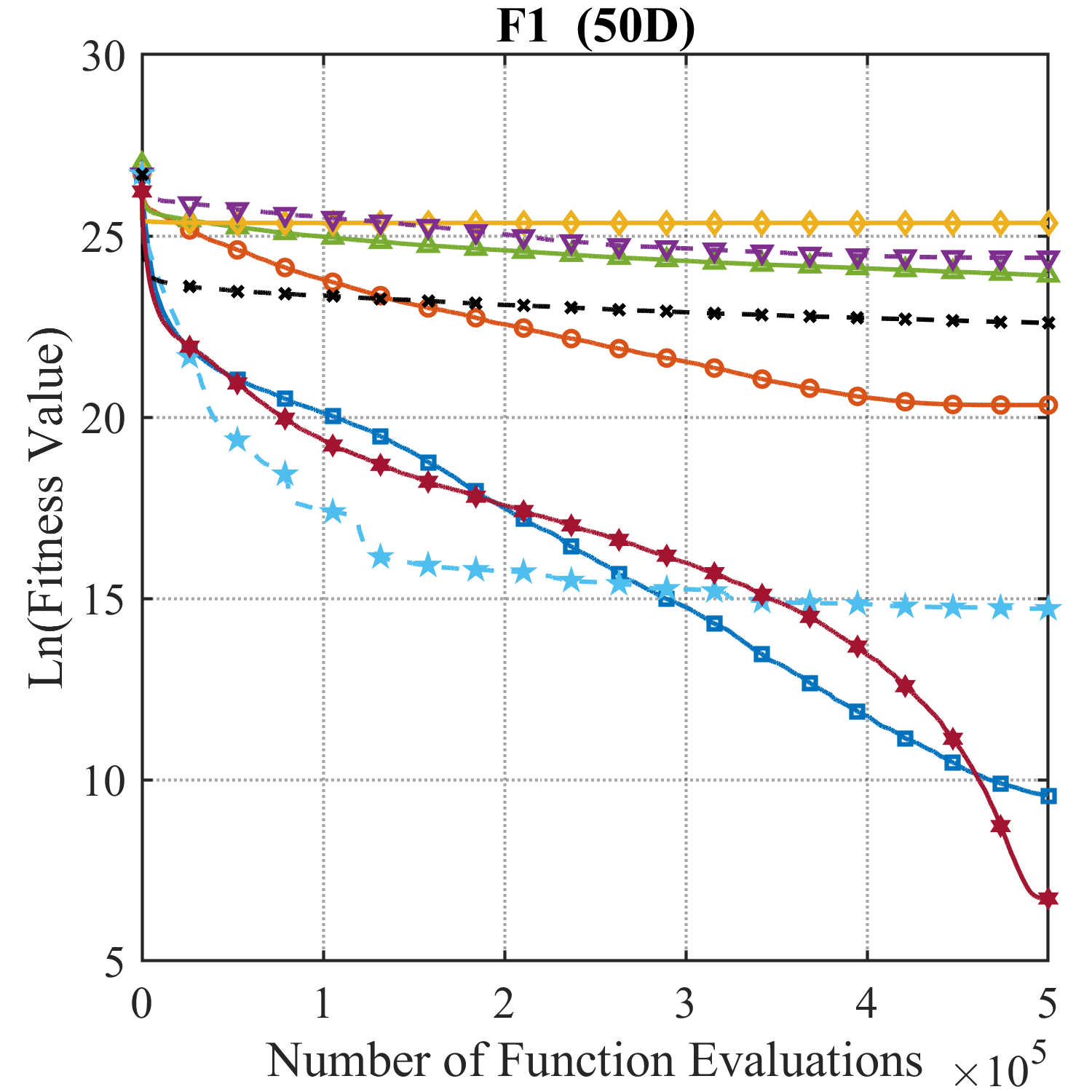}%
    \includegraphics[width=0.25\textwidth]{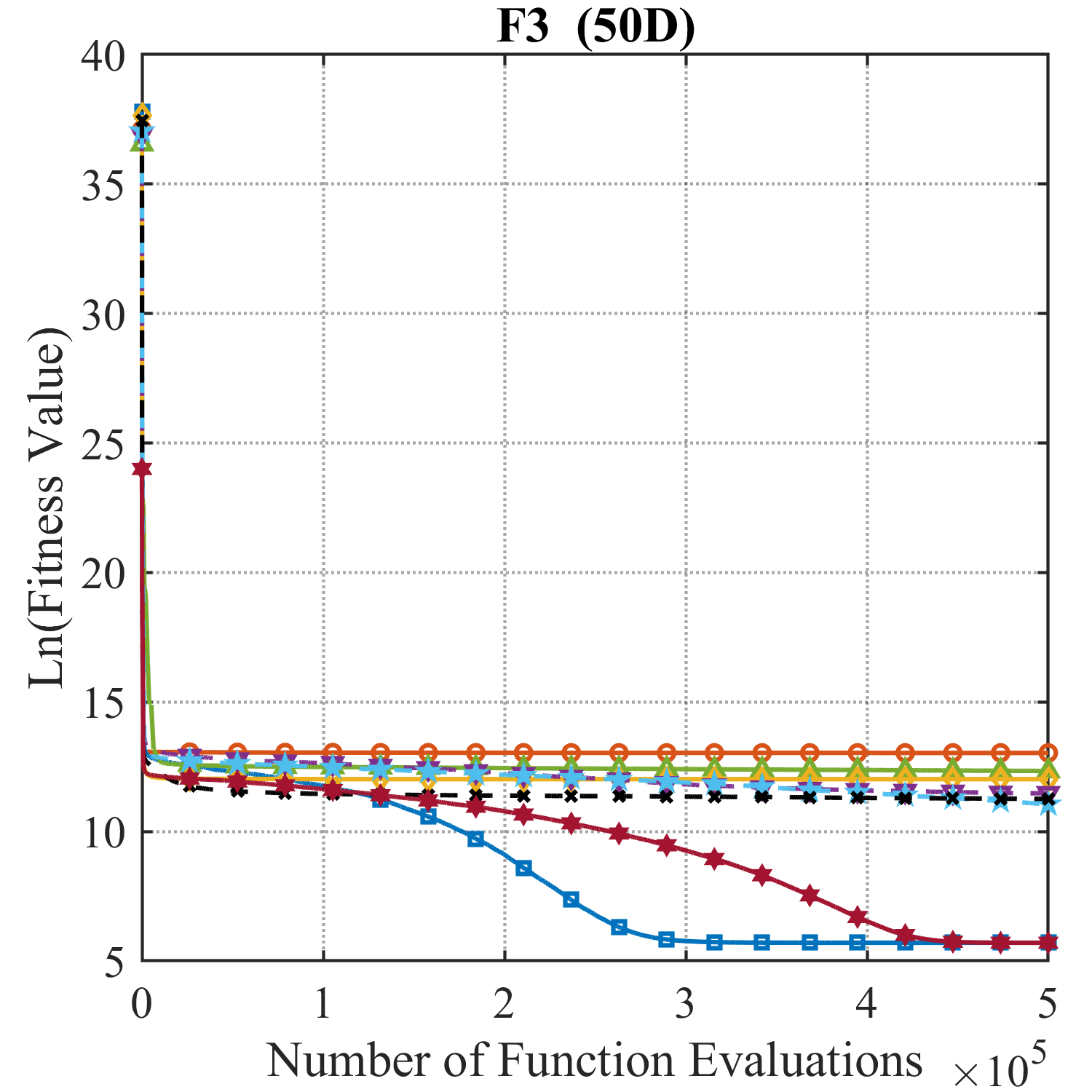}%
    \includegraphics[width=0.25\textwidth]{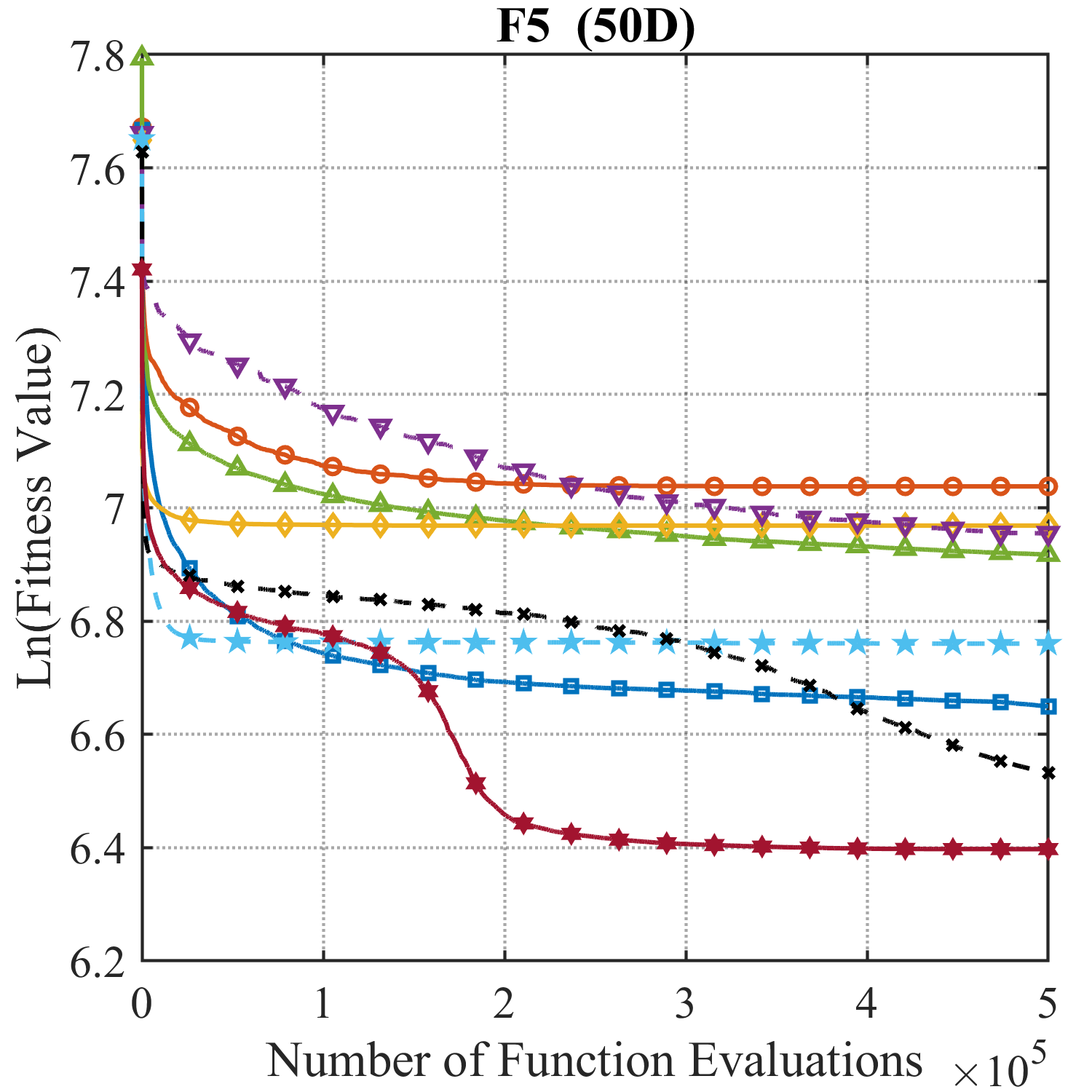}%
    \includegraphics[width=0.25\textwidth]{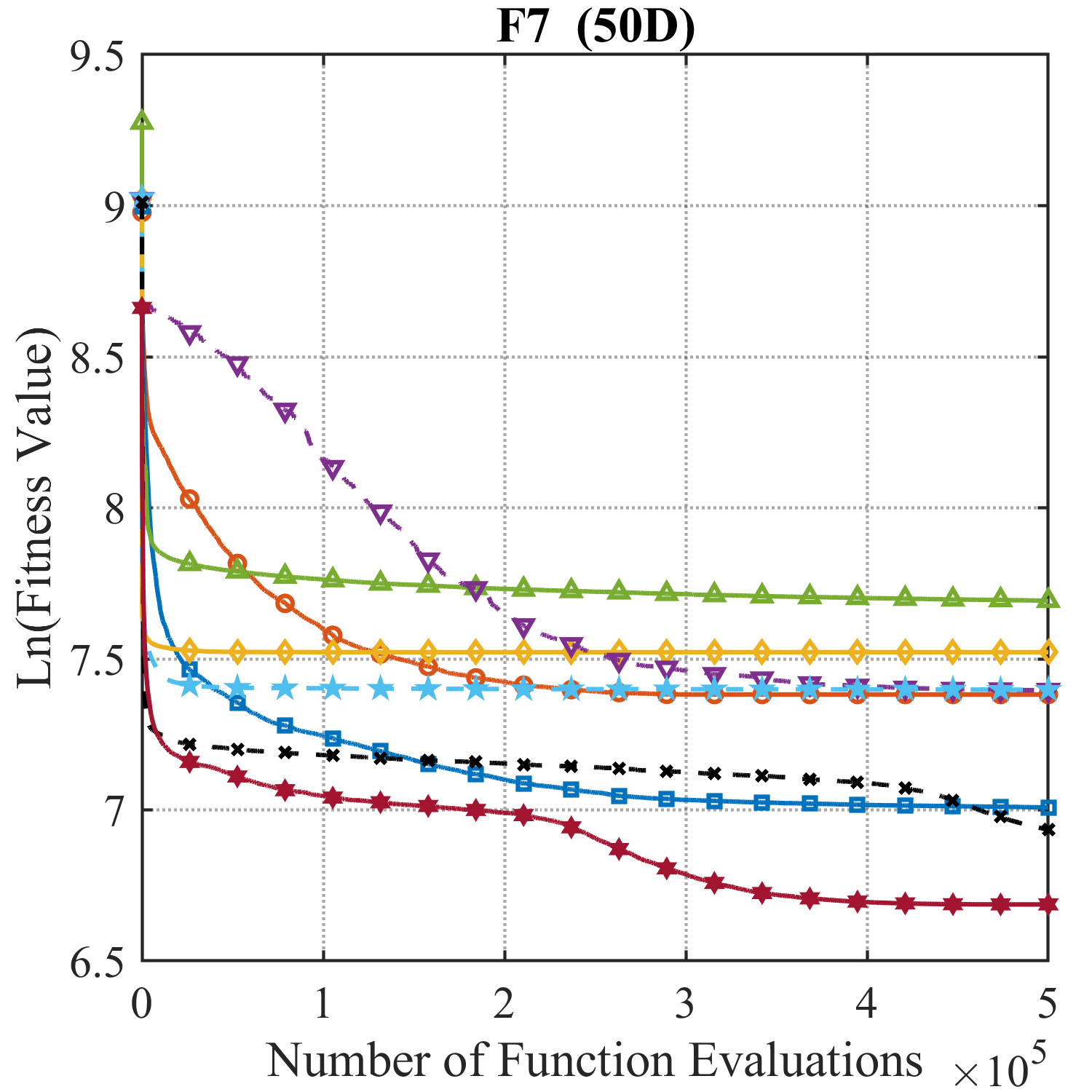}
    
    \includegraphics[width=0.25\textwidth]{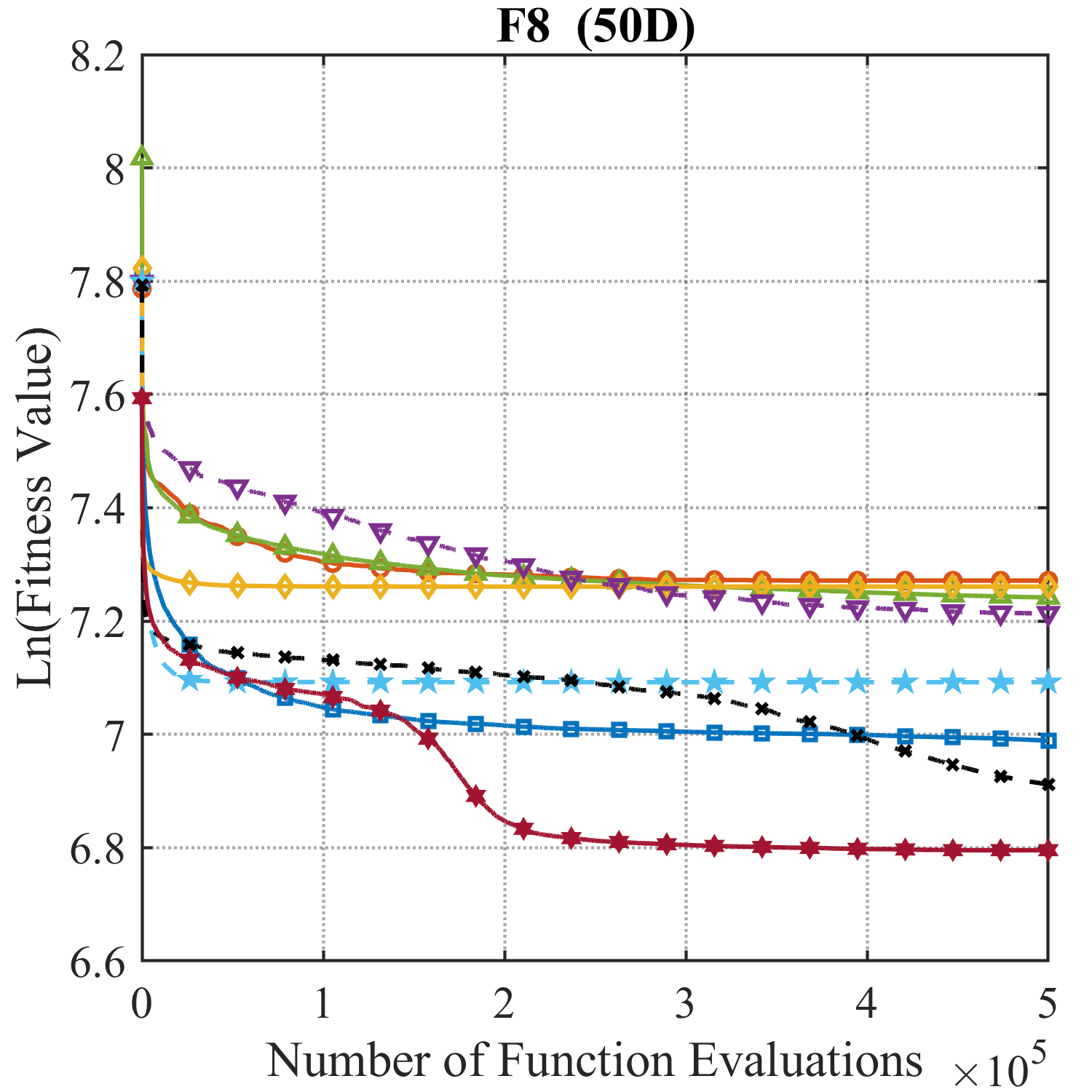}%
    \includegraphics[width=0.25\textwidth]{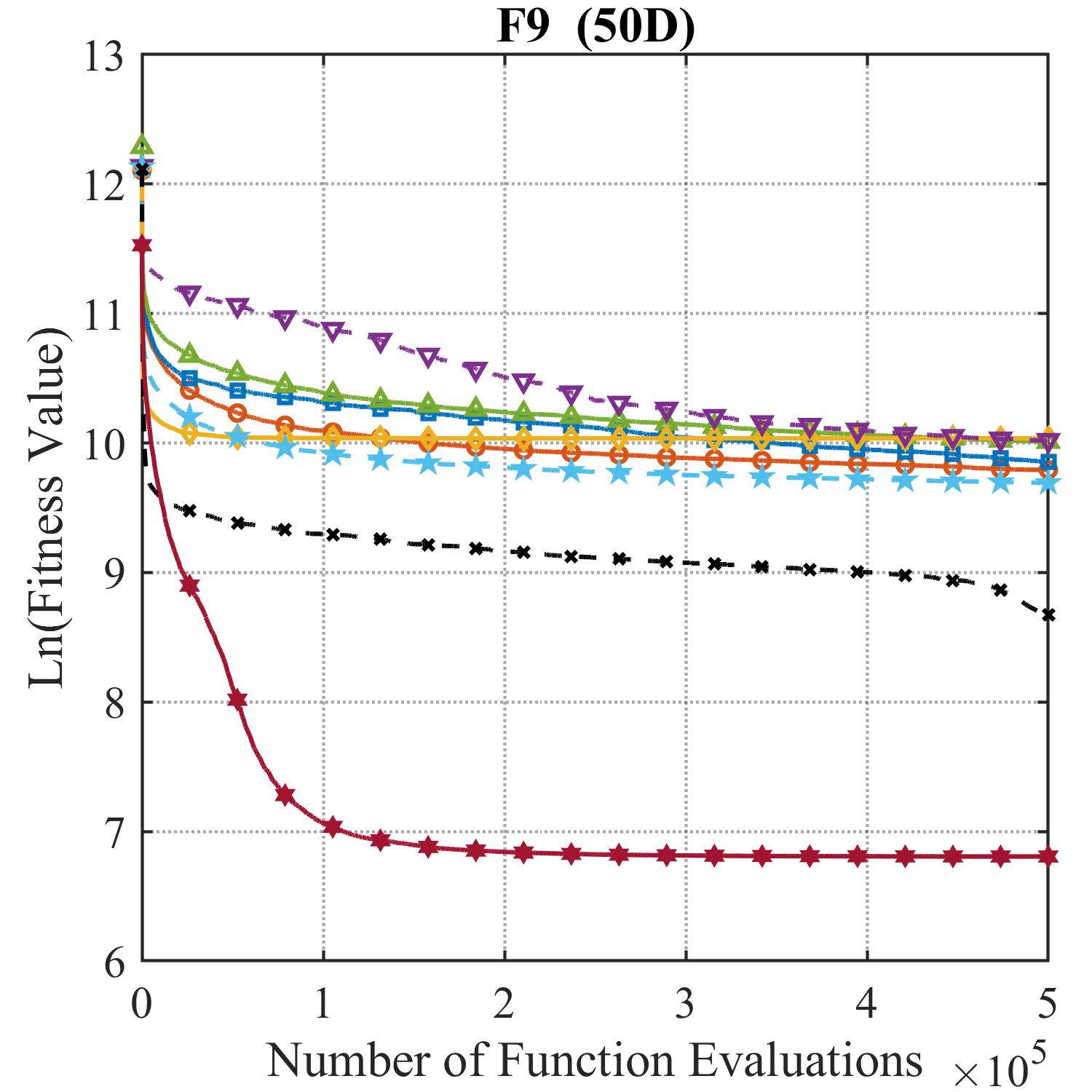}%
    \includegraphics[width=0.25\textwidth]{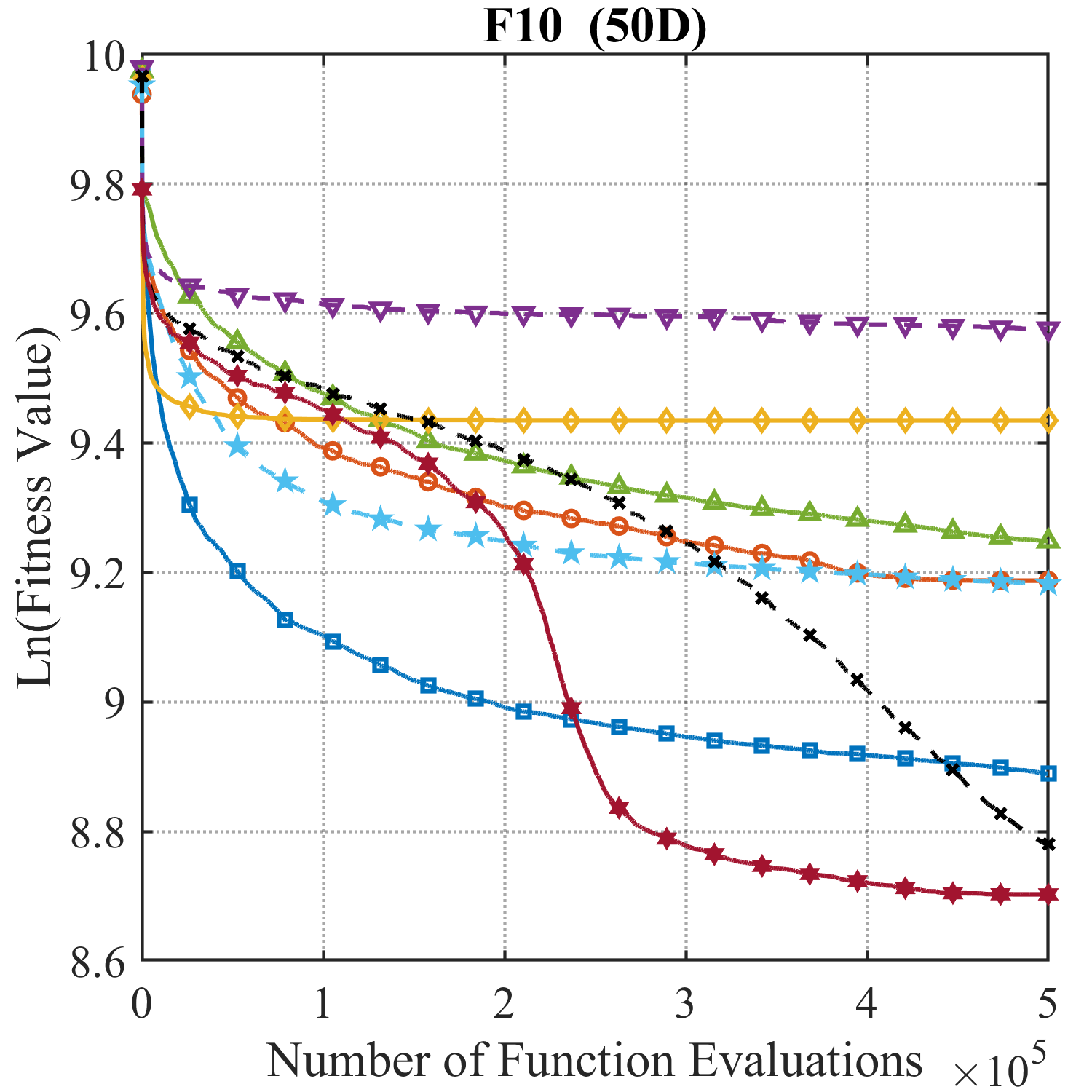}%
    \includegraphics[width=0.25\textwidth]{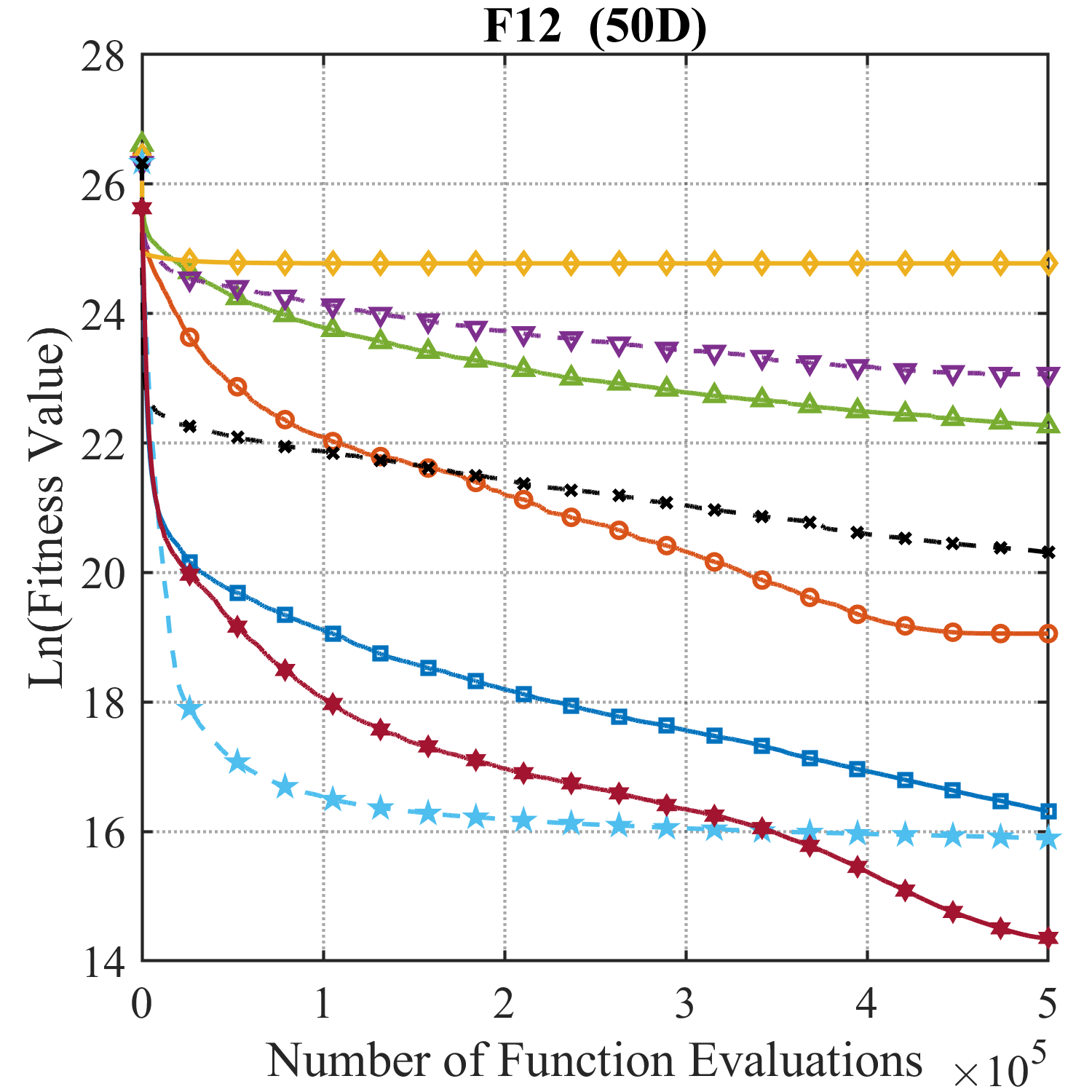}
     
    \includegraphics[width=0.25\textwidth]{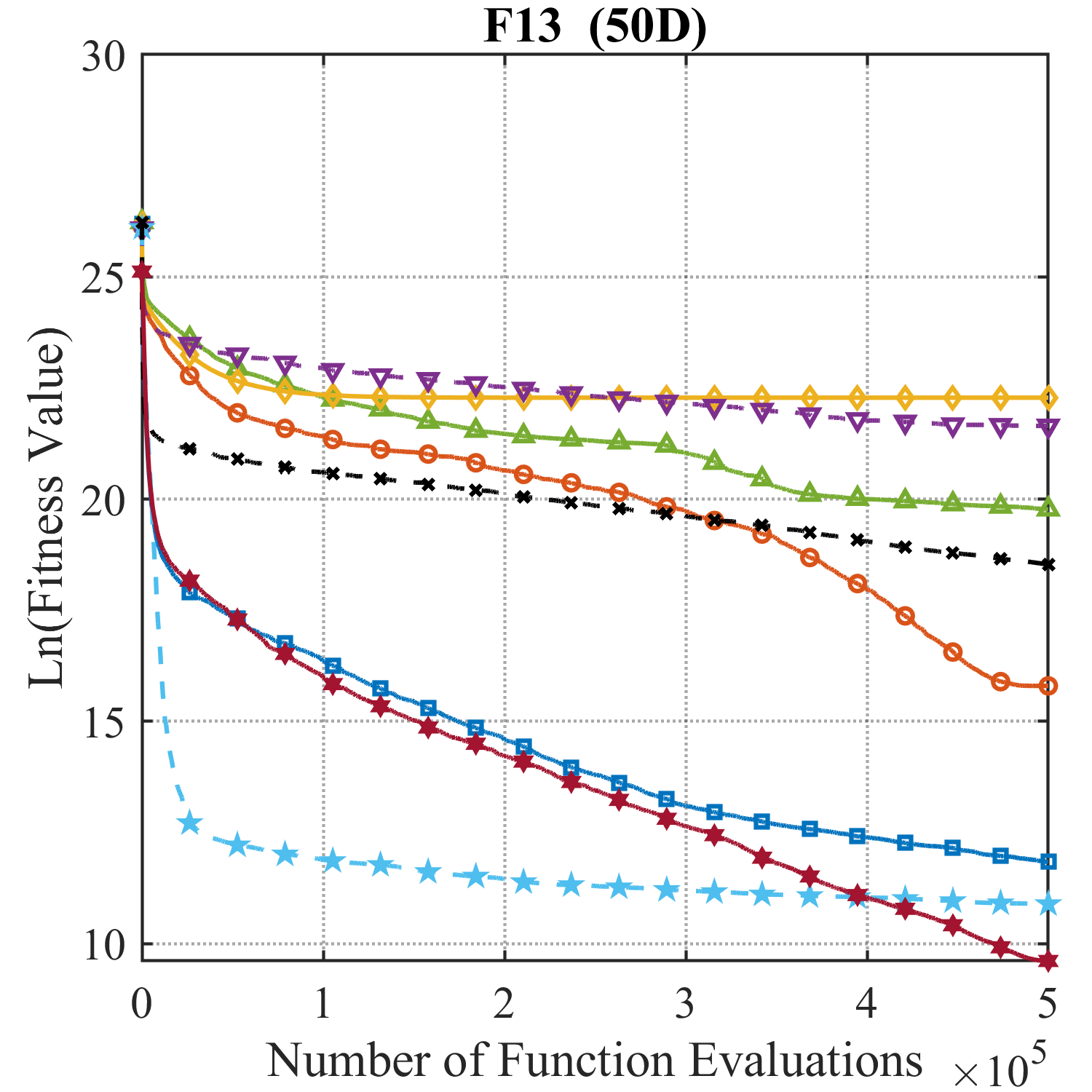}%
    \includegraphics[width=0.25\textwidth]{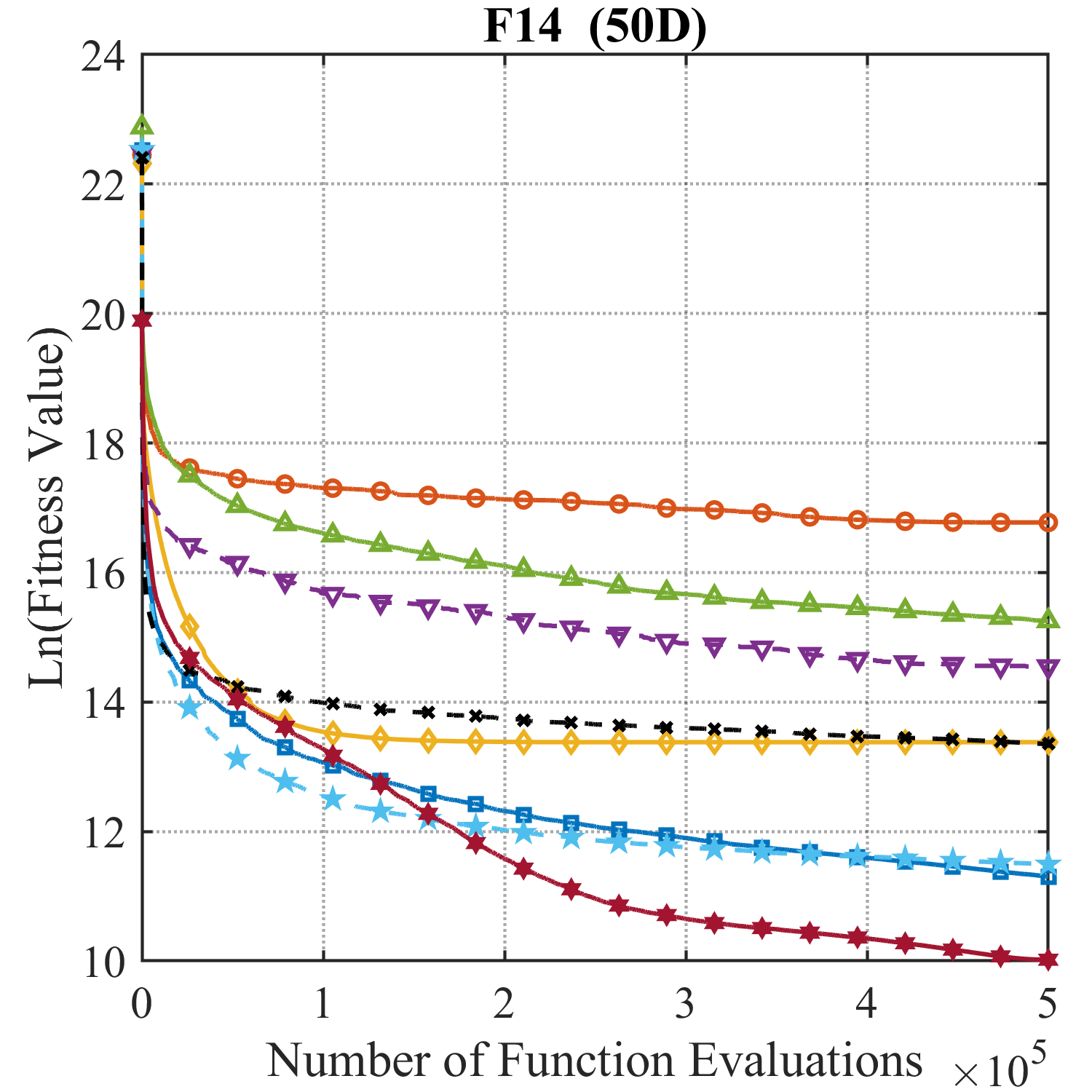}%
    \includegraphics[width=0.25\textwidth]{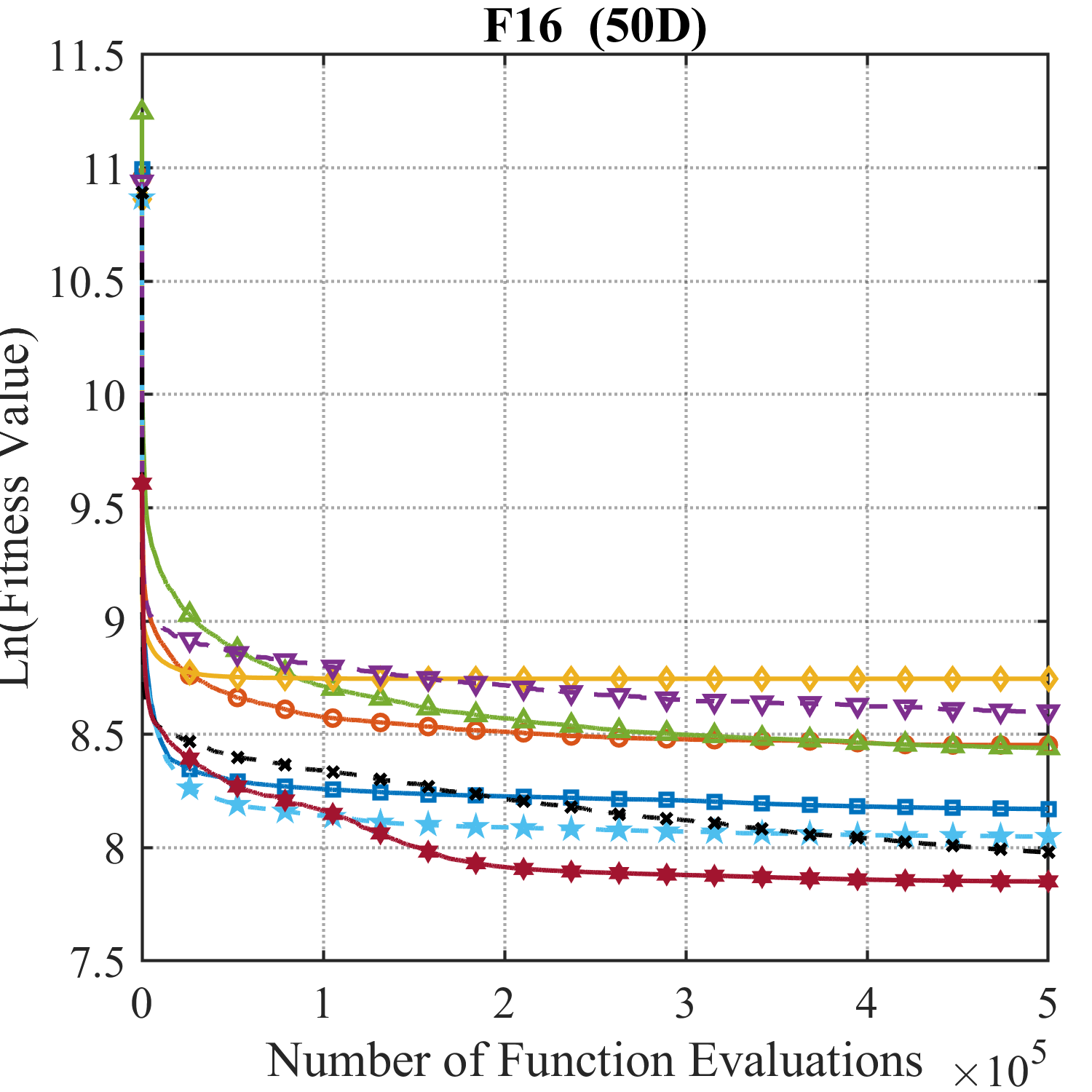}%
    \includegraphics[width=0.25\textwidth]{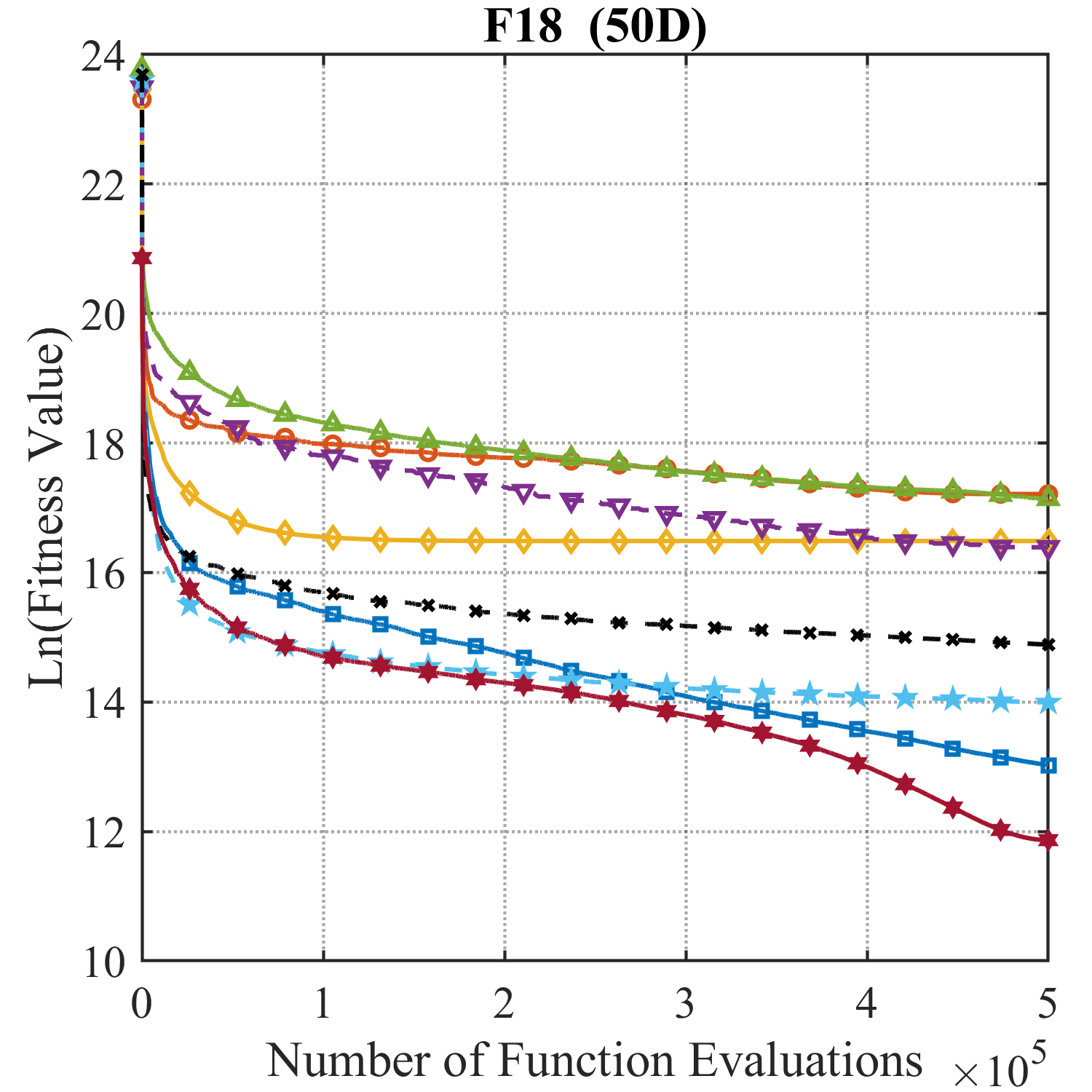}
       
    \includegraphics[width=0.25\textwidth]{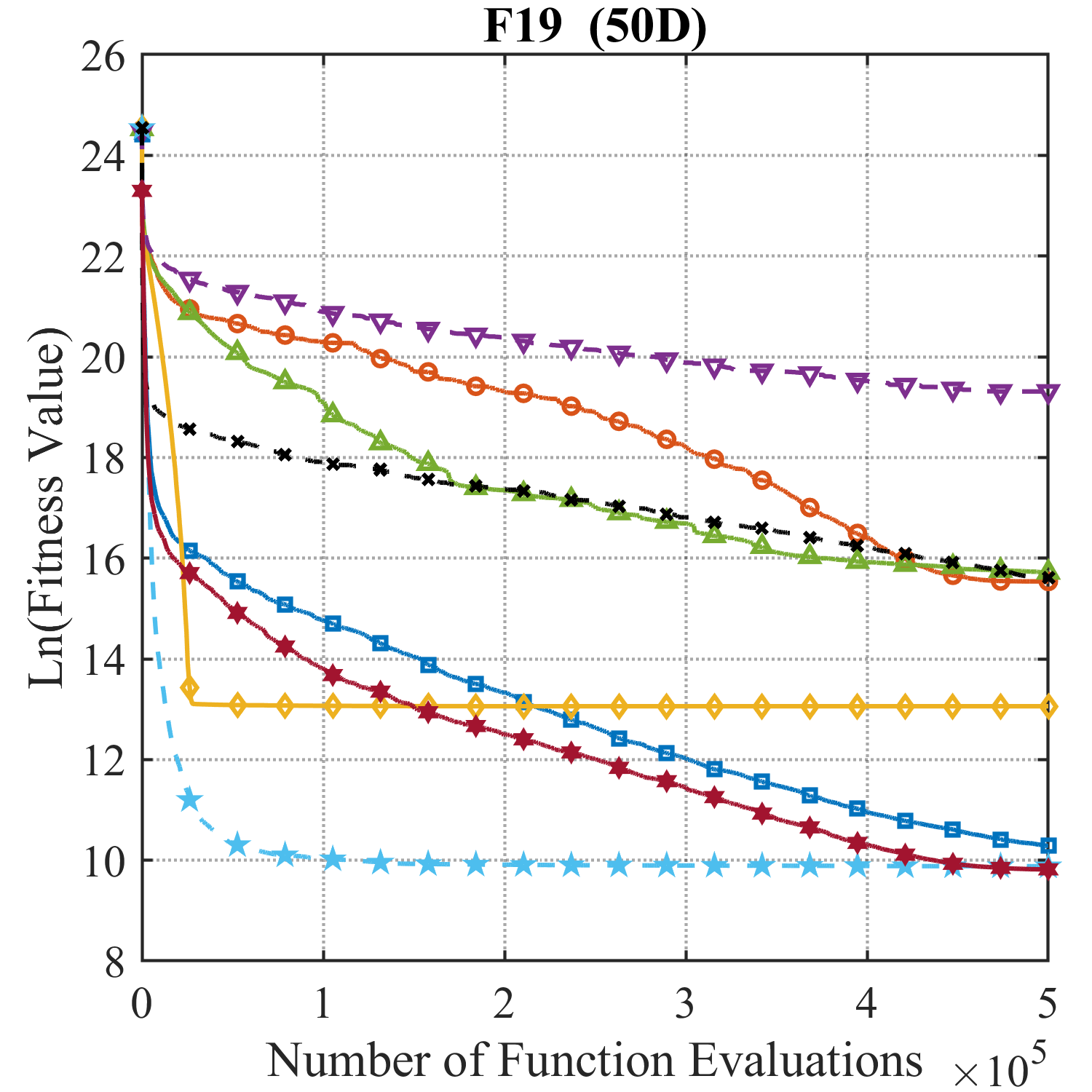}%
    \includegraphics[width=0.25\textwidth]{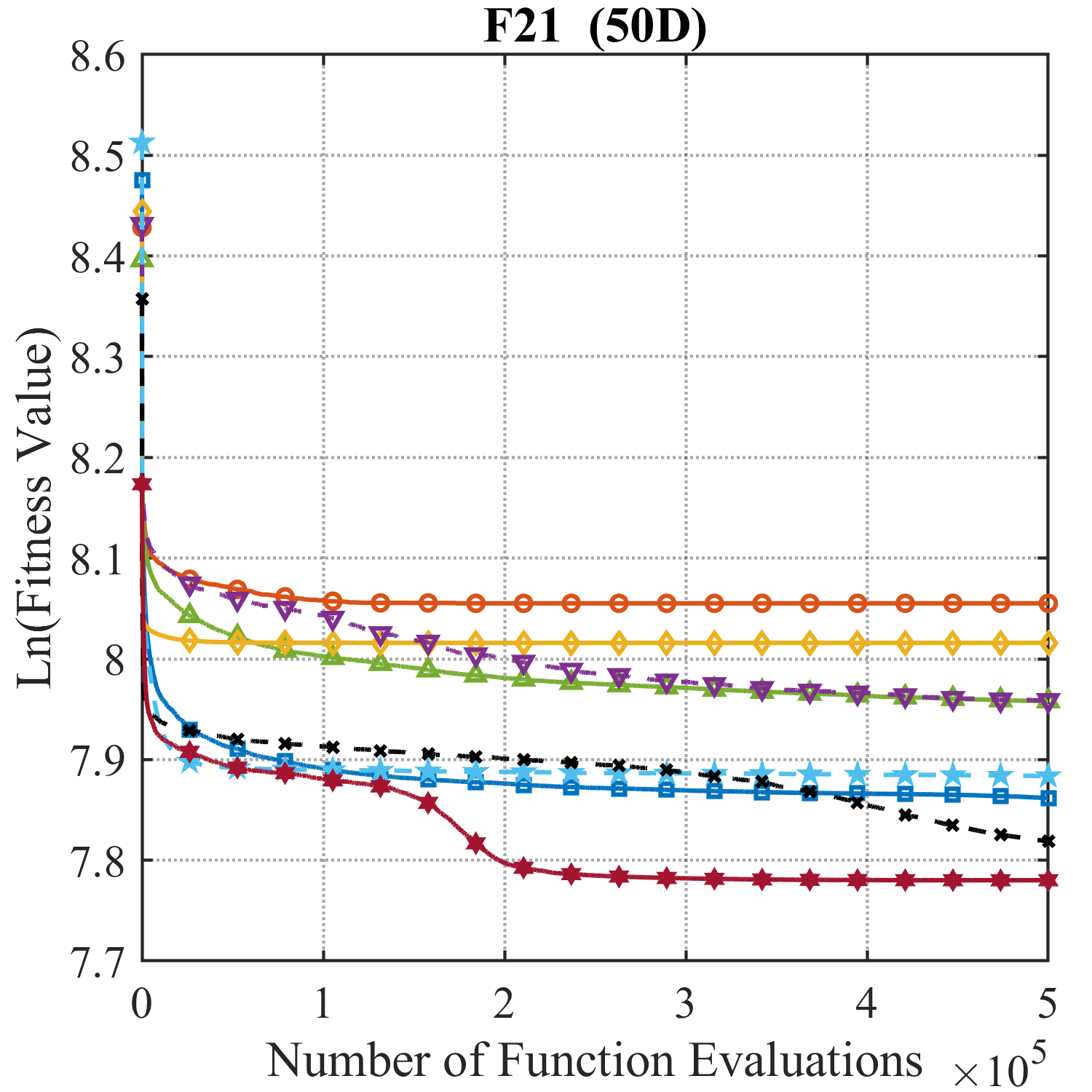}%
    \includegraphics[width=0.25\textwidth]{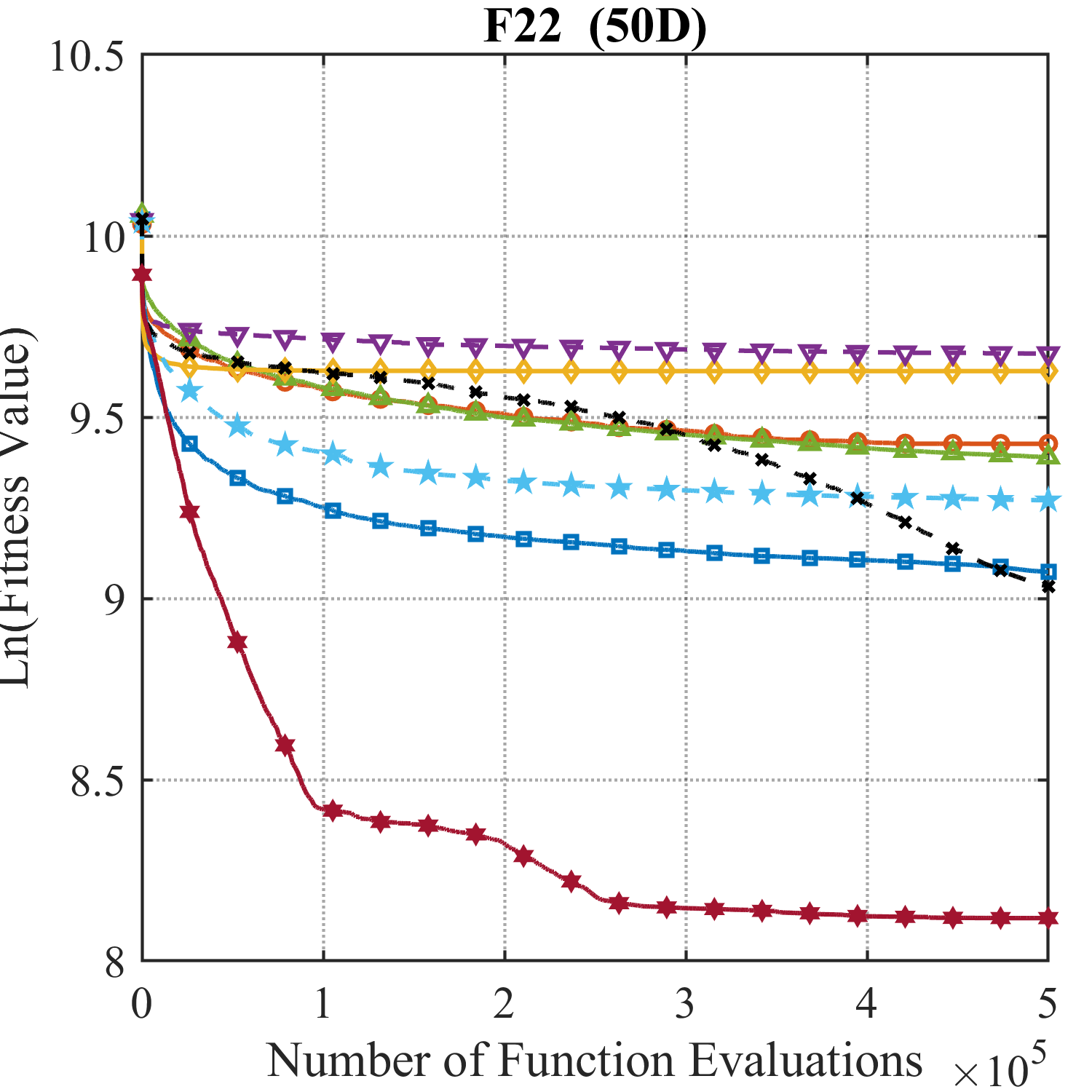}%
    \includegraphics[width=0.25\textwidth]{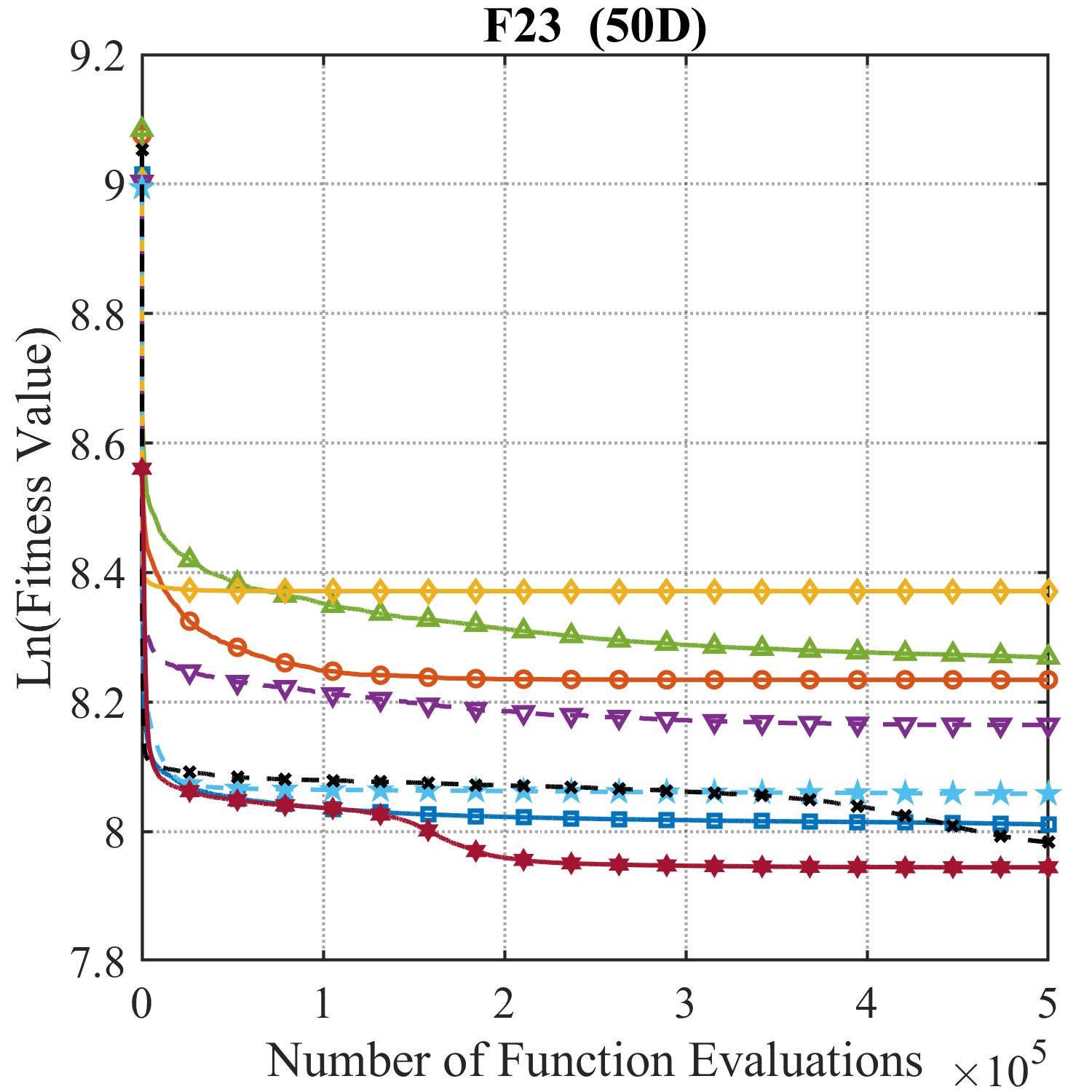}

    \includegraphics[width=0.25\textwidth]{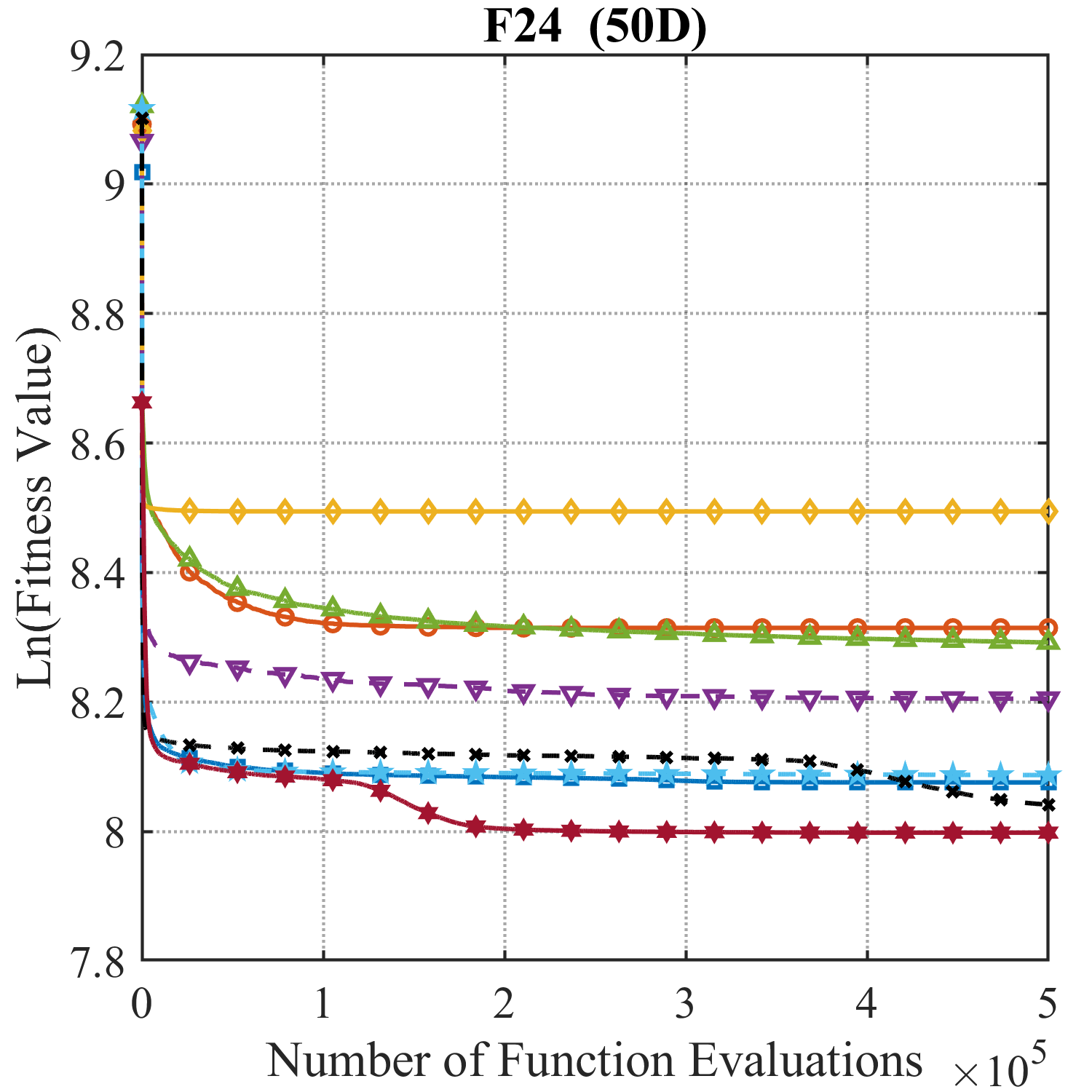}%
    \includegraphics[width=0.25\textwidth]{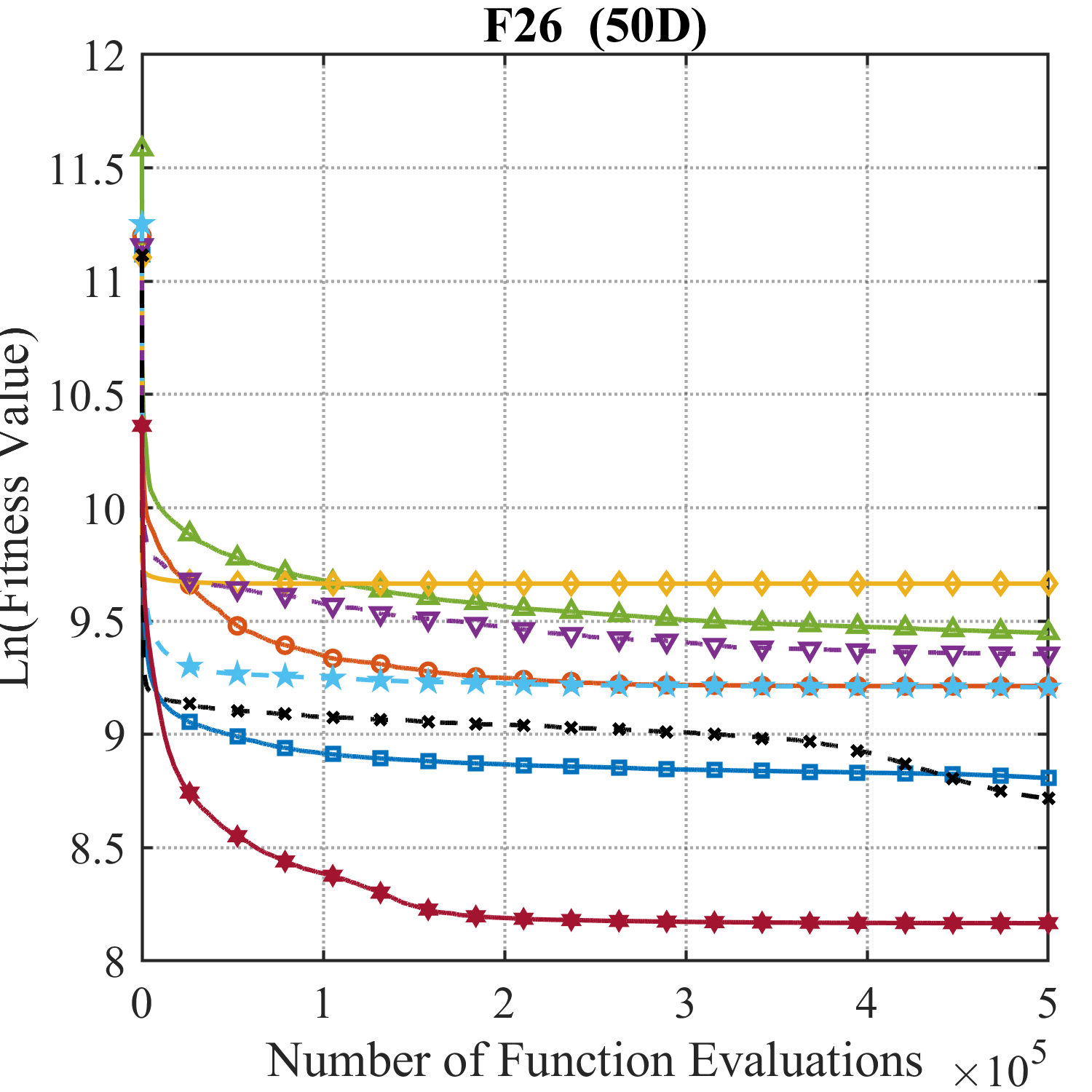}%
    \includegraphics[width=0.25\textwidth]{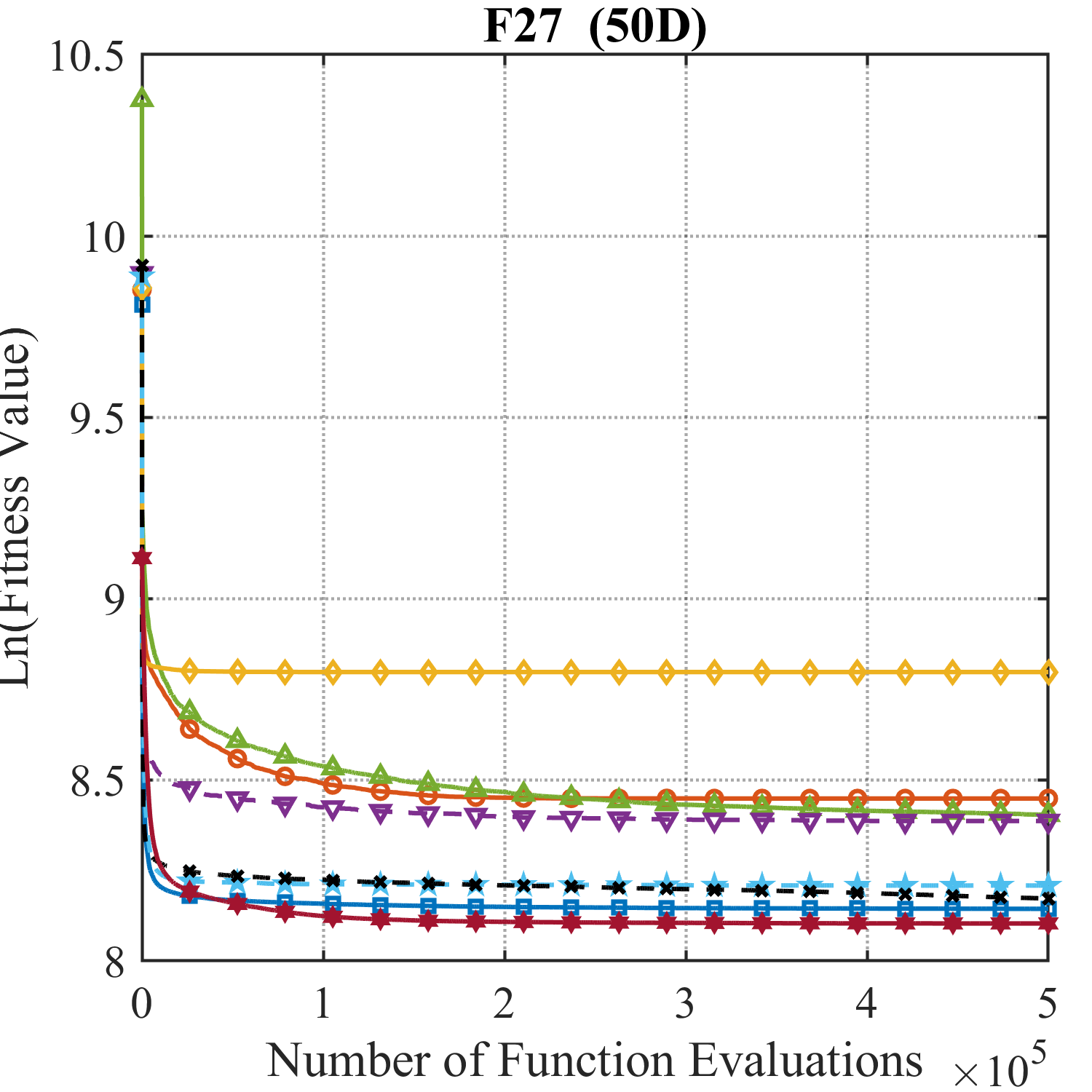}%
    \includegraphics[width=0.25\textwidth]{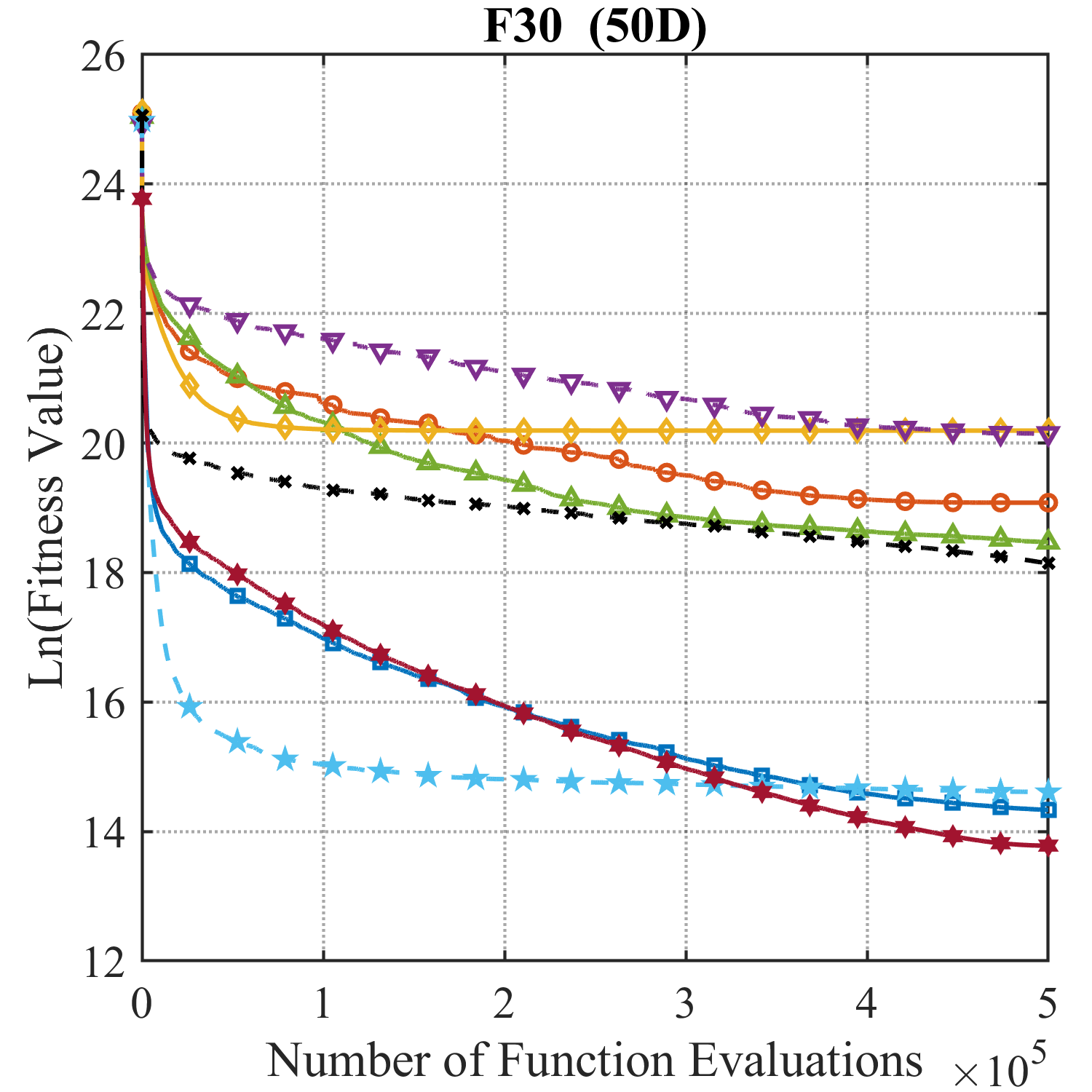}

    \caption{Convergence curves of BWE and comparative algorithms on some CEC2017 functions.}
    \label{fig:convergence_50D_part1}
\end{figure*}

\subsubsection{Convergence analysis}
Convergence speed and precision are pivotal metrics for evaluating evolutionary algorithms.
Representative convergence curves on the CEC2017 benchmark ($D=50$) are illustrated in \cref{fig:convergence_50D_part1}.
Initially, most algorithms exhibit a sharp descent, attributed to the transition from stochastic initialization to systematic operator guidance.
To balance global exploration and local exploitation, BWE prioritizes broad search space coverage early on, resulting in a characteristic steep decline followed by gradual stabilization.

For unimodal functions, BWE maintains a steady downward trend during the first 40\% of the $MaxFEs$, effectively preserving population diversity to prevent premature convergence.
Between 40\% and 80\% of the $MaxFEs$, it frequently identifies promising regions, triggering a secondary accelerated descent and significant precision refinement.
For instance, the rapid convergence on F1 in the final 20\% of the search highlights BWE's efficient local exploitation, while on F9, it achieves superior initial precision from the onset.

Furthermore, on multimodal like F5, F7, F8, and F10, BWE exhibits a distinctive rapid descent within the $20\% \sim 40\%$ iteration interval.
This unique "staircase" convergence behavior represents a critical transition from exploration to exploitation, which is largely absent in rival algorithms, validating BWE's effectiveness in escaping local optima.

Regarding hybrid and composition functions, BWE demonstrates sustained optimization capabilities across complex landscapes (e.g., F12, F13, F14, F18, F19).
Although it may not always exhibit the most aggressive initial speed, it maintains steady refinement throughout the search.
In contrast, algorithms like AOA and COA, while occasionally achieving high initial precision, are more susceptible to stagnation in local basins, making BWE's gradual but persistent convergence a more resilient strategy.
Additionally, on functions like F21, F22, and F24, BWE exhibits a significant mid-stage downward trend (20\%$\sim$50\%), successfully executing leaps from local optima and resuming productive search in later evolutionary stages.
On remaining functions like F27, BWE remains highly competitive, consistently reaching higher precision levels before stagnation.

\subsubsection{Box plot analysis} 
\begin{figure*}[!htbp]
    \centering

    \includegraphics[width=0.23\textwidth]{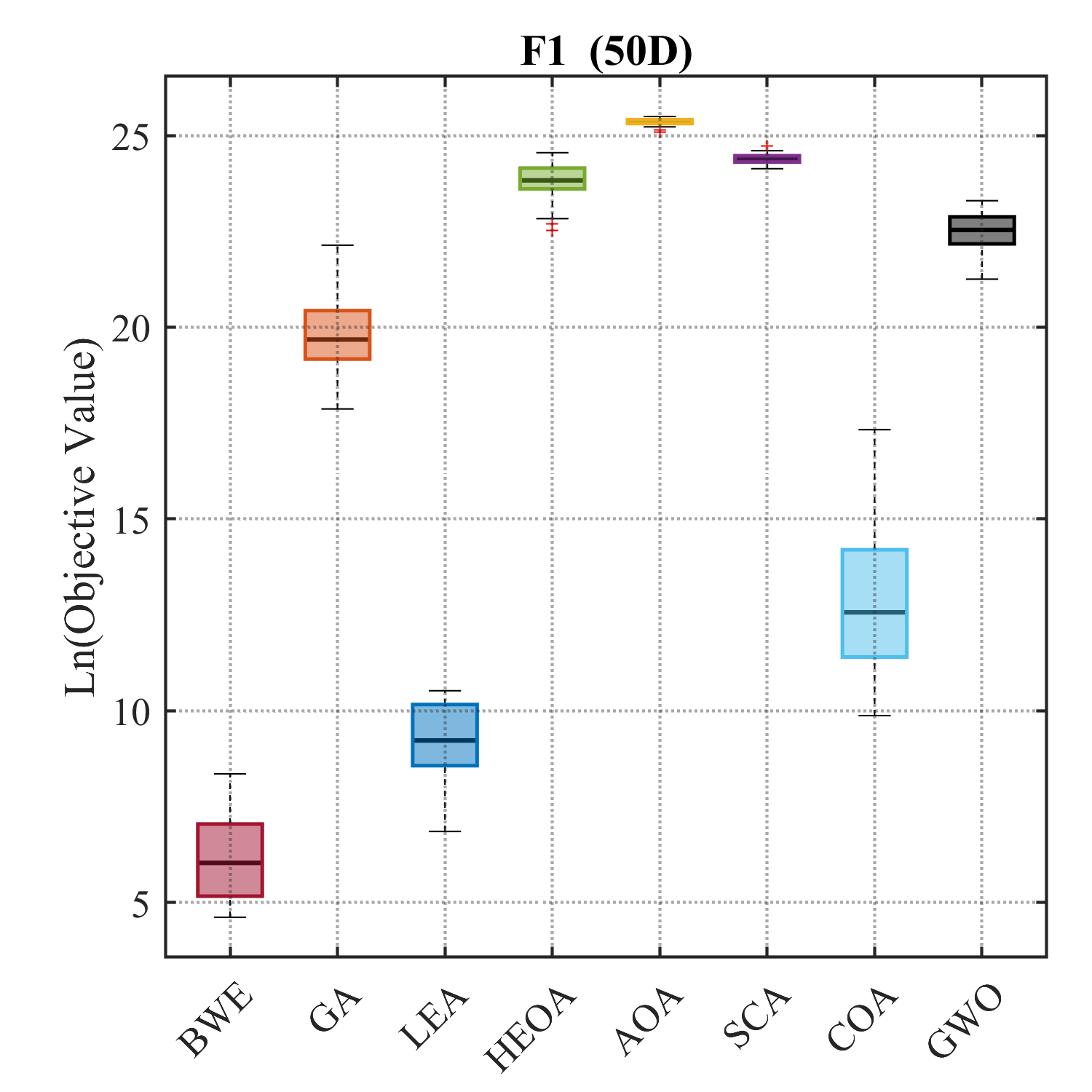}%
    \includegraphics[width=0.23\textwidth]{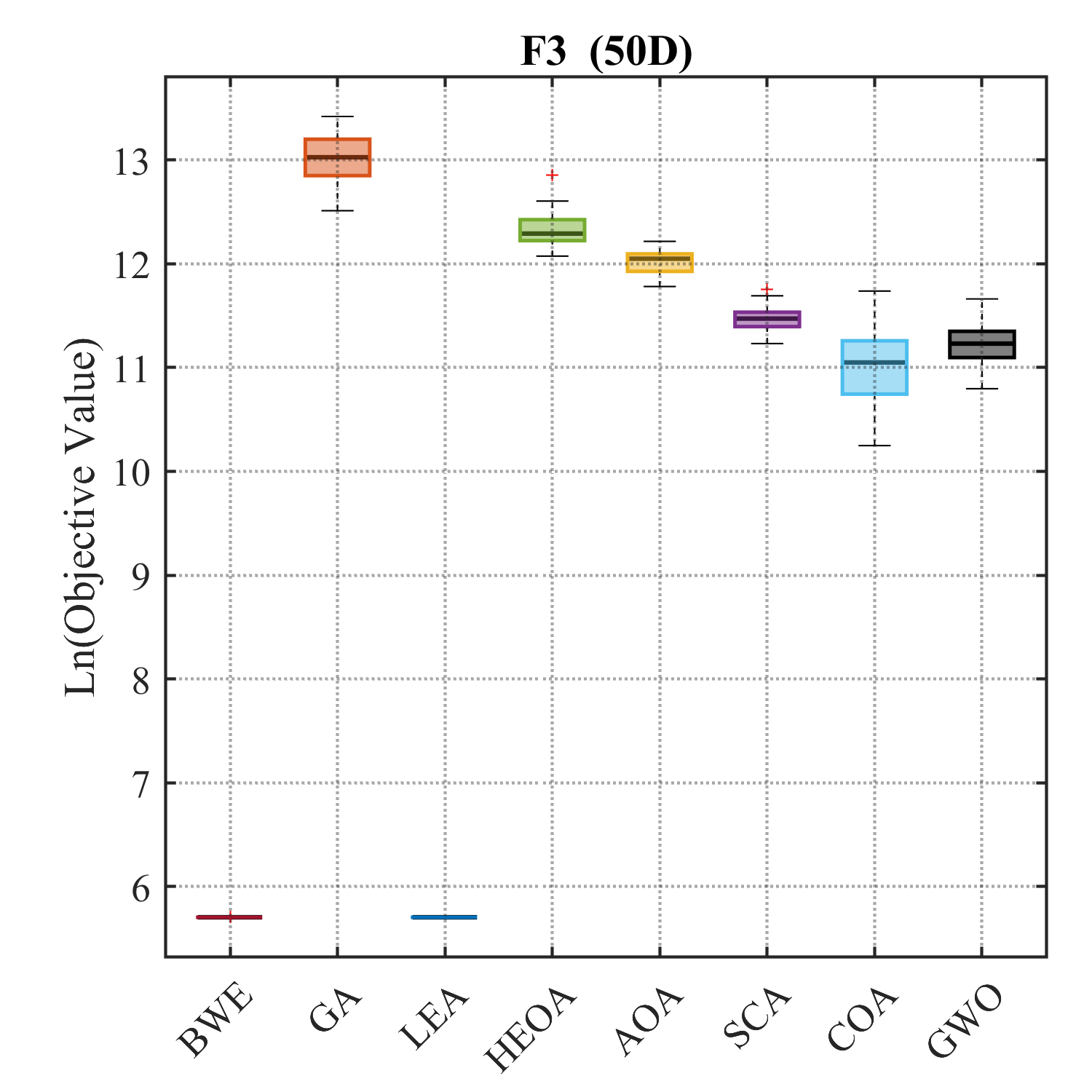}%
    \includegraphics[width=0.23\textwidth]{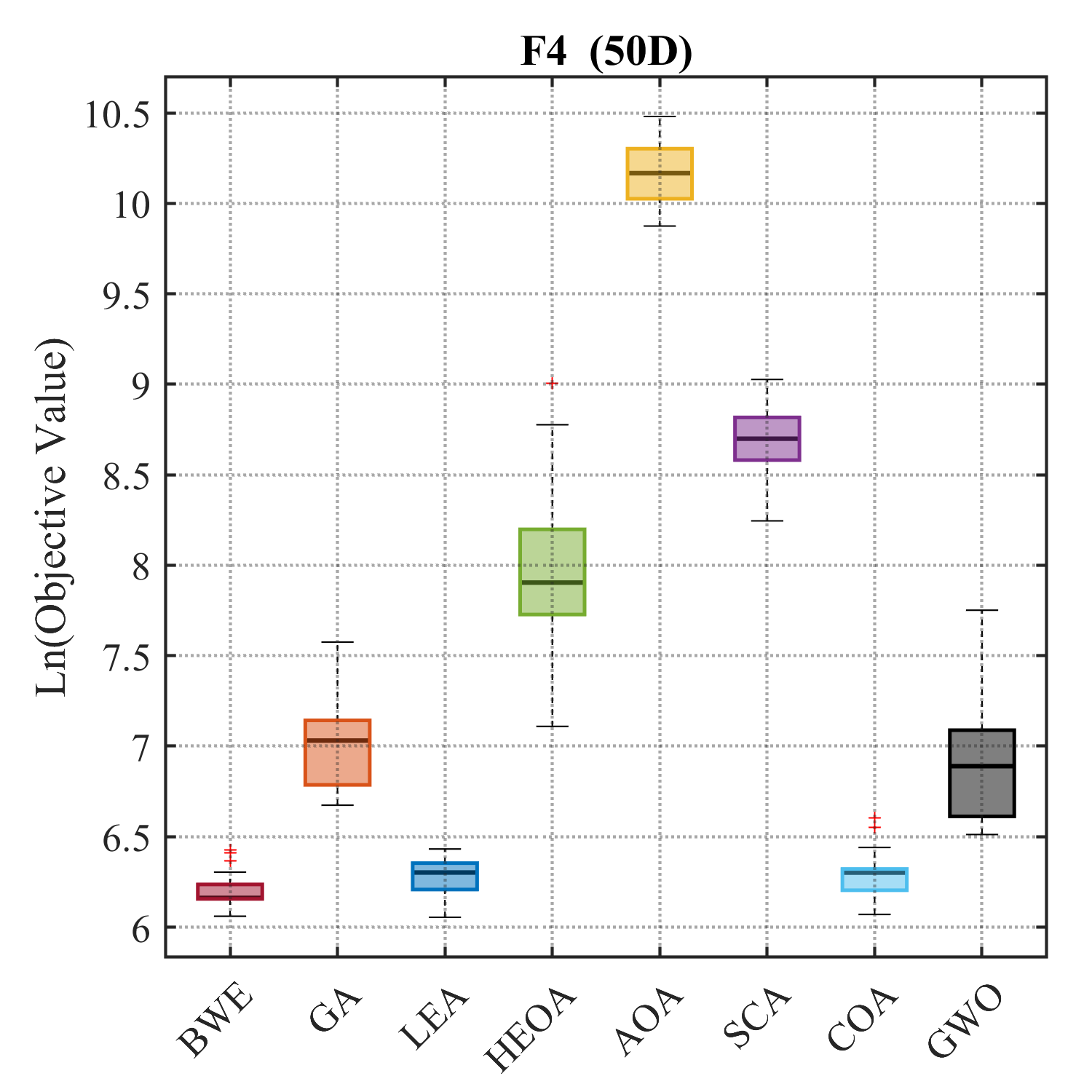}%
    \includegraphics[width=0.23\textwidth]{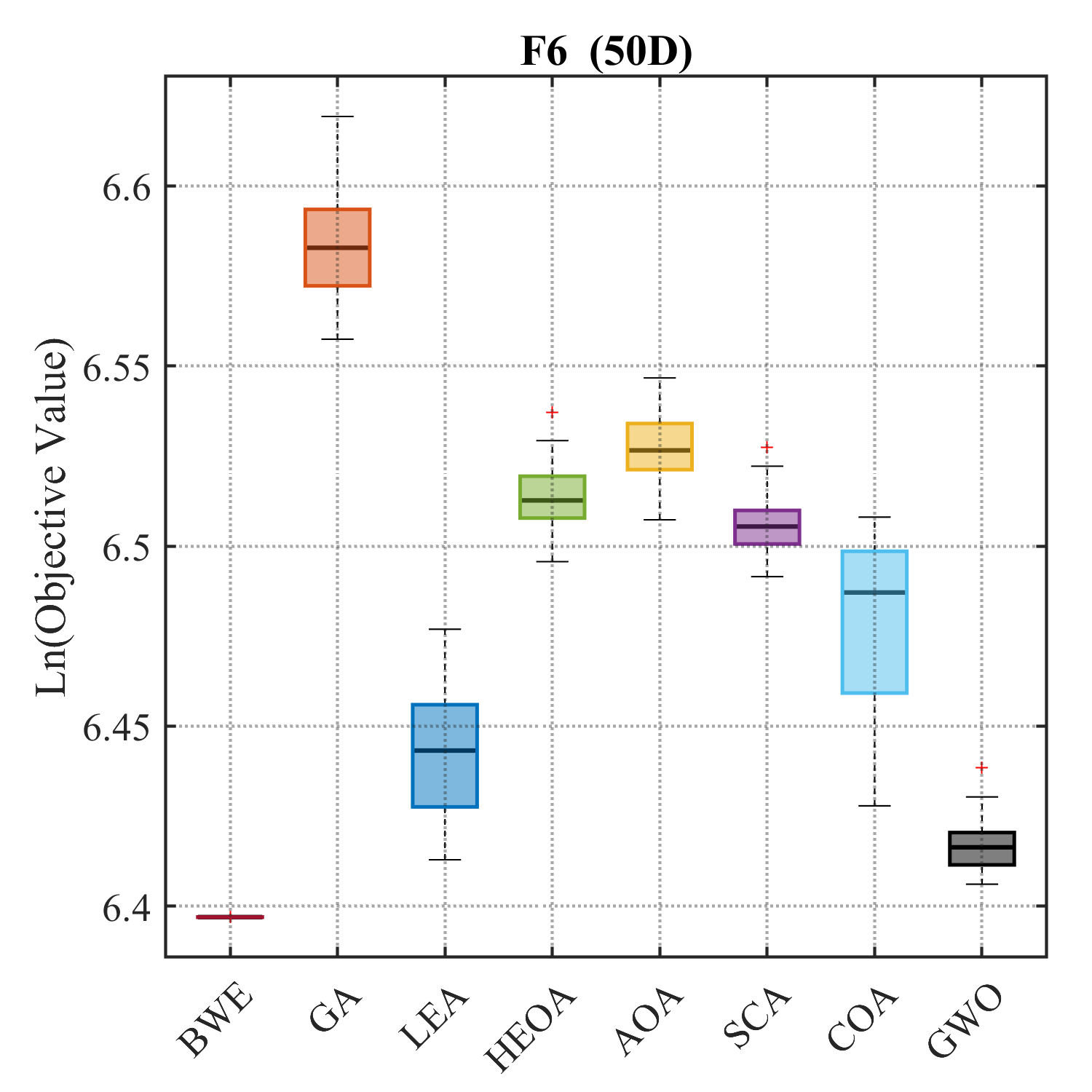}
   
    \includegraphics[width=0.23\textwidth]{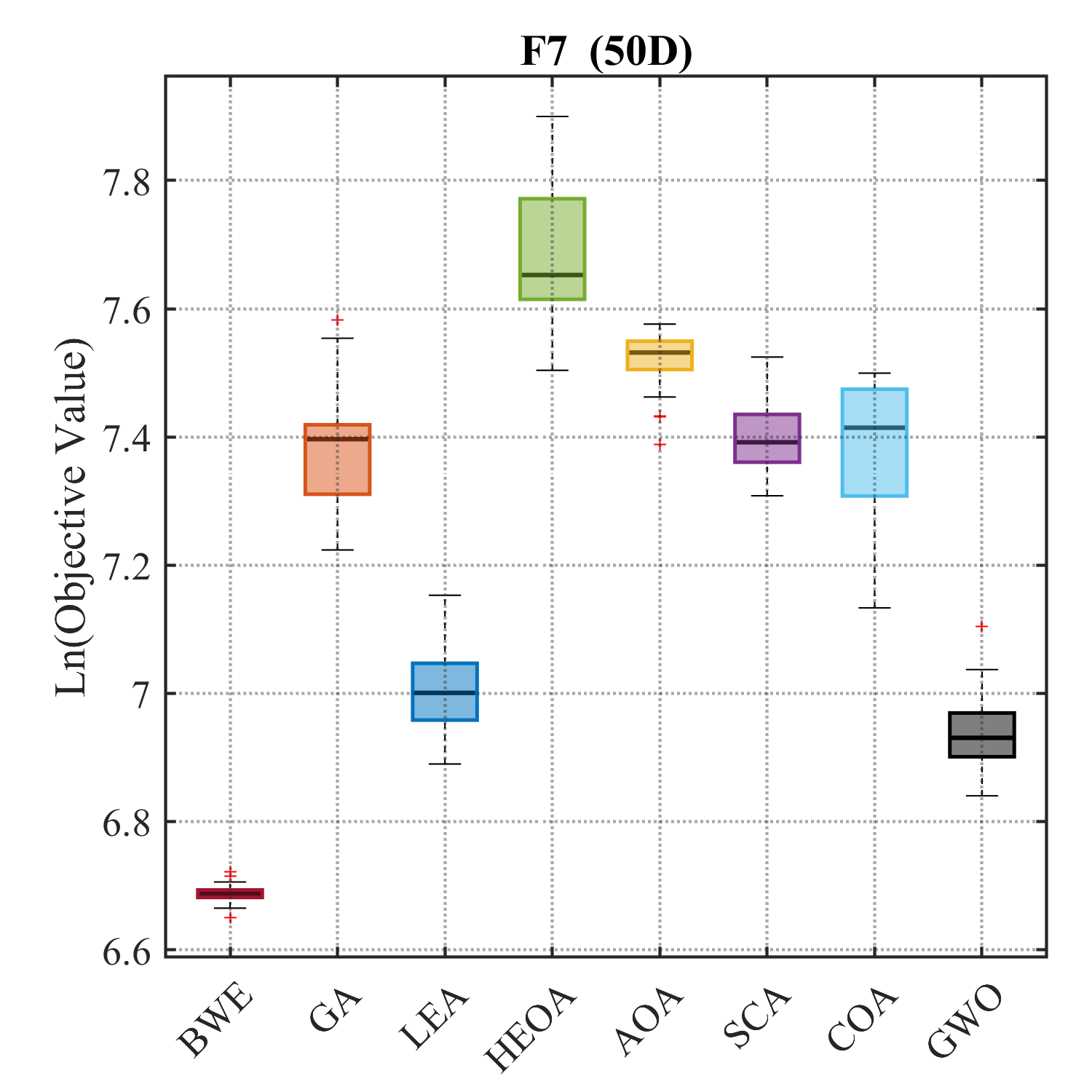}%
    \includegraphics[width=0.23\textwidth]{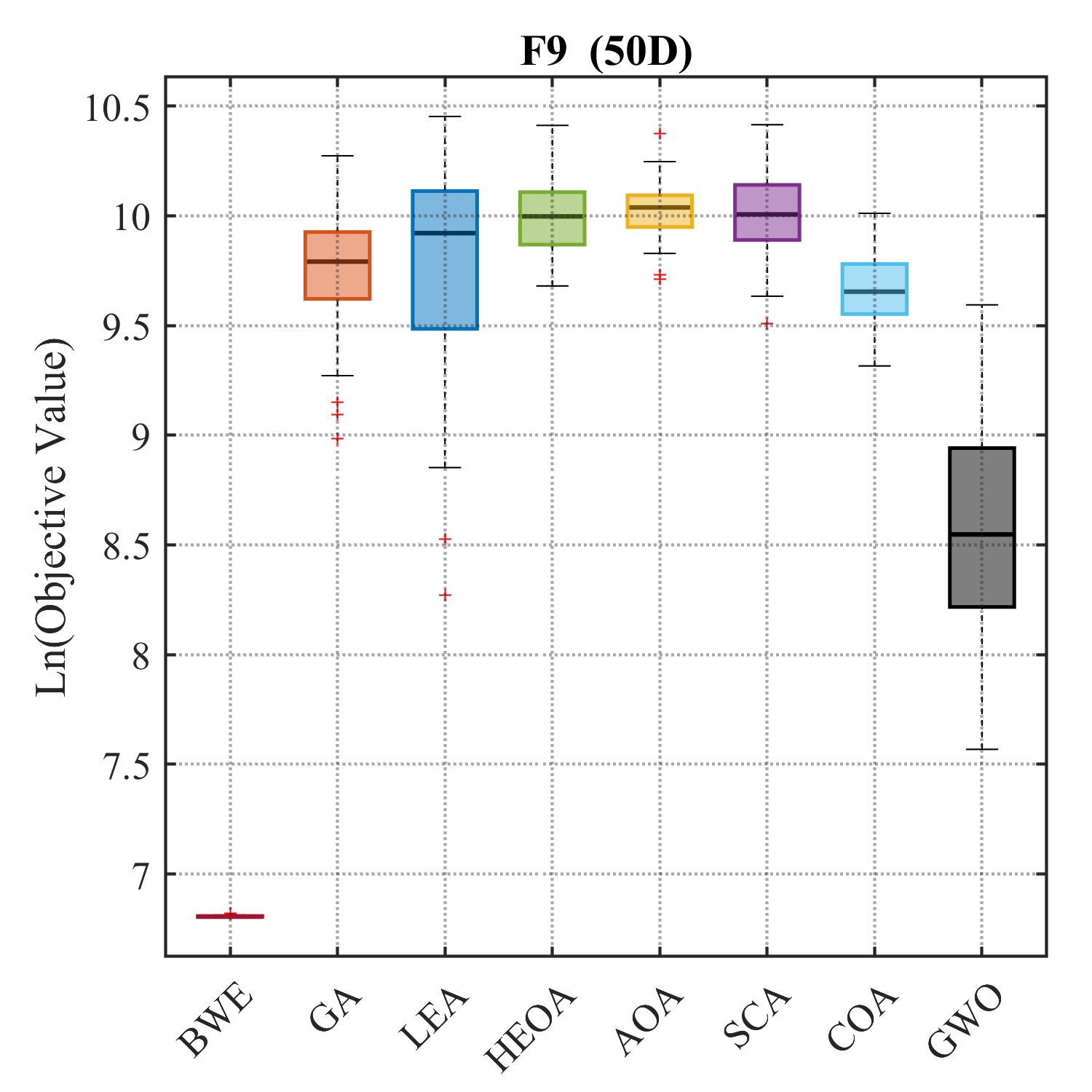}%
    \includegraphics[width=0.23\textwidth]{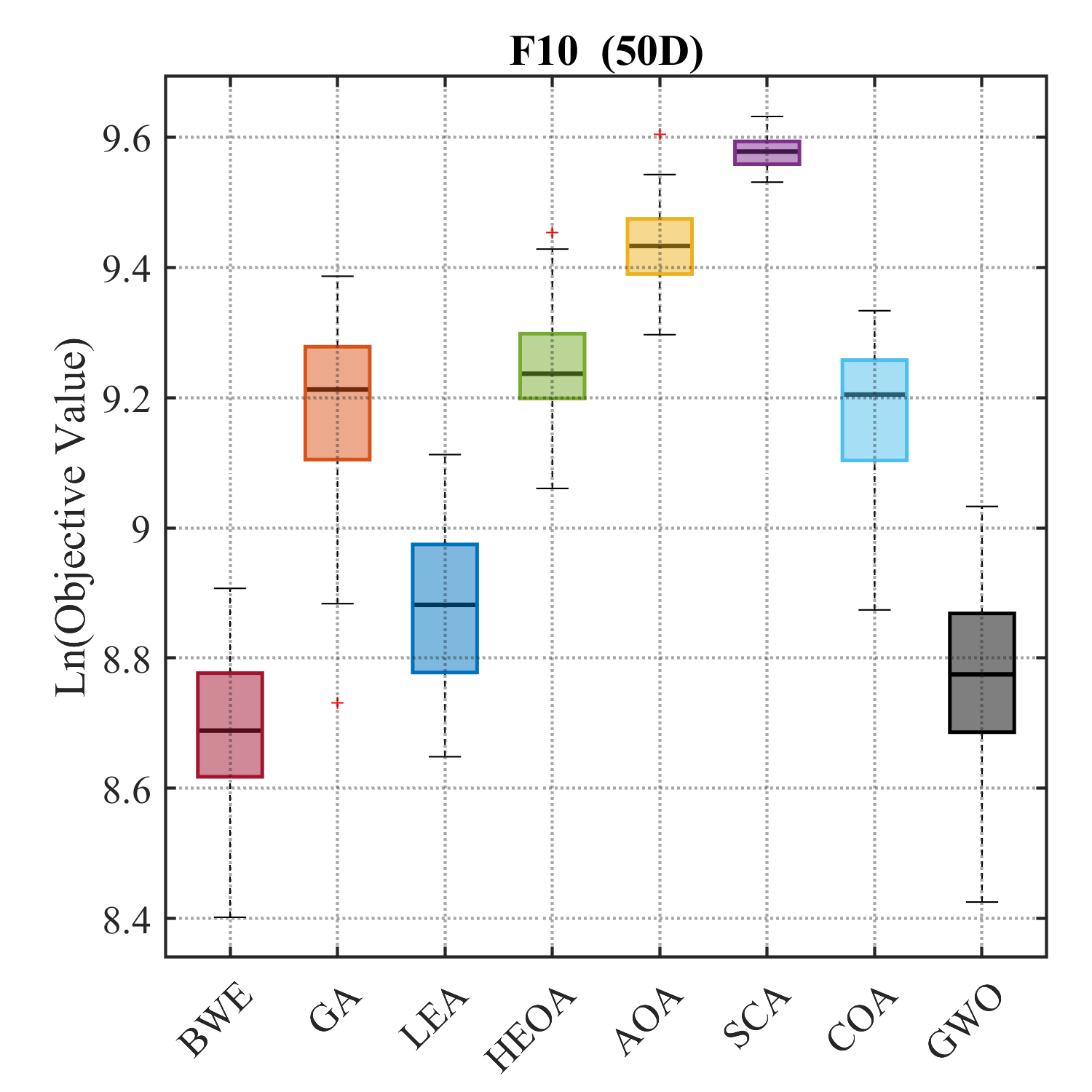}%
    \includegraphics[width=0.23\textwidth]{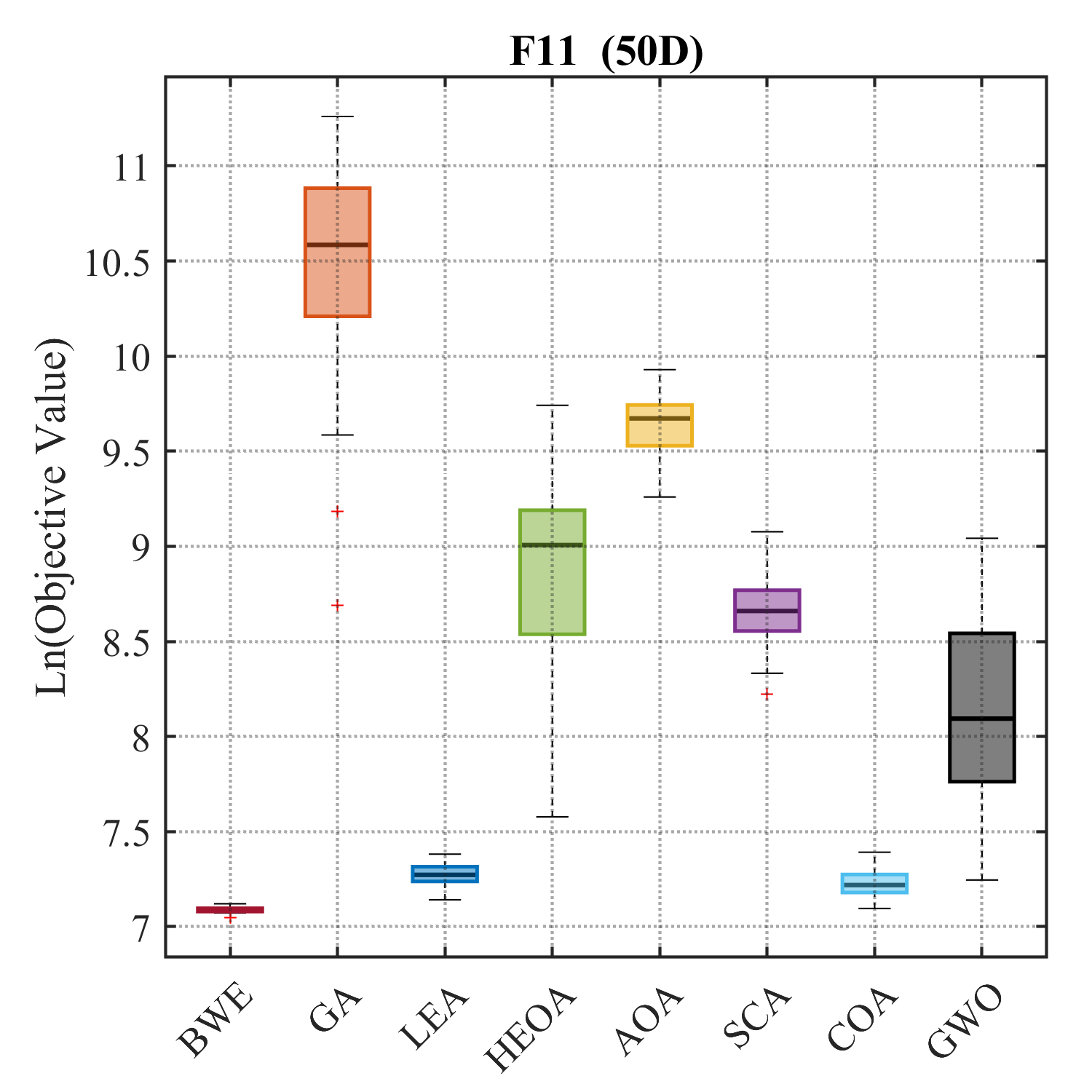}
    
    \includegraphics[width=0.23\textwidth]{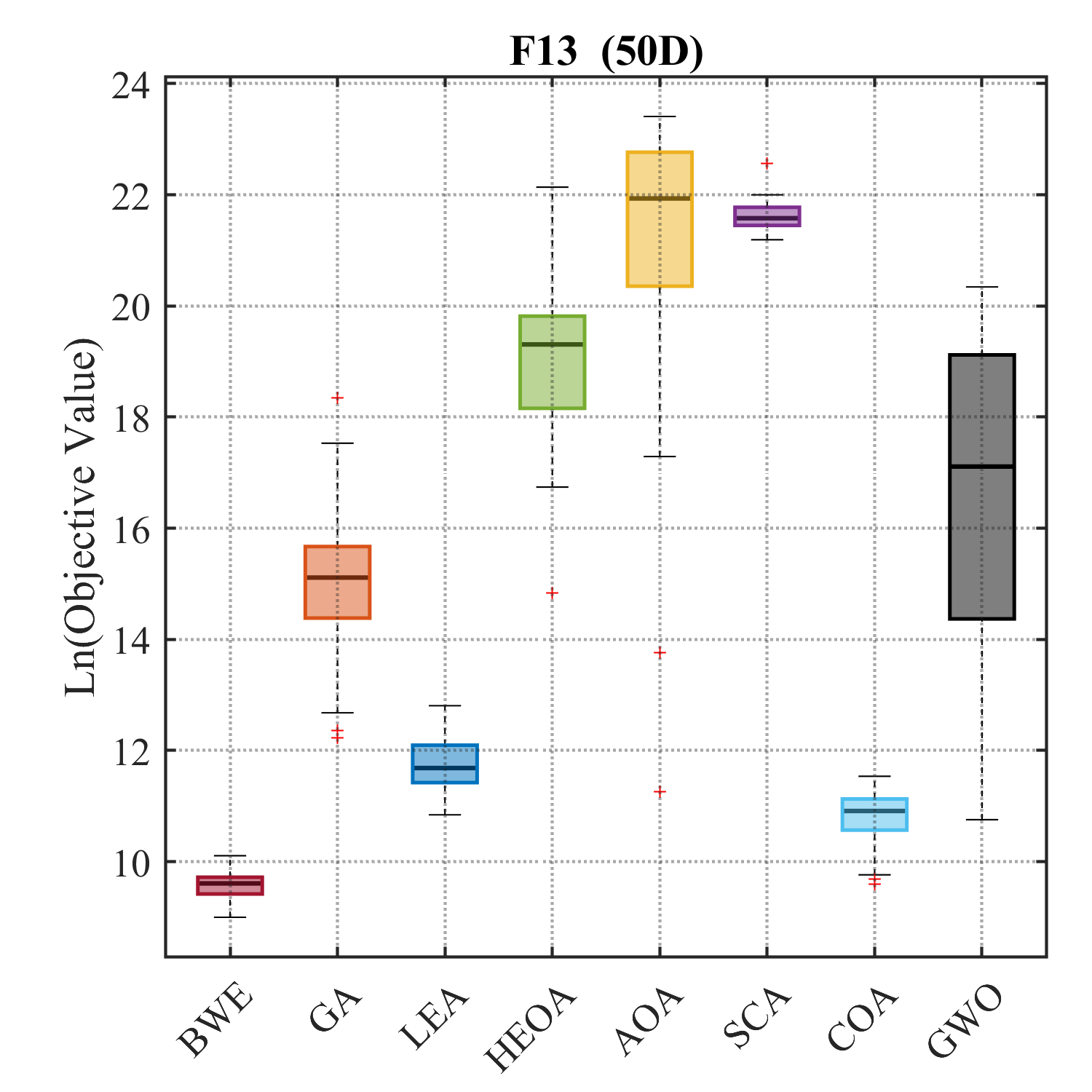}%
    \includegraphics[width=0.23\textwidth]{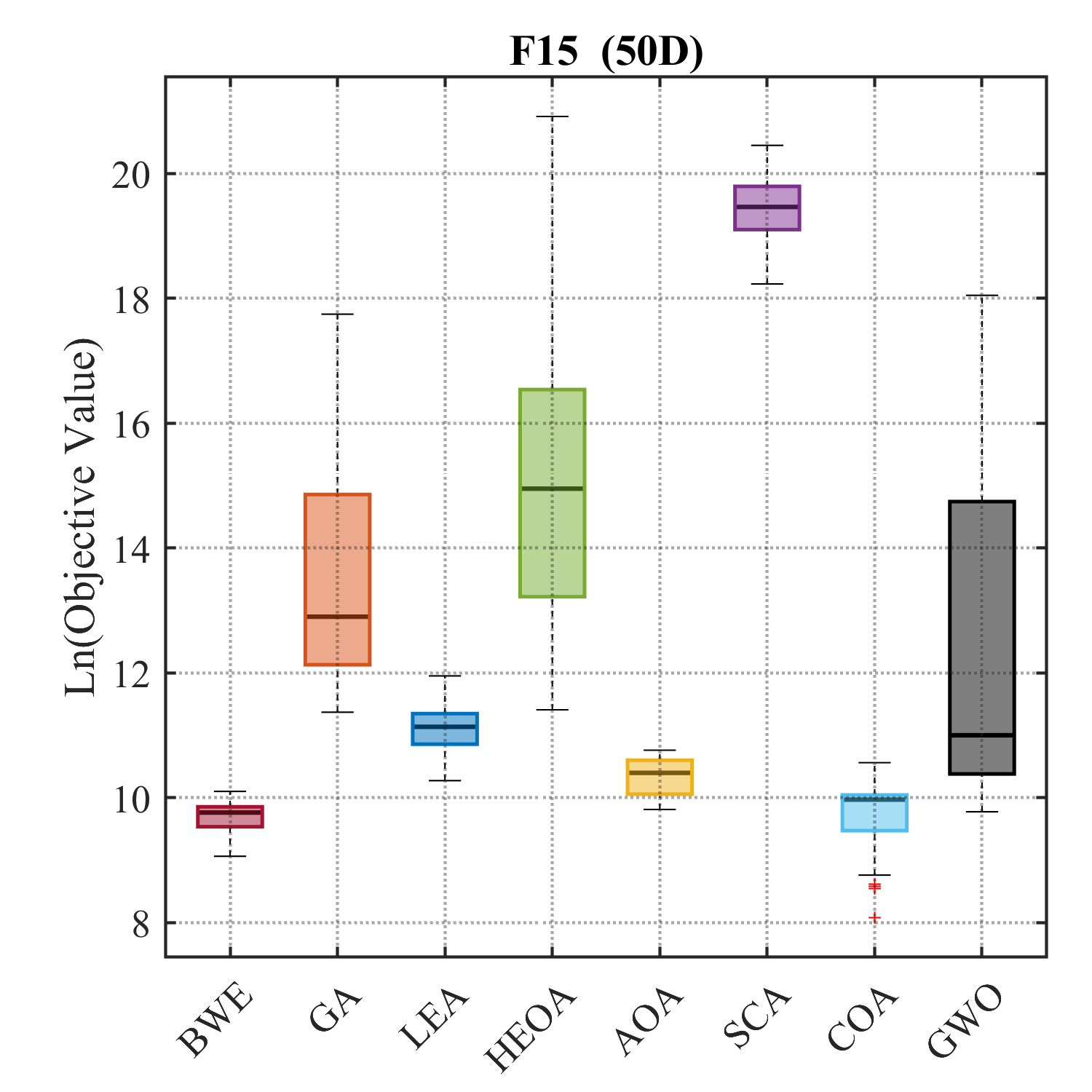}%
    \includegraphics[width=0.23\textwidth]{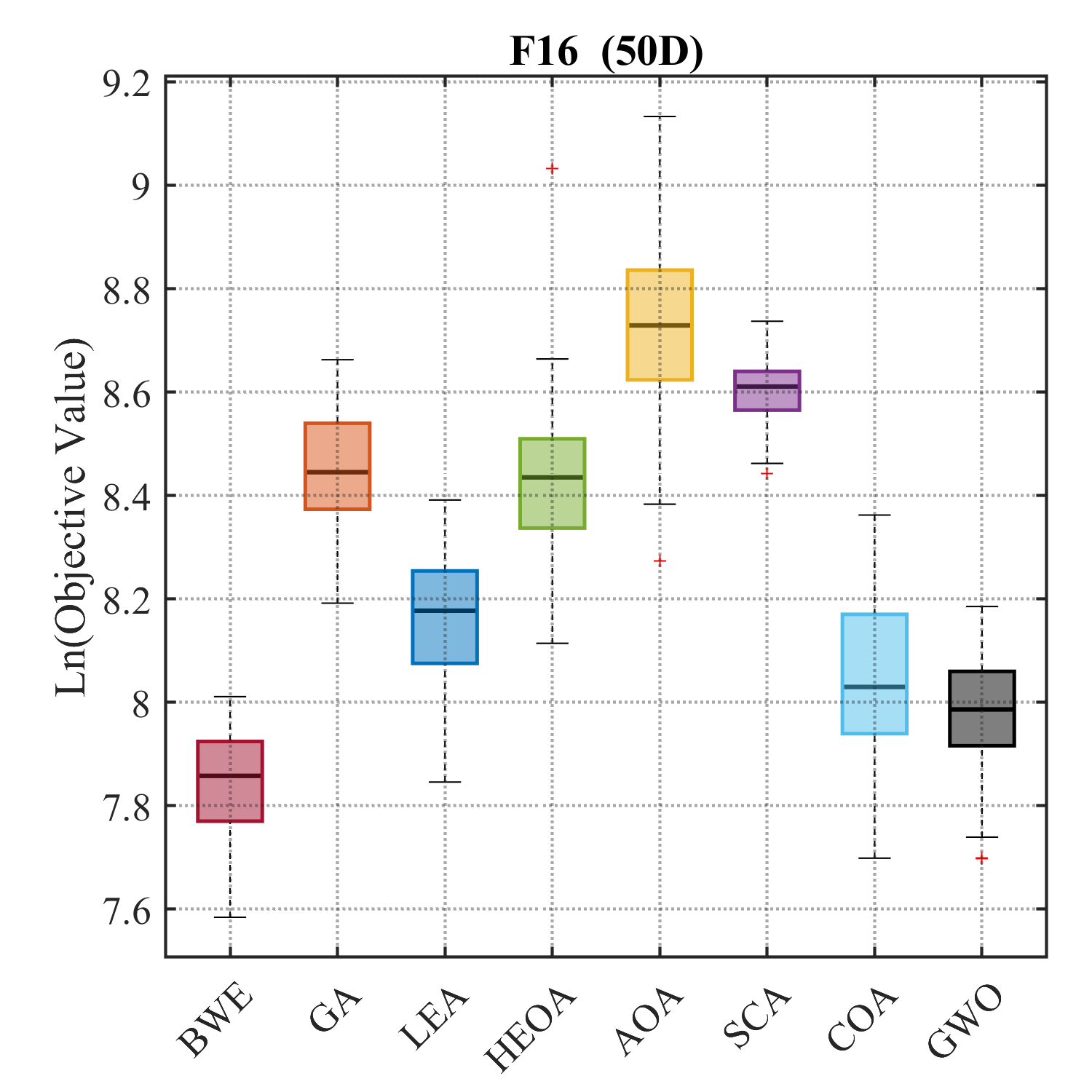}%
    \includegraphics[width=0.23\textwidth]{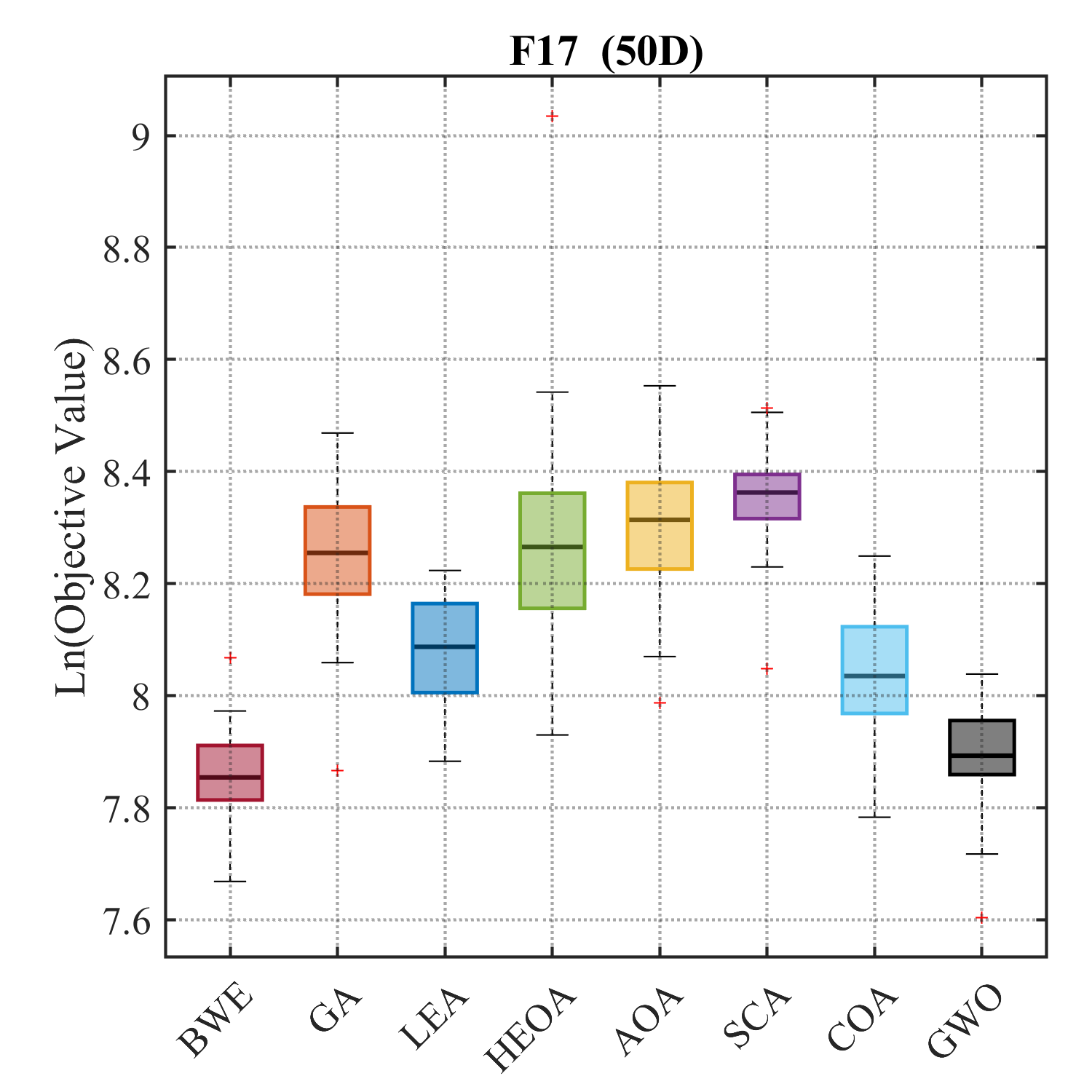}
 
    \includegraphics[width=0.23\textwidth]{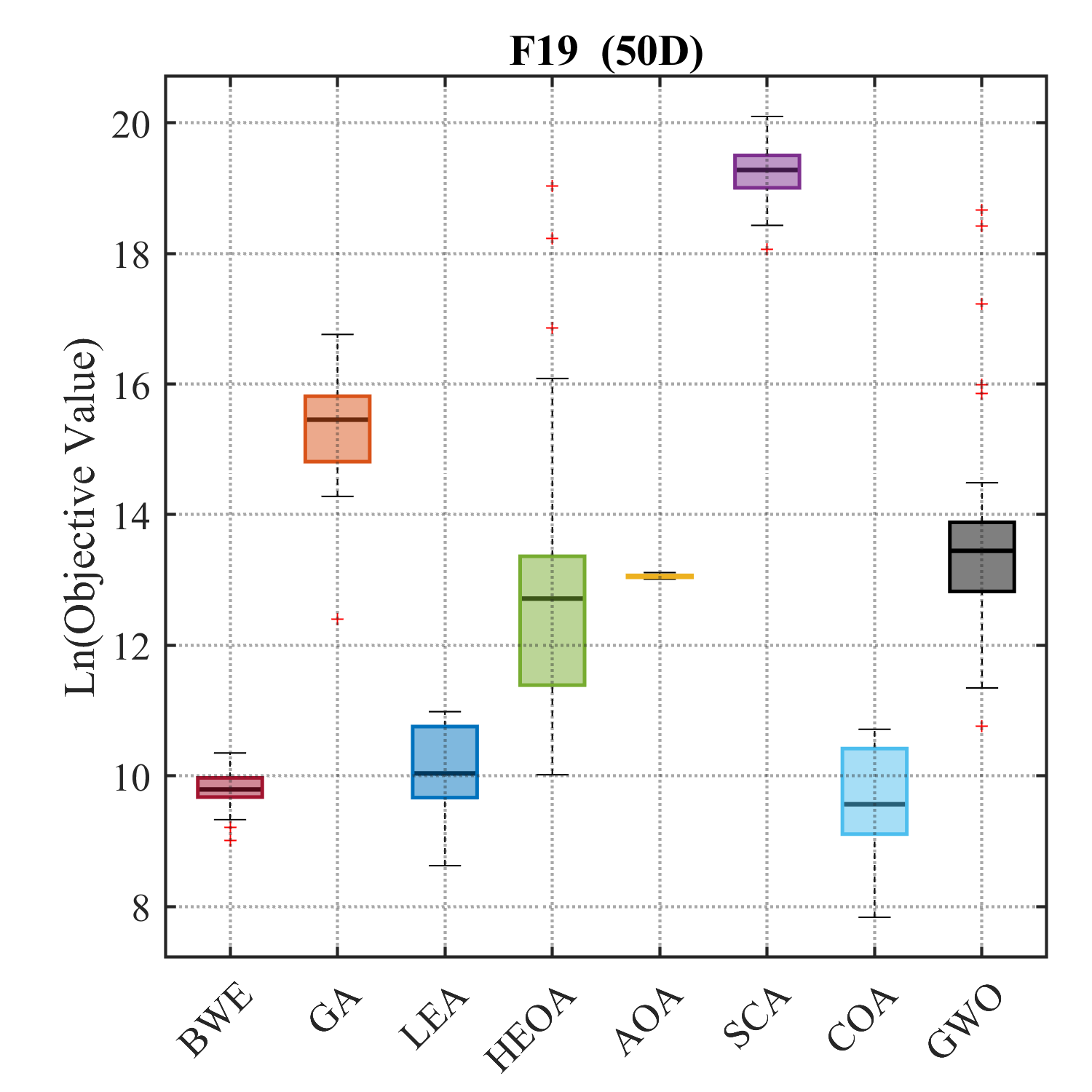}%
    \includegraphics[width=0.23\textwidth]{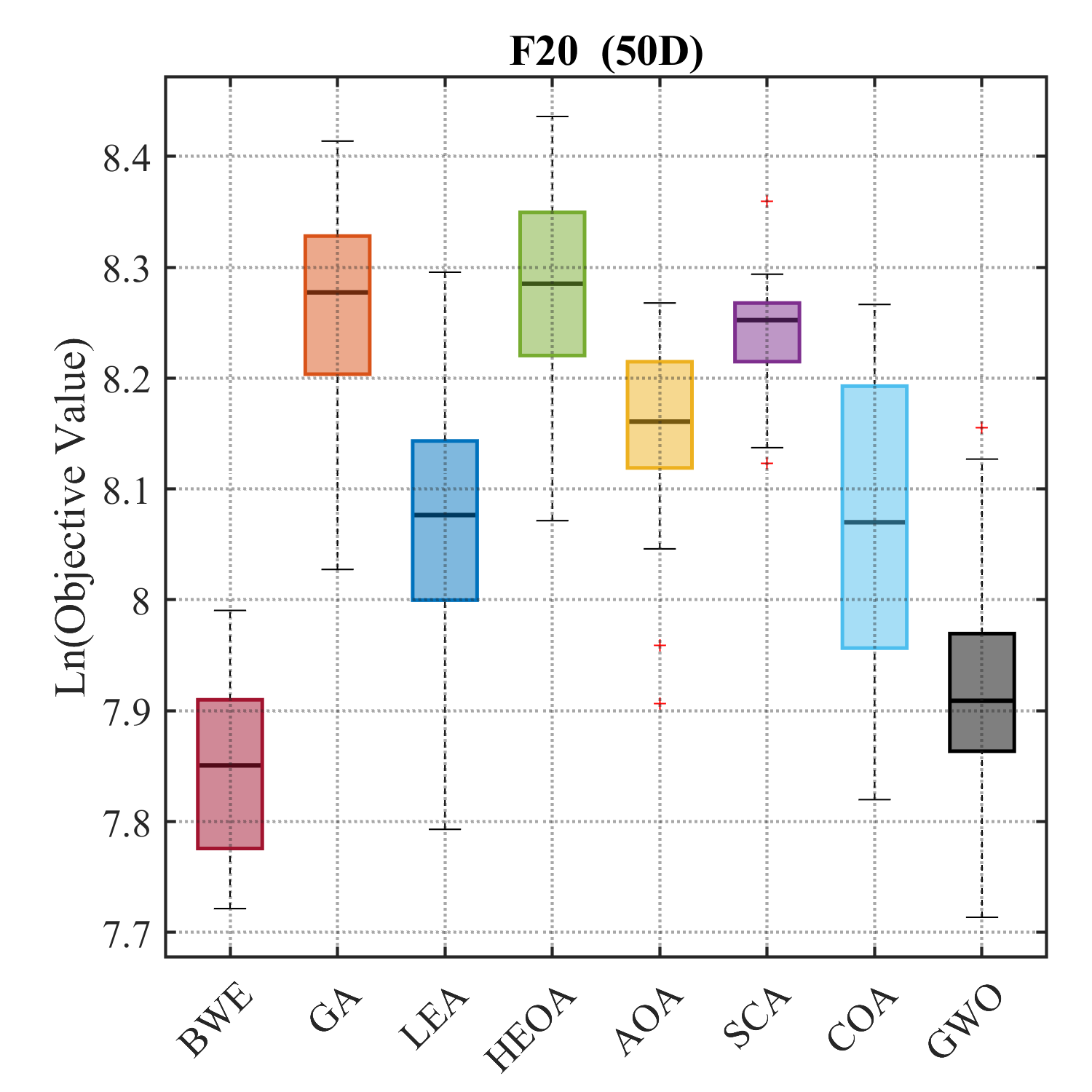}%
    \includegraphics[width=0.23\textwidth]{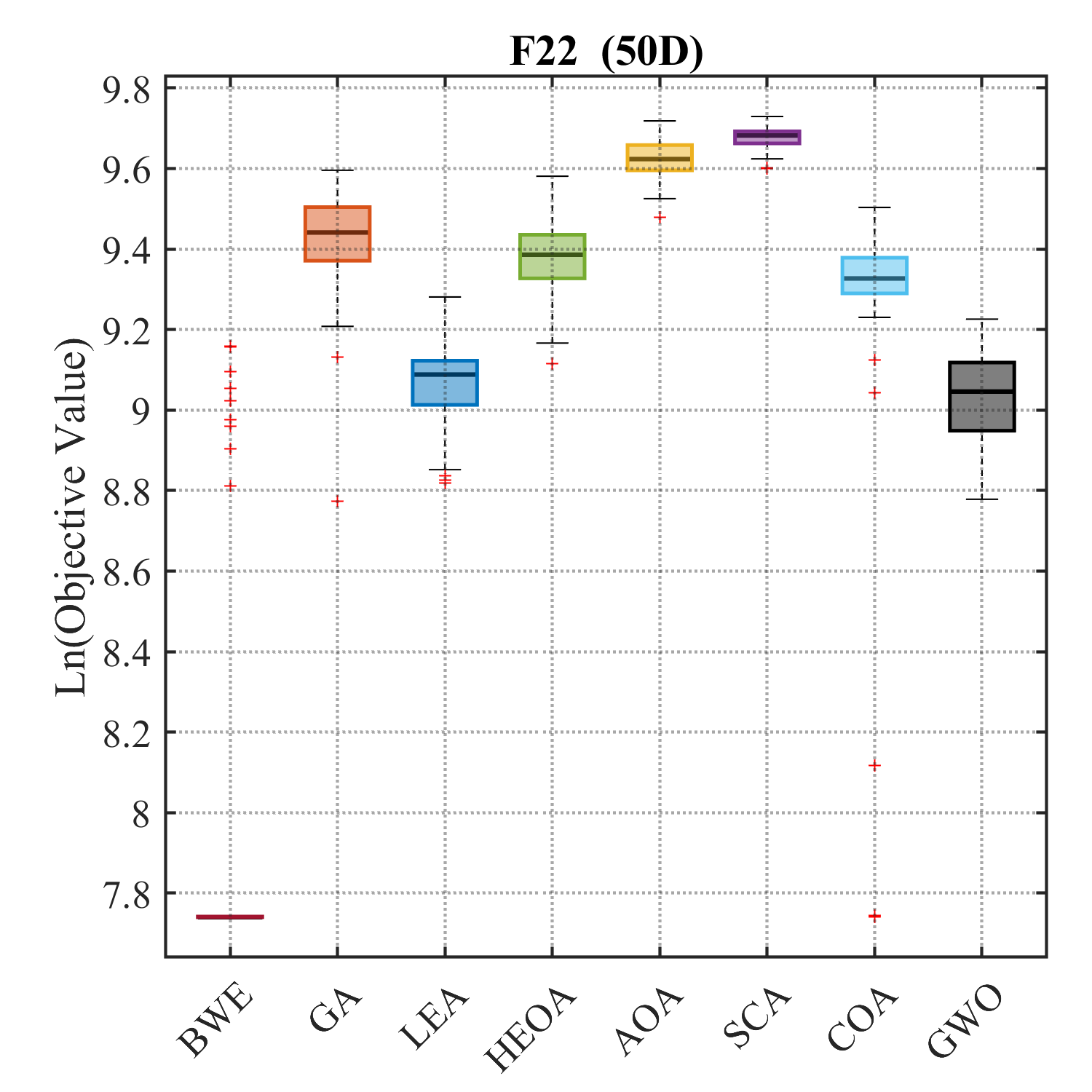}%
    \includegraphics[width=0.23\textwidth]{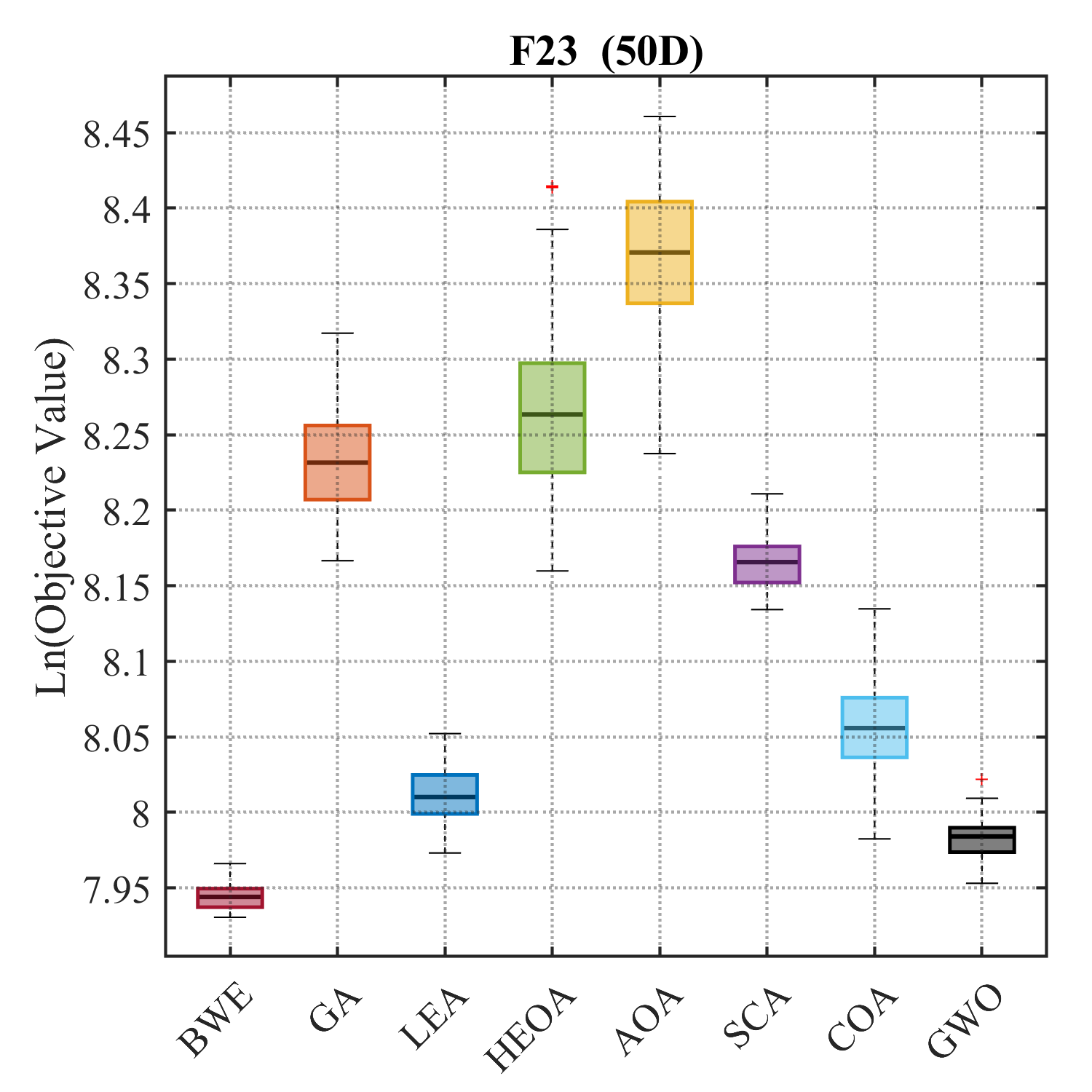}    

    \includegraphics[width=0.23\textwidth]{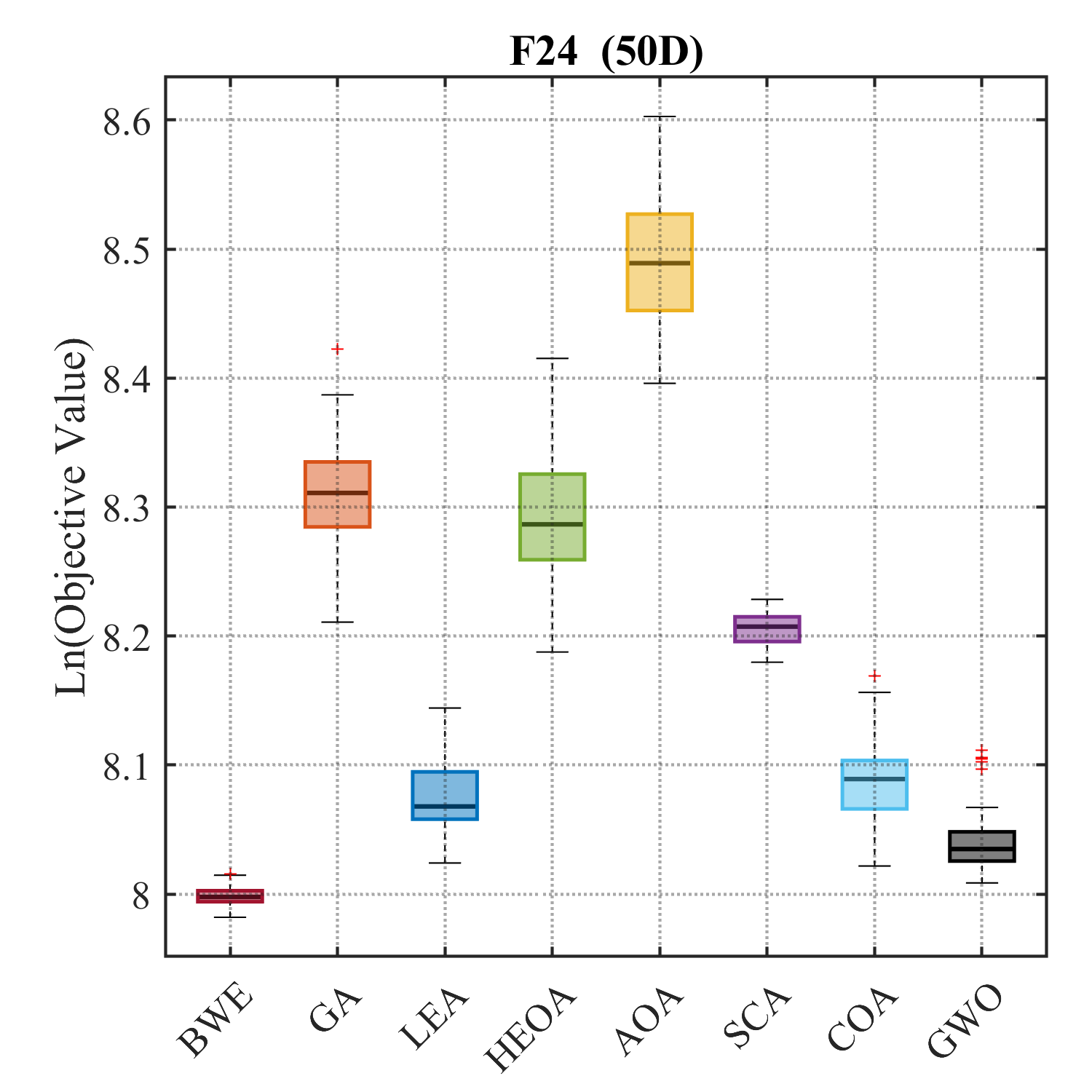}%
    \includegraphics[width=0.23\textwidth]{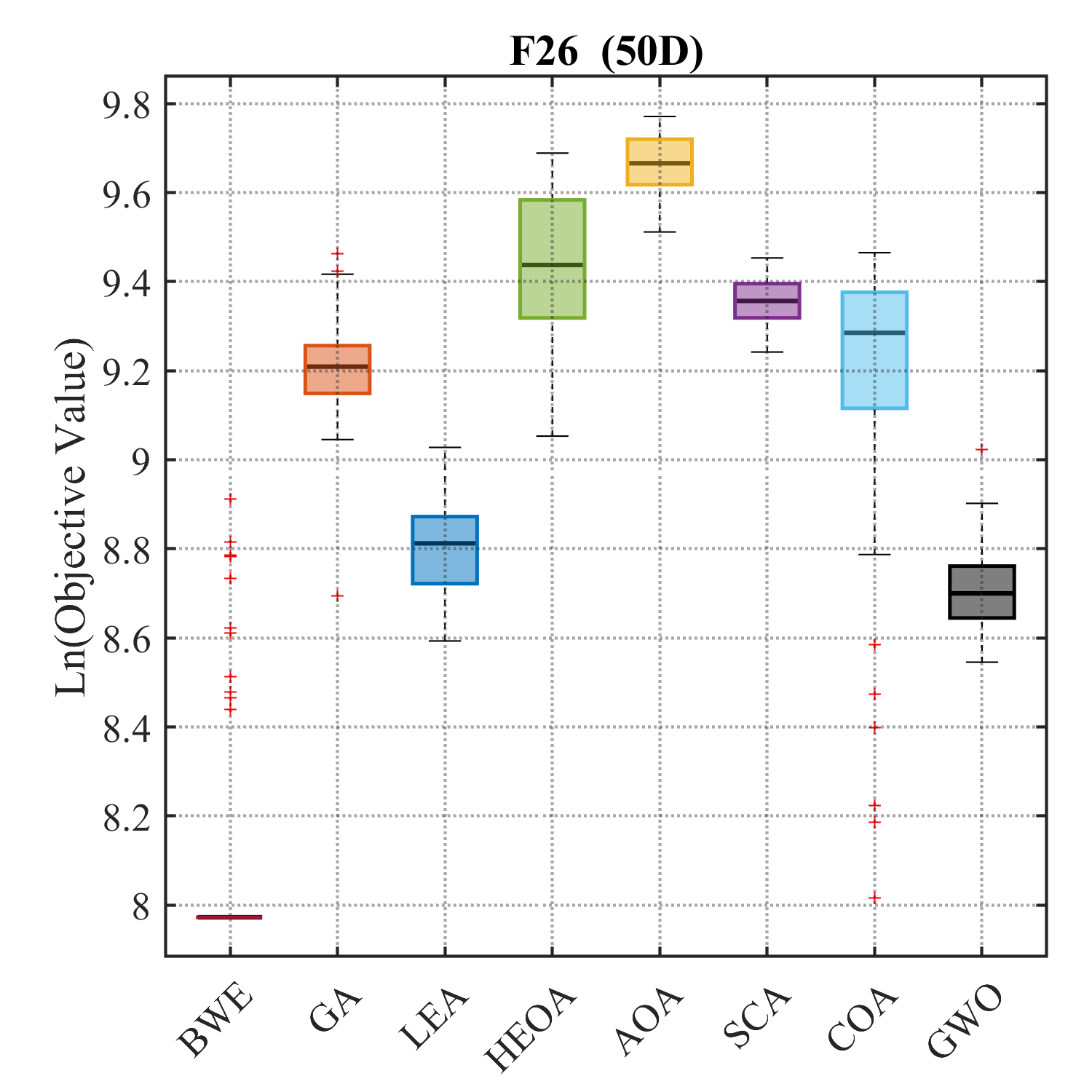}%
    \includegraphics[width=0.23\textwidth]{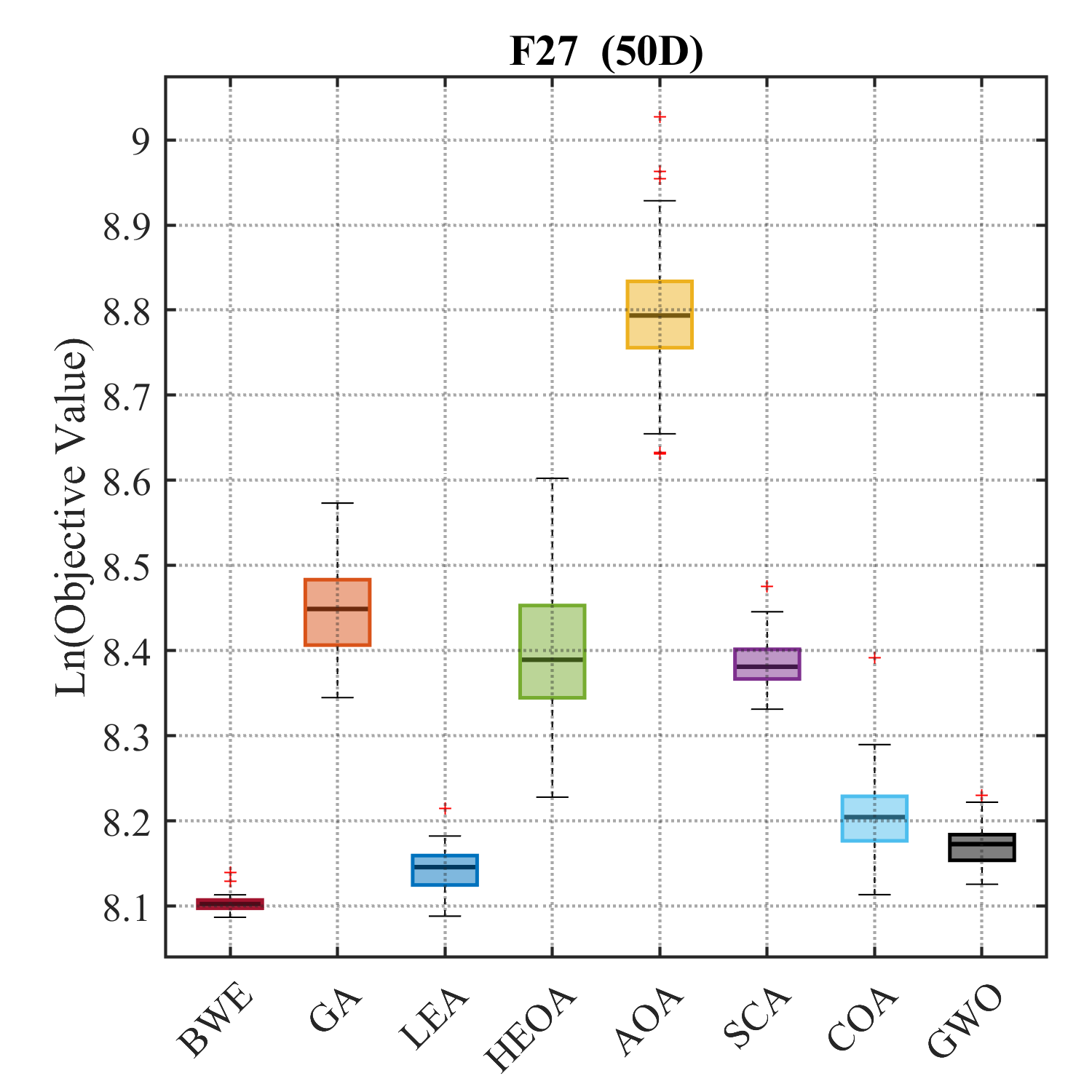}%
    \includegraphics[width=0.23\textwidth]{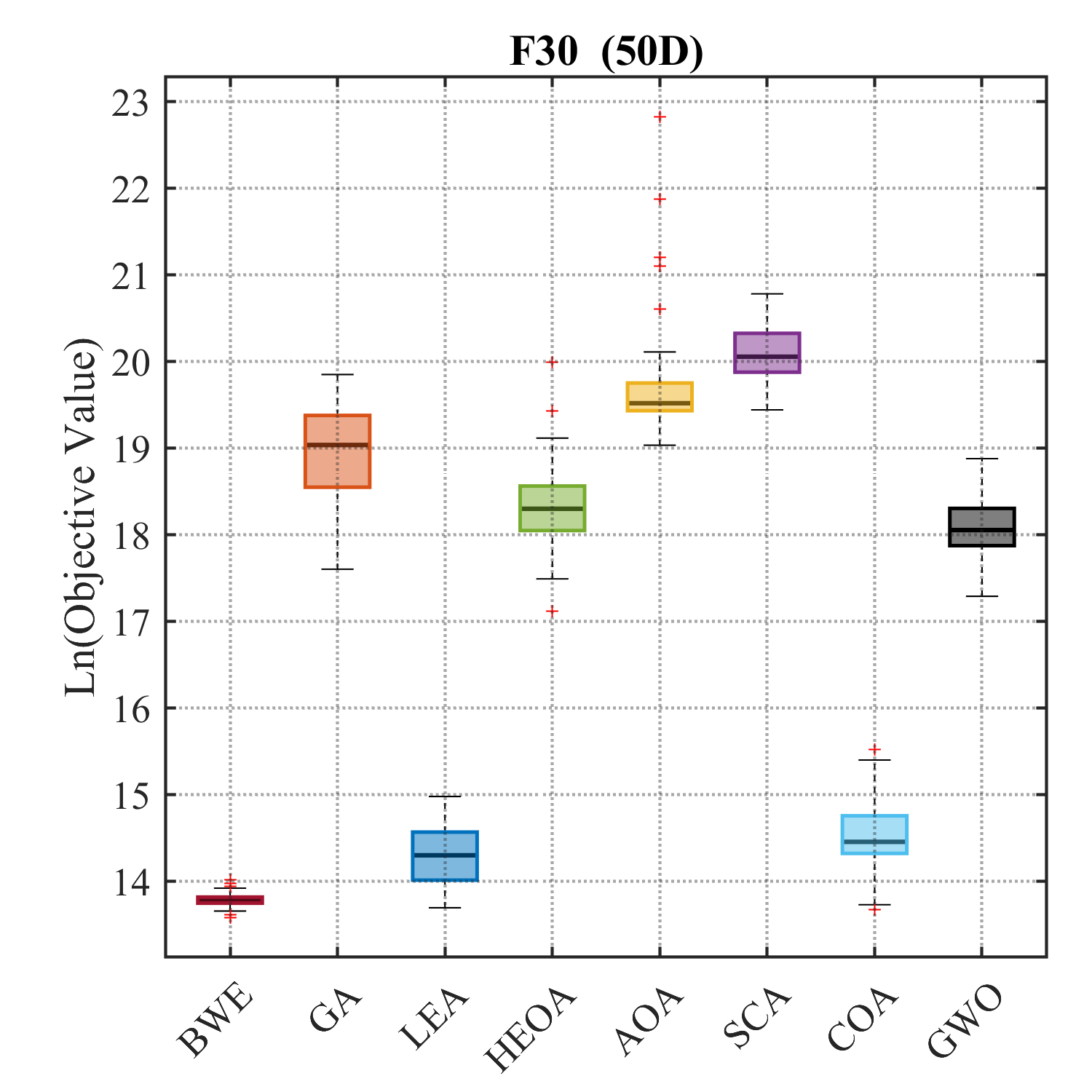}    

    \caption{Box plot of BWE and comparative algorithms on some CEC2017 functions.}
    \label{fig:boxplot_50D_part1}
\end{figure*}

Box plot analyses based on 51 independent runs visually evaluate the distribution, stability, and robustness of the algorithms.
\cref{fig:boxplot_50D_part1} displays representative functions from the CEC2017 suite ($D=50$), where the central line, "+" symbols, and whiskers denote the median, statistical outliers, and distribution spread beyond the first and third quartiles, respectively.

BWE exhibits remarkable performance across most test cases, characterized by compact, low-positioned boxes.
On numerous functions such as F3, F4, F6, and F7, the narrow interquartile ranges (IQRs) achieved by BWE confirm high optimization precision and superior stability.
Even when not the absolute top performer, BWE remains highly competitive; for instance, on F1 and F16, its maximum fitness values approach or fall below the minimum values achieved by most competitors.

Notably, BWE produces significantly fewer outliers than its counterparts in high-dimensional scenarios.
Although its box scale on F10 and F20 is comparable to some competitors, BWE’s median and mean consistently remain at a superior, lower level.
For hybrid and composition, the consistently narrow boxes underscore BWE’s strong structural stability when navigating complex, multimodal landscapes.

Despite its overall dominance, the analysis highlights certain limitations.
On F3, BWE’s statistical distribution is slightly inferior to LEA, indicating sensitivity to specific landscape characteristics.
Furthermore, a noticeably wider IQR on F26 suggests that BWE’s reliability on certain intricate composition functions faces challenges, though it still maintains an acceptable performance margin over the majority of compared metaheuristics.

\subsection{Behavior analysis}
\begin{figure*}[!htbp]
    \centering

    \includegraphics[width=0.25\textwidth]{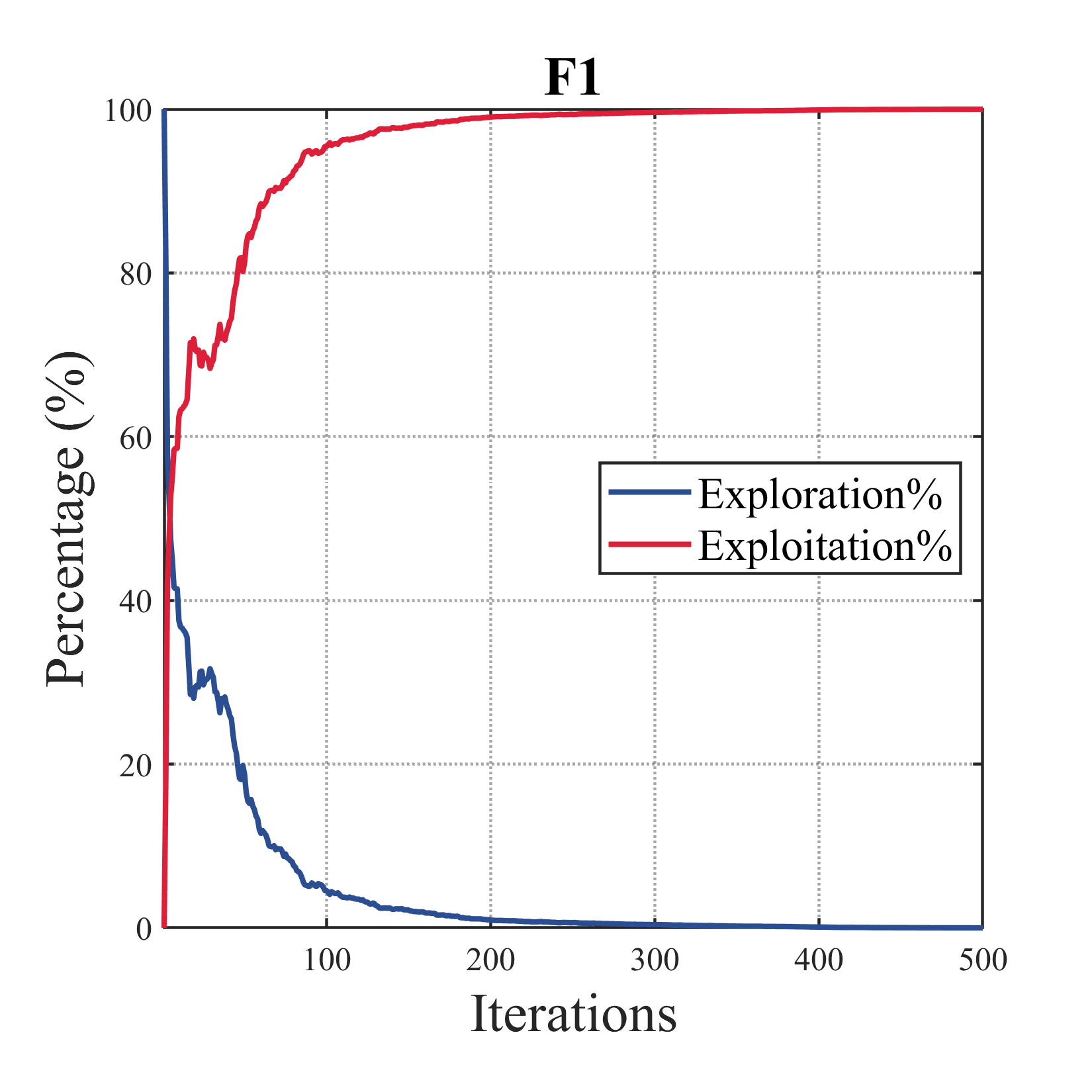}%
    \includegraphics[width=0.25\textwidth]{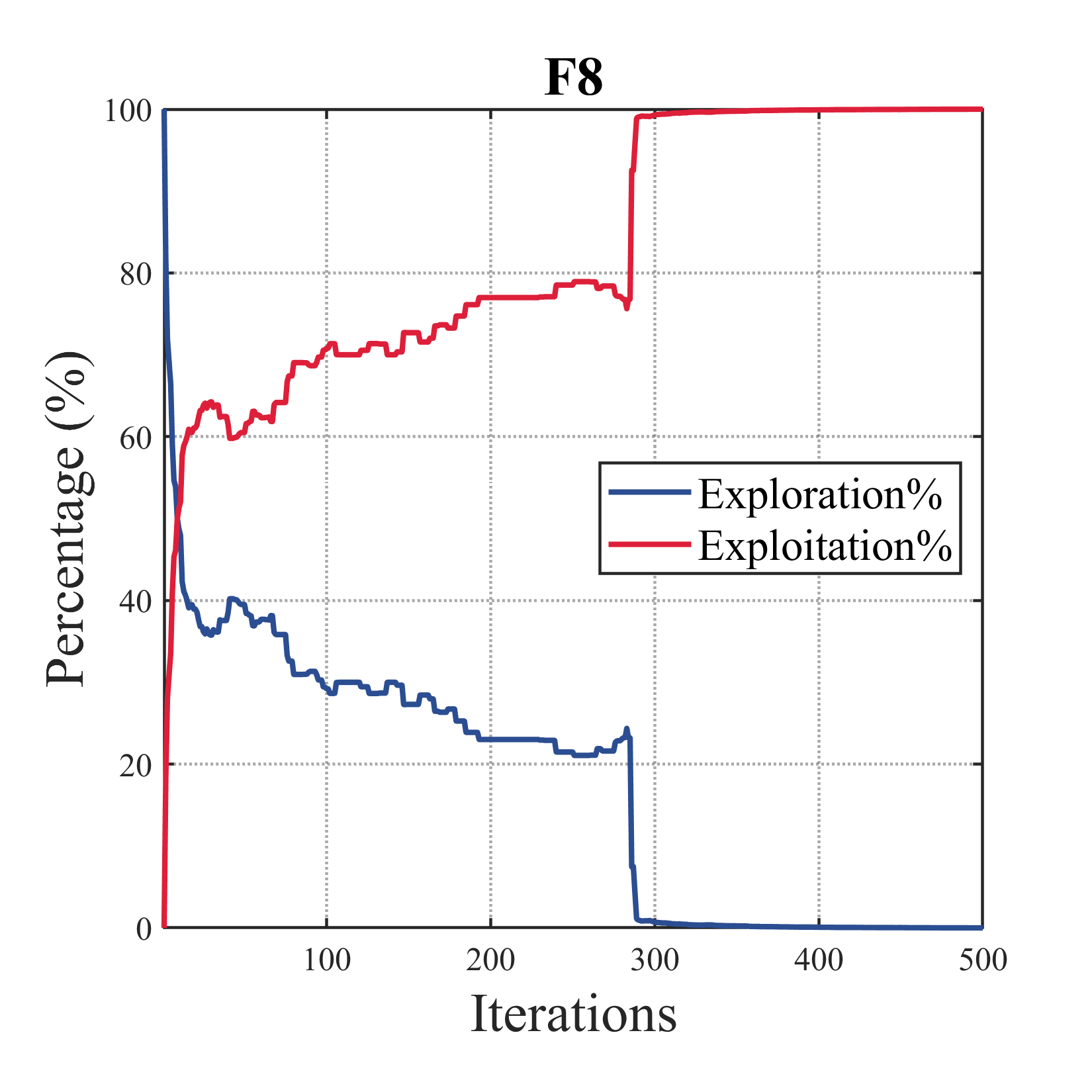}%
    \includegraphics[width=0.25\textwidth]{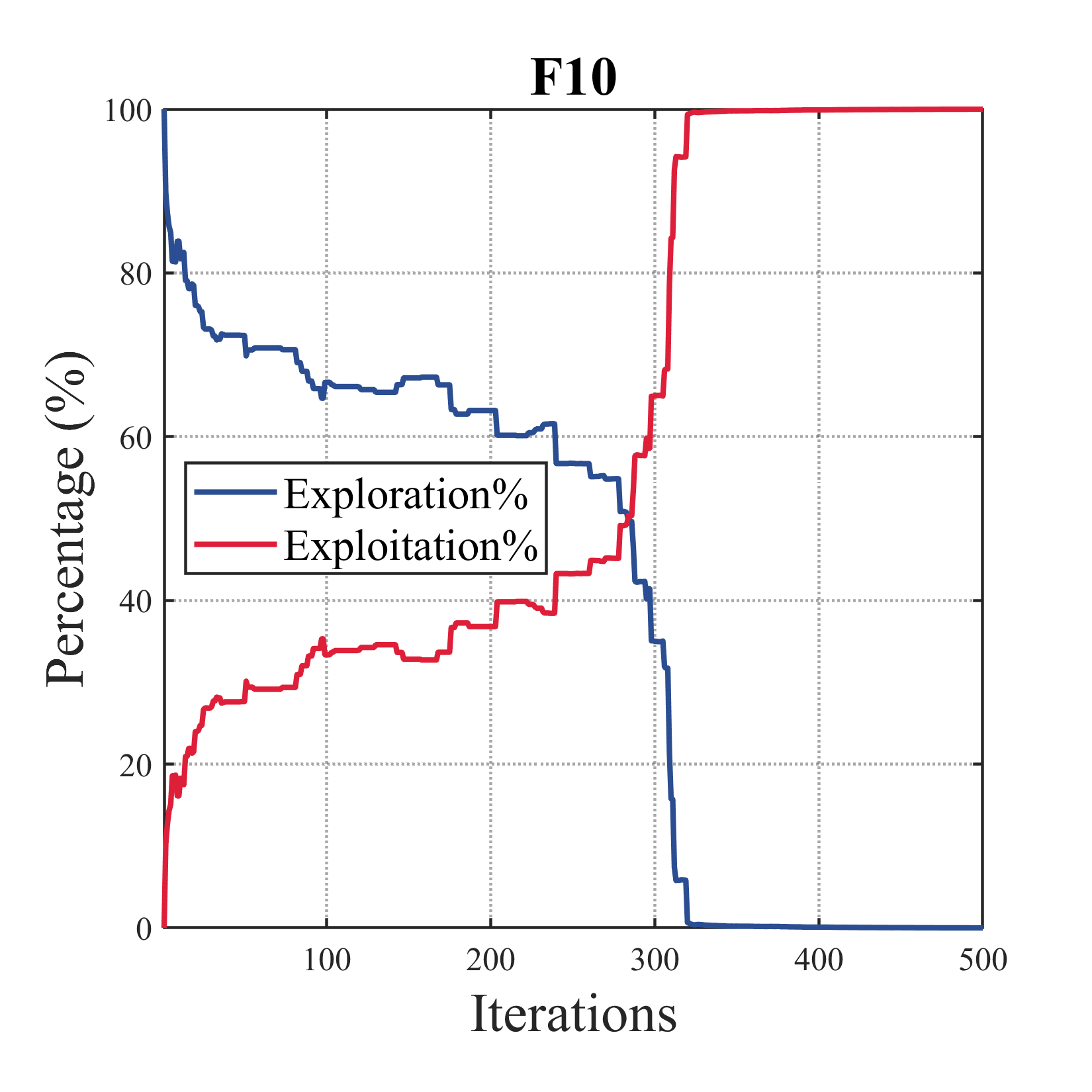}%
    \includegraphics[width=0.25\textwidth]{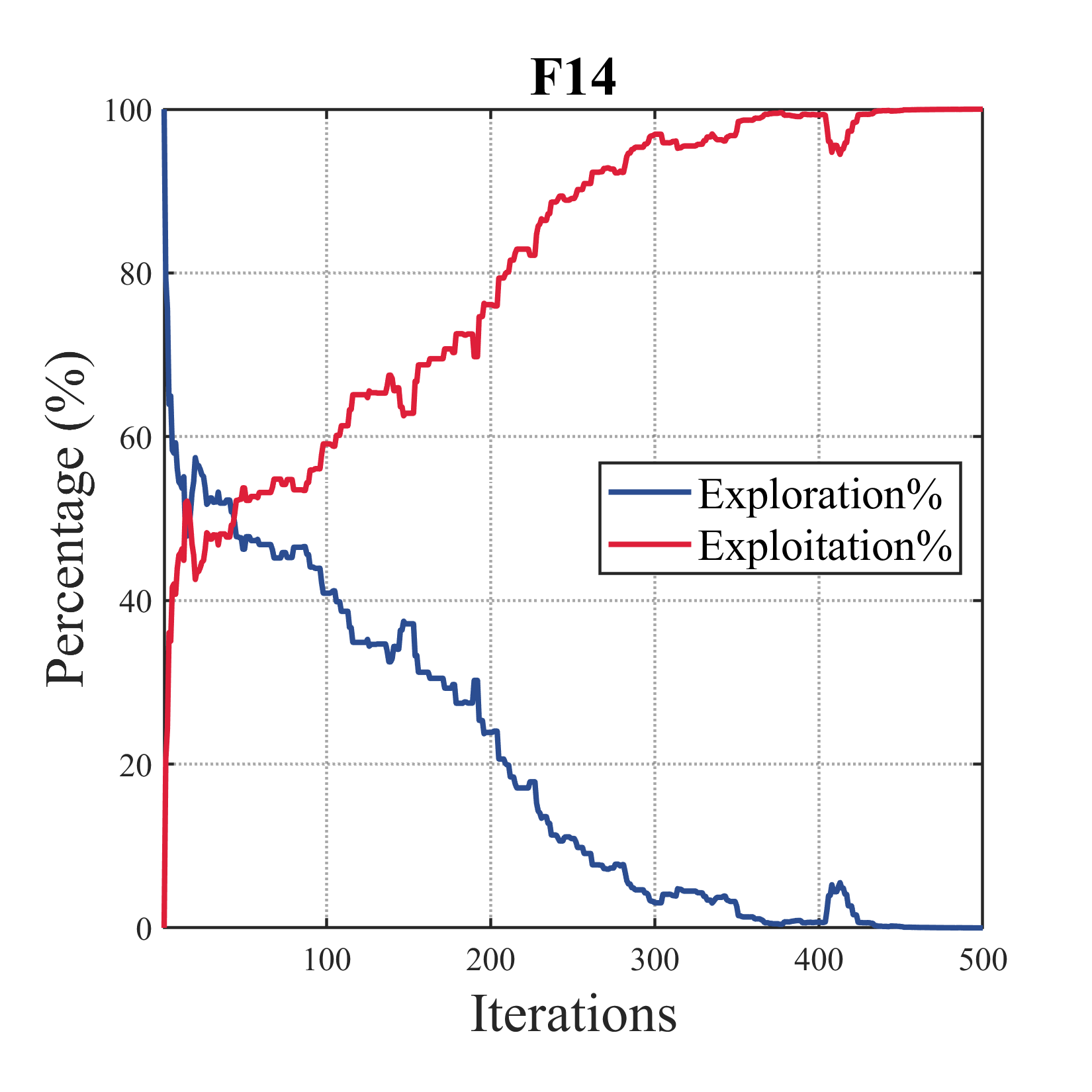}
    
    \includegraphics[width=0.25\textwidth]{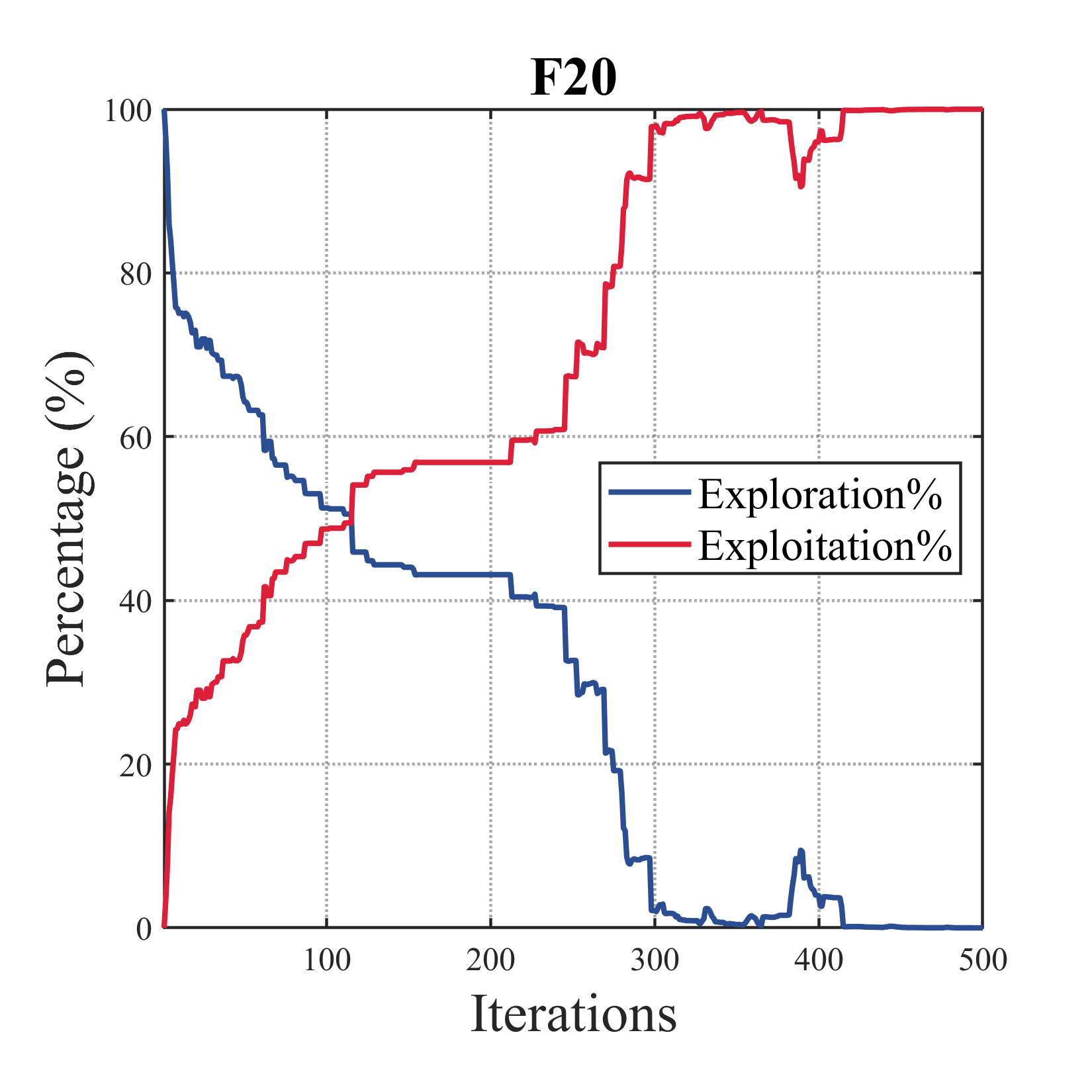}%
    \includegraphics[width=0.25\textwidth]{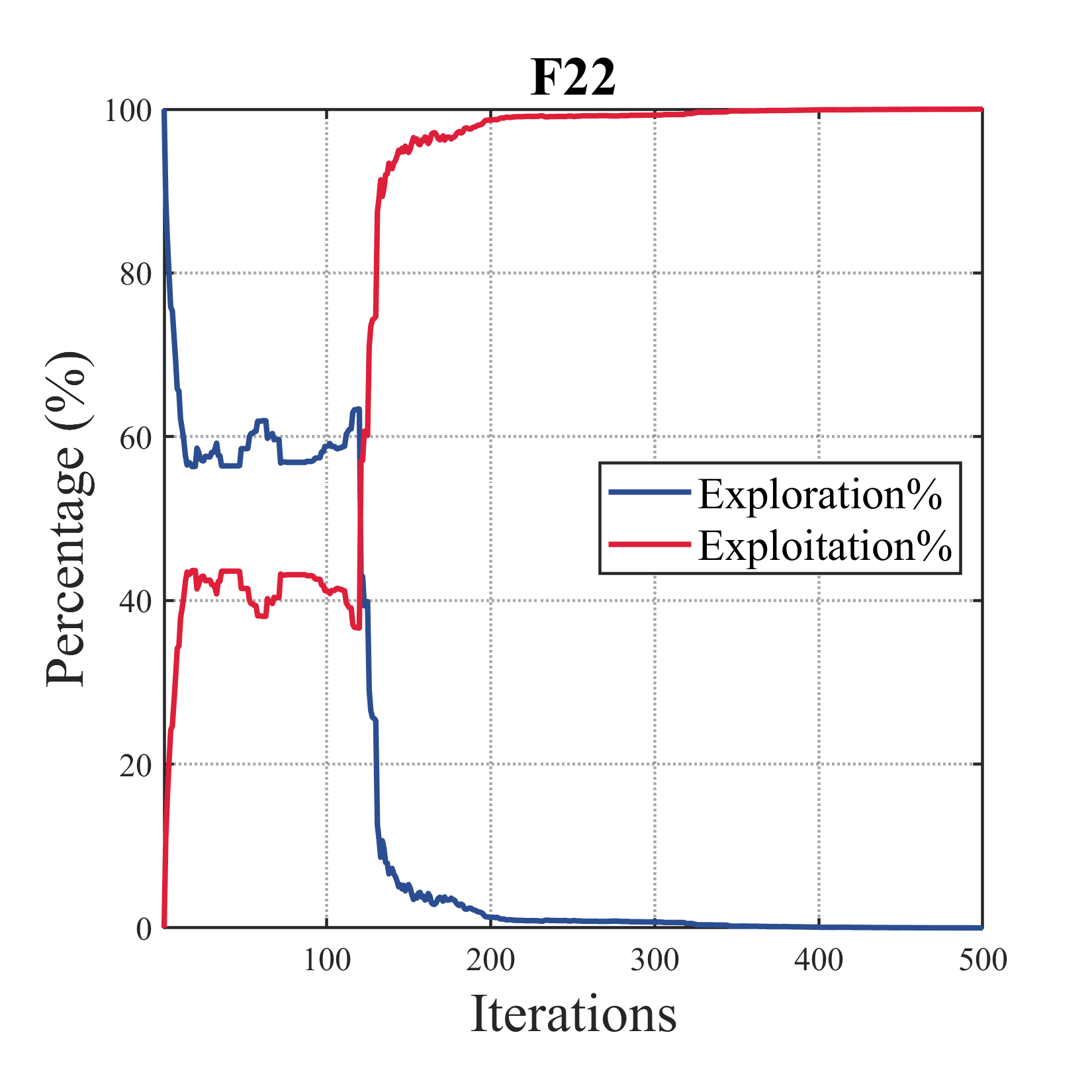}%
    \includegraphics[width=0.25\textwidth]{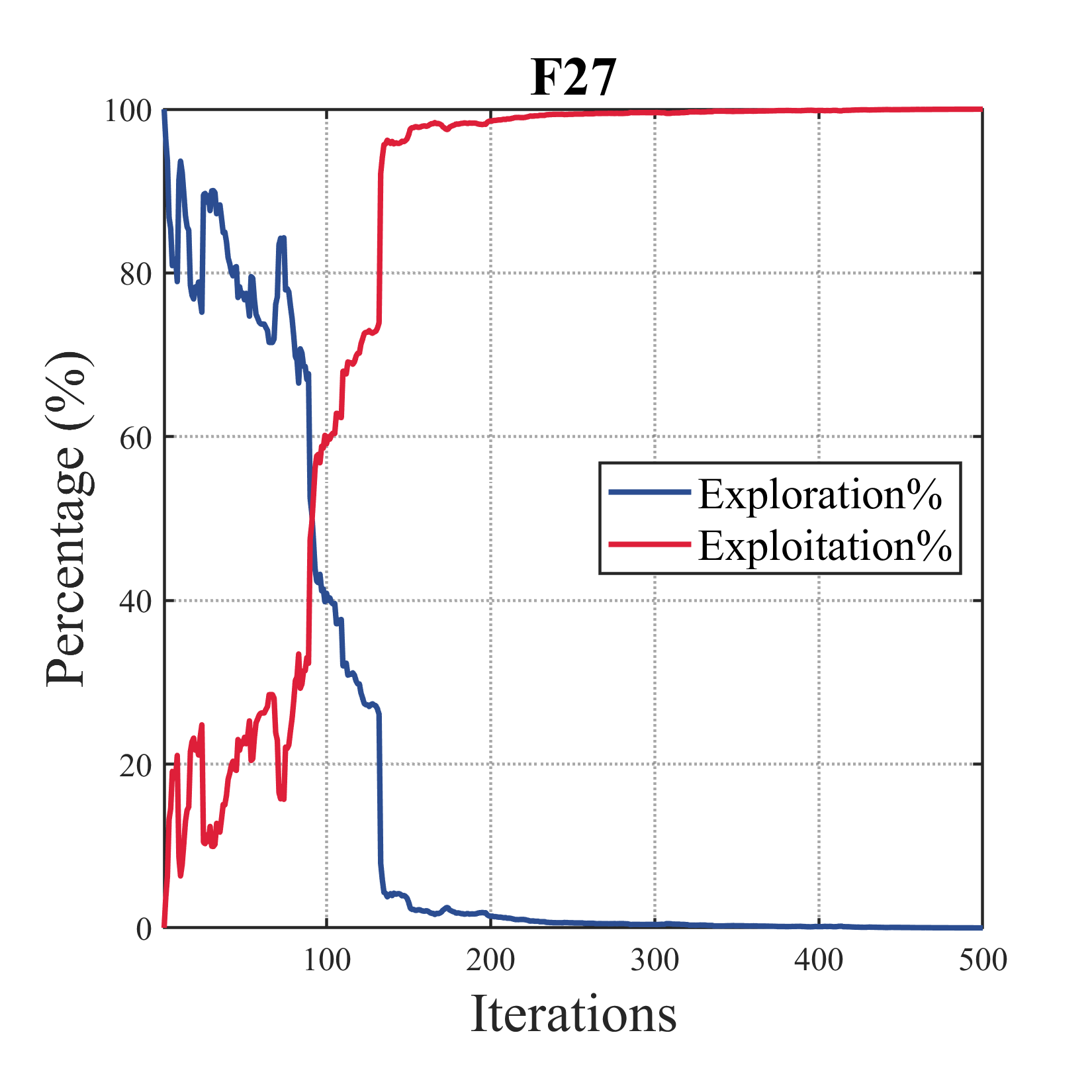}%
    \includegraphics[width=0.25\textwidth]{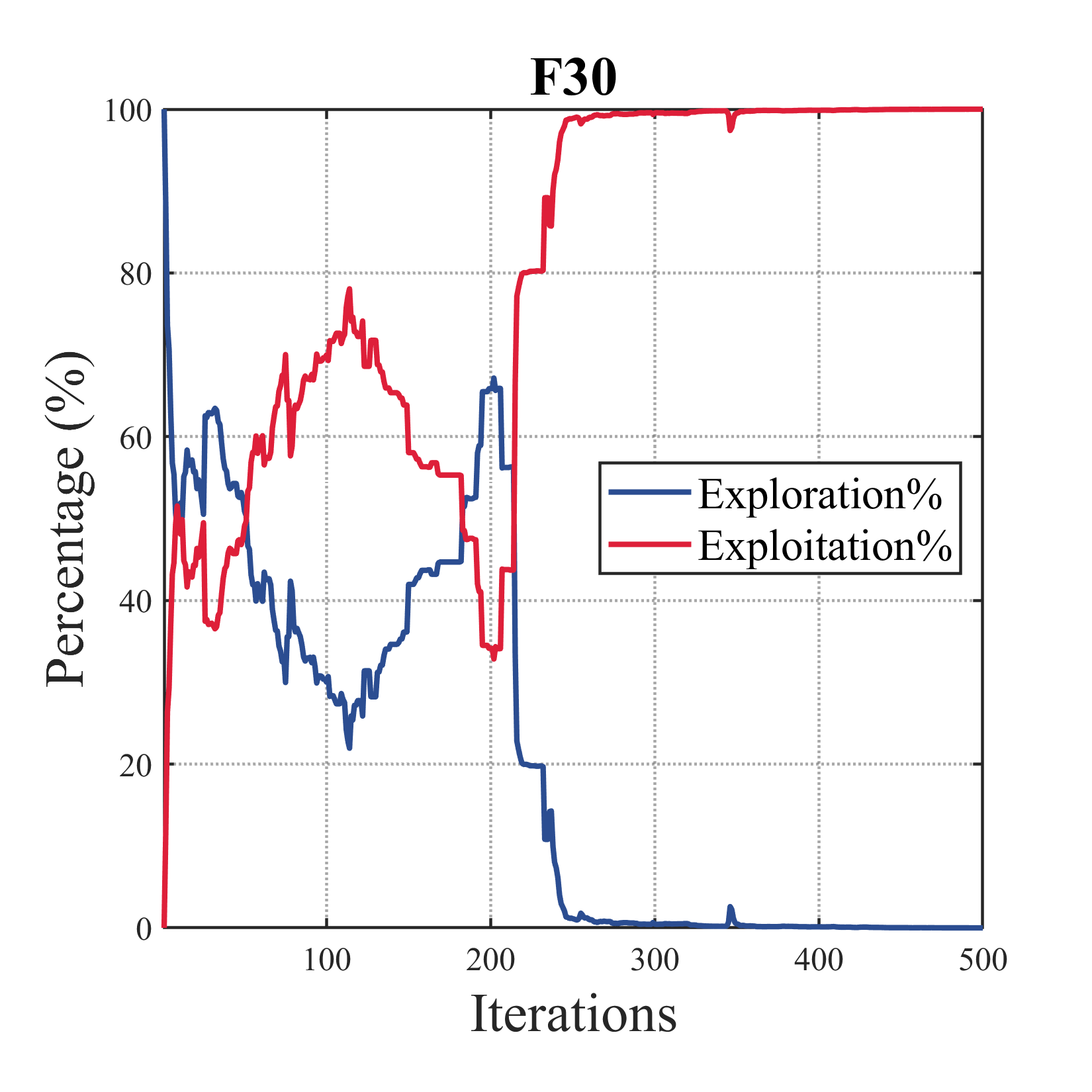}       

    \caption{Percentages of exploration and exploitation.}
    \label{fig:explor_exploi}
\end{figure*}

\cref{fig:explor_exploi} presents the percentage distribution of exploration and exploitation rates of BWE across selected CEC2017 benchmark functions. To illustrate its adaptive behavior, functions from different categories were sampled. The results show that BWE dynamically adjusts the exploration--exploitation balance according to problem characteristics.

Specifically:
(1) On unimodal functions, the exploration rate surpassed the exploitation rate early, exceeding 90\% at around 100 iterations, demonstrating strong exploration capability;
(2) On multimodal functions, although the exploration rate also exceeded 50\% early, it remained significant until roughly half of the iterations. In particular, on F10, a high exploration rate persisted even in later iterations;
(3) On hybrid and composition functions, exploration and exploitation fluctuate more noticeably. For example, on F30, dominance alternated multiple times, indicating BWE’s ability to adaptively balance both behaviors.

\begin{figure*}[!htbp]
\centering

\includegraphics[width=.92\linewidth]{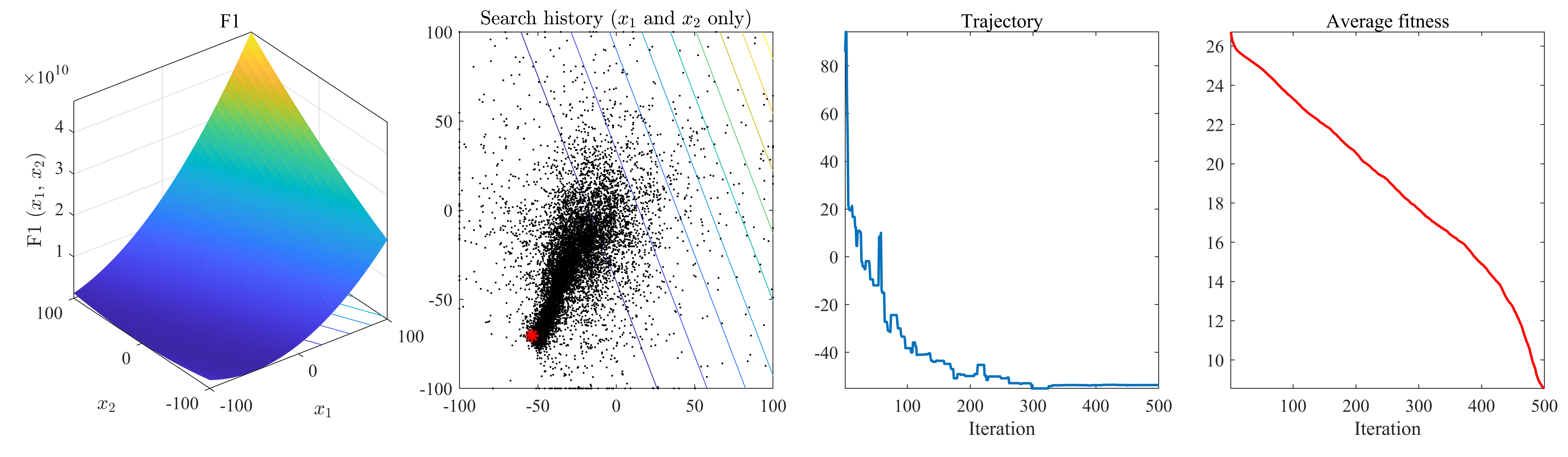}  
\includegraphics[width=.92\linewidth]{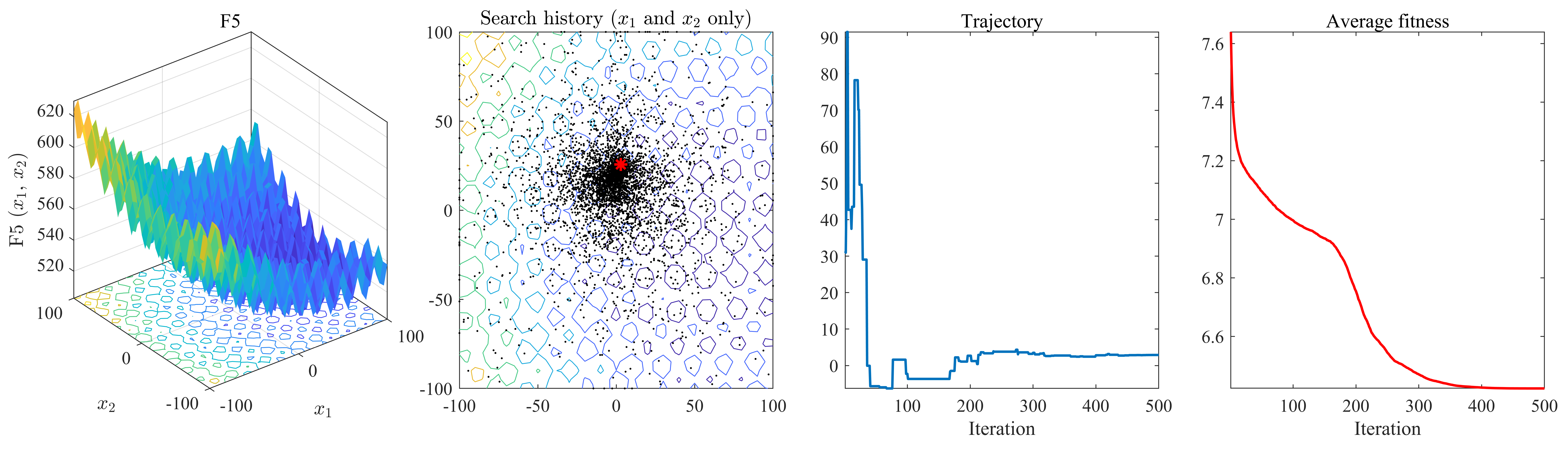}  
\includegraphics[width=.92\linewidth]{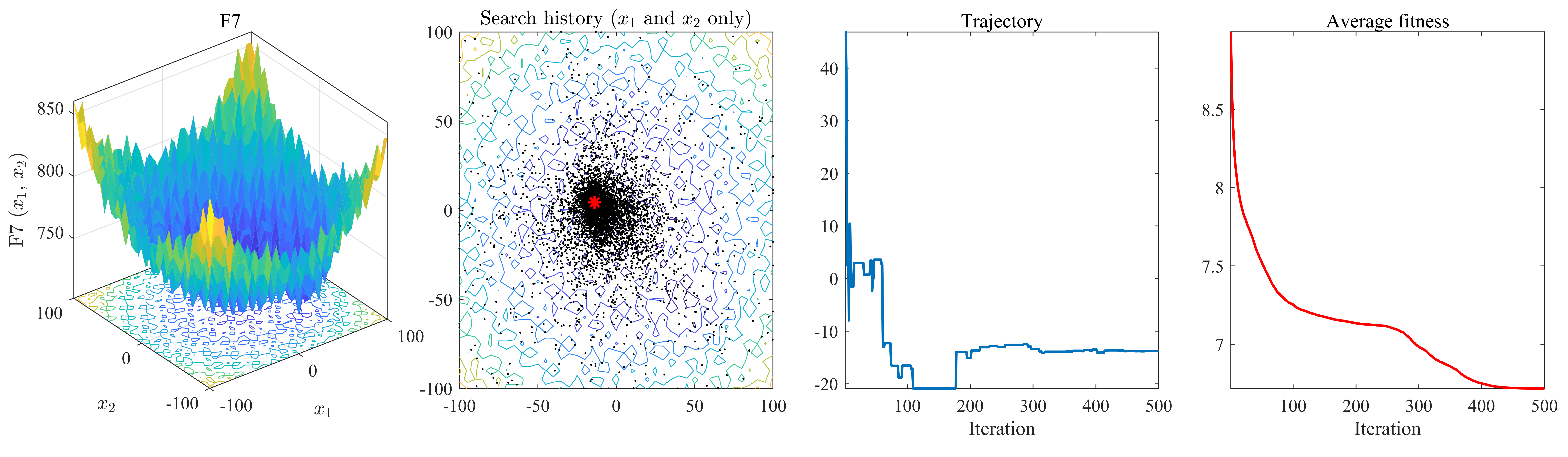}  
\includegraphics[width=.92\linewidth]{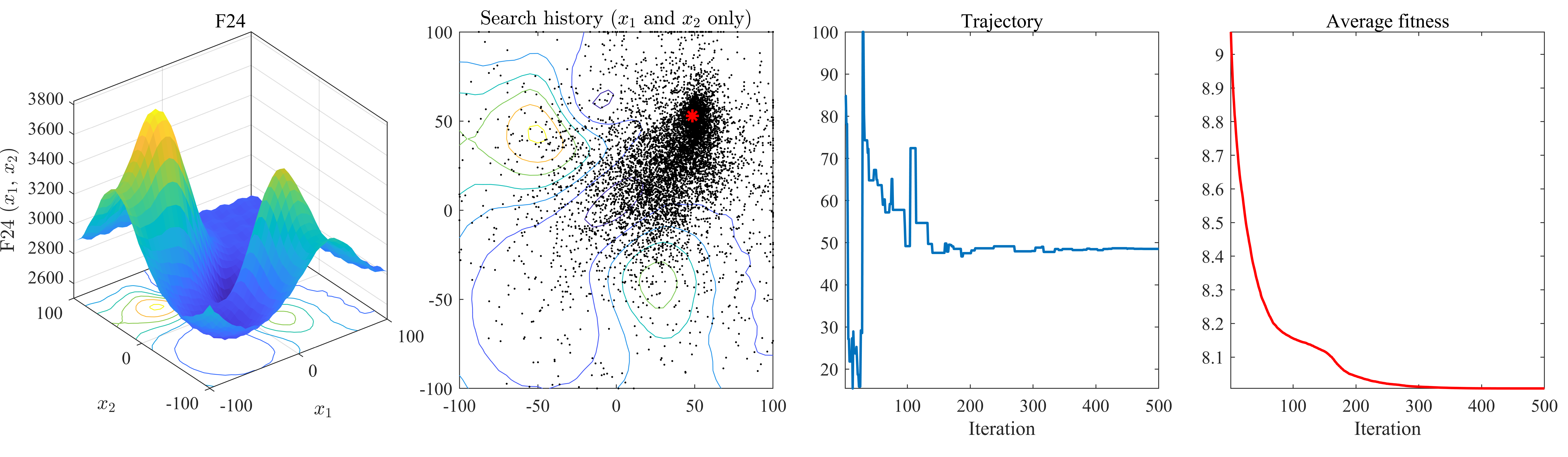}  
\includegraphics[width=.92\linewidth]{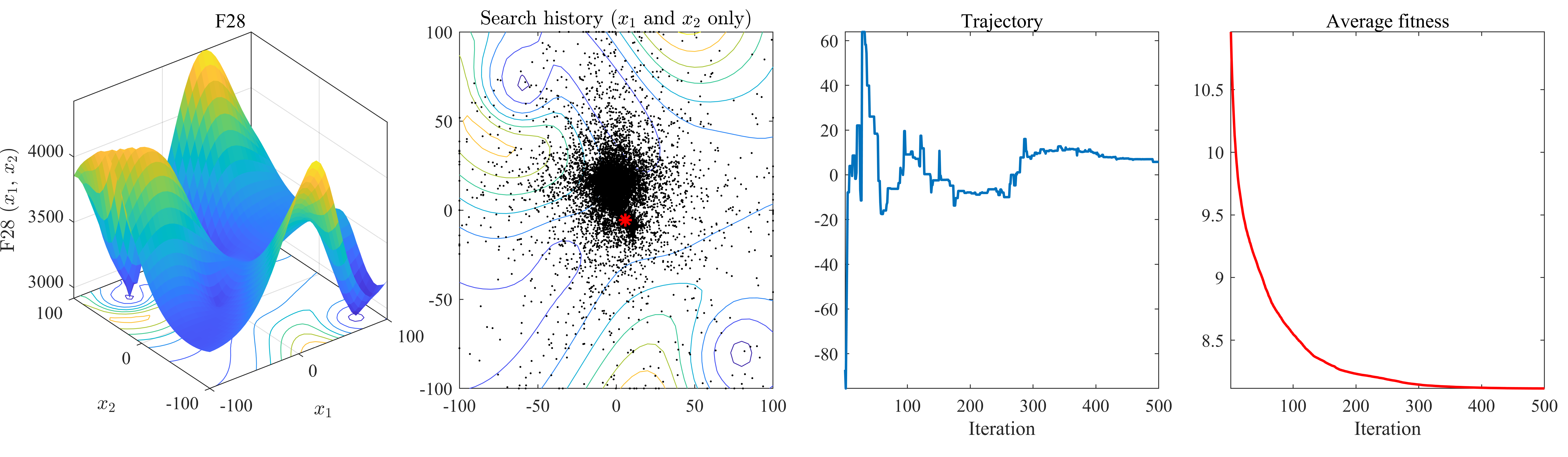}

\caption{Function images, search history, trajectory and average fitness of some functions ($2D$).}
\label{fig:4plots}
\end{figure*}

\cref{fig:4plots} illustrates the optimization process through five representative functions, showing $2D$ function landscapes, search history, trajectories, and mean fitness convergence. Three key observations emerge. First, all search history plots exhibit a globally sparse yet locally dense pattern, reflecting BWE’s adaptive balance between exploration and exploitation. Second, trajectory analysis reveals an initial phase of large-step global exploration, followed by focused local exploitation near promising optima. Third, the rapid convergence of mean fitness across all functions indicates strong optimization capability, with fitness values decreasing sharply as agents exploit promising regions.

For F7, the average fitness drops steeply at around 300 iterations, likely because BWE escapes local optimum. This is supported by the abrupt trajectory change at the same stage, suggesting that BWE retains exploration even in the middle and late stages without sacrificing exploitation.

\subsection{Competitive analysis}
Based on the statistical results in \cref{table:cec2022_10D,table:cec2022_20D} and \cref{tab:cec2022_Statistical}, BWE demonstrates strong optimization performance on the CEC2022 benchmark suite. Across both $10D$ and $20D$, BWE achieves Wilcoxon rank-sum results of nearly 12/0/0 or 11/0/1 against GA, AOA, HEOA, SCA, JADE, and GWO, indicating statistically significant superiority. Even against strong competitors such as COA and LEA, BWE maintains a clear advantage (e.g., 10 wins, 2 ties, and 0 losses against COA in $10D$). In terms of convergence accuracy, BWE often outperforms these algorithms by several to more than ten orders of magnitude, validating the effectiveness of its core search mechanism.

Against industry-standard benchmarks and SOTA algorithms, BWE also remains highly competitive. Compared with CMA-ES, BWE shows better scalability as dimensionality increases, achieving 7 wins, 1 tie, and 4 losses in $20D$, while avoiding the performance degradation of CMA-ES on challenging functions such as C10 and C12. BWE also performs competitively against the L-SHADE family. In the $10D$ Friedman ranking, BWE (Rank: 4.17) surpasses LSHADE-SPACMA (Rank: 4.25), ranking fourth. In the $20D$ tests, a notable result appears on C1, where several SOTA algorithms, including L-SHADE and LSHADE-SPACMA, are trapped in local optima (Mean $>1.63\times10^3$), whereas BWE converges to the theoretical optimum (300). This suggests stronger global exploration capability on deceptive landscapes.

Although BWE still trails LSHADE-cnEpSin in overall ranking, this is expected given its early stage of development. The L-SHADE reflects over a decade of advances in evolutionary computation, incorporating mature strategies such as linear population size reduction and adaptive parameter control. Despite its relatively simple structure, BWE ranks among the top-performing algorithms in $10D$ and remains competitive with LSHADE-SPACMA in $20D$ (5 wins, 1 tie, 6 losses). These results highlight both the effectiveness of BWE’s native mechanism and its potential for further improvement through adaptive enhancement strategies.

\begin{table*}[!htbp]
\centering
\caption{Experimental results of BWE, well-known algorithm, and state-of-the-art DE variants on CEC2022 suite ($10D$).}
\label{table:cec2022_10D}
\sffamily
\scriptsize
\begin{tabular}{llllllllll}
\toprule
Metrics & Mean & Min & Std & Mean & Min & Std & Mean & Min & Std \\ \hline
No. & C1 & & & C2 & & & C3 & & \\ \hline
BWE & 3.00E+02 & 3.00E+02 & 1.04E-12 & 4.02E+02 & 4.00E+02 & 1.29E+01 & 6.00E+02 & 6.00E+02 & 1.97E-06 \\
GA & 3.14E+04 ($>$) & 1.06E+04 & 1.42E+04 & 4.28E+02 ($>$) & 4.02E+02 & 2.85E+01 & 6.15E+02 ($>$) & 6.08E+02 & 4.24E+00 \\
LEA & 3.00E+02 ($>$) & 3.00E+02 & 5.81E-10 & 4.05E+02 ($>$) & 4.00E+02 & 3.62E+00 & 6.00E+02 ($>$) & 6.00E+02 & 9.65E-03 \\
HEOA & 8.25E+02 ($>$) & 3.04E+02 & 5.04E+02 & 4.08E+02 ($>$) & 4.00E+02 & 1.28E+01 & 6.30E+02 ($>$) & 6.07E+02 & 1.20E+01 \\
AOA & 3.00E+02 ($>$) & 3.00E+02 & 3.02E-03 & 4.19E+02 ($>$) & 4.00E+02 & 2.74E+01 & 6.36E+02 ($>$) & 6.22E+02 & 6.78E+00 \\
SCA & 5.07E+02 ($>$) & 3.86E+02 & 5.83E+01 & 4.28E+02 ($>$) & 4.12E+02 & 7.00E+00 & 6.11E+02 ($>$) & 6.07E+02 & 1.96E+00 \\
COA & 3.00E+02 ($>$) & 3.00E+02 & 4.84E-08 & 4.12E+02 ($>$) & 4.00E+02 & 2.08E+01 & 6.04E+02 ($>$) & 6.00E+02 & 1.18E+01 \\
GWO & 9.96E+02 ($>$) & 3.04E+02 & 1.23E+03 & 4.13E+02 ($>$) & 4.00E+02 & 1.33E+01 & 6.01E+02 ($>$) & 6.00E+02 & 1.89E+00 \\
JADE & 3.97E+03 ($>$) & 3.05E+02 & 1.51E+03 & 5.14E+02 ($>$) & 4.04E+02 & 5.20E+01 & 6.26E+02 ($>$) & 6.00E+02 & 8.08E+00 \\
CMA-ES & 3.00E+02 ($<$) & 3.00E+02 & 0.00E+00 & 4.01E+02 ($<$) & 4.00E+02 & 1.38E+00 & 6.59E+02 ($>$) & 6.55E+02 & 1.15E+00 \\
L-SHADE & 3.00E+02 ($<$) & 3.00E+02 & 0.00E+00 & 4.06E+02 ($>$) & 4.04E+02 & 2.48E+00 & 6.00E+02 ($<$) & 6.00E+02 & 2.07E-06 \\
AL-SHADE & 3.00E+02 ($<$) & 3.00E+02 & 0.00E+00 & 4.06E+02 ($>$) & 4.00E+02 & 2.70E+00 & 6.00E+02 ($<$) & 6.00E+02 & 0.00E+00 \\
LSHADE-cnEpSin & 3.00E+02 ($<$) & 3.00E+02 & 0.00E+00 & 4.07E+02 ($>$) & 4.00E+02 & 2.54E+00 & 6.00E+02 ($<$) & 6.00E+02 & 0.00E+00 \\
LSHADE-SPACMA & 3.00E+02 ($>$) & 3.00E+02 & 6.84E-04 & 4.05E+02 ($>$) & 4.00E+02 & 2.90E+00 & 6.00E+02 ($>$) & 6.00E+02 & 1.05E-05 \\ \hline
No. & C4 & & & C5 & & & C6 & & \\ \hline
BWE & 8.06E+02 & 8.03E+02 & 2.10E+00 & 9.00E+02 & 9.00E+02 & 7.58E-13 & 2.22E+03 & 1.81E+03 & 5.77E+02 \\
GA & 8.39E+02 ($>$) & 8.21E+02 & 8.03E+00 & 1.00E+03 ($>$) & 9.02E+02 & 1.88E+02 & 3.91E+03 ($>$) & 1.85E+03 & 2.53E+03 \\
LEA & 8.23E+02 ($>$) & 8.08E+02 & 7.41E+00 & 9.00E+02 ($>$) & 9.00E+02 & 1.16E-01 & 4.48E+03 ($>$) & 1.91E+03 & 2.07E+03 \\
HEOA & 8.34E+02 ($>$) & 8.21E+02 & 9.26E+00 & 1.41E+03 ($>$) & 1.06E+03 & 2.06E+02 & 4.57E+03 ($>$) & 1.91E+03 & 2.12E+03 \\
AOA & 8.30E+02 ($>$) & 8.10E+02 & 9.87E+00 & 1.26E+03 ($>$) & 1.05E+03 & 1.13E+02 & 3.88E+03 ($>$) & 1.94E+03 & 1.61E+03 \\
SCA & 8.26E+02 ($>$) & 8.18E+02 & 4.21E+00 & 9.30E+02 ($>$) & 9.10E+02 & 1.39E+01 & 2.46E+05 ($>$) & 9.86E+03 & 3.41E+05 \\
COA & 8.29E+02 ($>$) & 8.14E+02 & 5.70E+00 & 9.16E+02 ($>$) & 9.00E+02 & 6.29E+01 & 2.45E+03 ($\approx$) & 1.86E+03 & 9.43E+02 \\
GWO & 8.10E+02 ($>$) & 8.02E+02 & 5.08E+00 & 9.04E+02 ($>$) & 9.00E+02 & 7.79E+00 & 5.43E+03 ($>$) & 1.94E+03 & 2.37E+03 \\
JADE & 8.40E+02 ($>$) & 8.17E+02 & 7.46E+00 & 1.21E+03 ($>$) & 9.03E+02 & 1.52E+02 & 1.45E+06 ($>$) & 1.95E+03 & 1.42E+06 \\
CMA-ES & 8.32E+02 ($>$) & 8.30E+02 & 8.15E-01 & 1.46E+03 ($>$) & 1.43E+03 & 1.53E+01 & 1.81E+03 ($<$) & 1.80E+03 & 9.62E+00 \\
L-SHADE & 8.04E+02 ($<$) & 8.02E+02 & 1.84E+00 & 9.00E+02 ($<$) & 9.00E+02 & 0.00E+00 & 1.80E+03 ($<$) & 1.80E+03 & 1.63E+00 \\
AL-SHADE & 8.03E+02 ($<$) & 8.01E+02 & 1.97E+00 & 9.00E+02 ($<$) & 9.00E+02 & 0.00E+00 & 1.80E+03 ($<$) & 1.80E+03 & 2.95E-01 \\
LSHADE-cnEpSin & 8.02E+02 ($<$) & 8.00E+02 & 8.15E-01 & 9.00E+02 ($<$) & 9.00E+02 & 0.00E+00 & 1.80E+03 ($<$) & 1.80E+03 & 1.83E-01 \\
LSHADE-SPACMA & 8.03E+02 ($<$) & 8.00E+02 & 1.71E+00 & 9.00E+02 ($<$) & 9.00E+02 & 0.00E+00 & 1.81E+03 ($<$) & 1.80E+03 & 5.91E+00 \\ \hline
No. & C7 & & & C8 & & & C9 & & \\ \hline
BWE & 2.01E+03 & 2.00E+03 & 7.68E+00 & 2.21E+03 & 2.20E+03 & 1.00E+01 & 2.53E+03 & 2.53E+03 & 2.31E-11 \\
GA & 2.04E+03 ($>$) & 2.03E+03 & 7.86E+00 & 2.23E+03 ($>$) & 2.22E+03 & 1.56E+01 & 2.56E+03 ($>$) & 2.51E+03 & 4.39E+01 \\
LEA & 2.02E+03 ($>$) & 2.00E+03 & 4.94E+00 & 2.22E+03 ($>$) & 2.20E+03 & 8.01E+00 & 2.53E+03 ($>$) & 2.53E+03 & 1.12E-06 \\
HEOA & 2.08E+03 ($>$) & 2.02E+03 & 4.67E+01 & 2.23E+03 ($>$) & 2.21E+03 & 1.34E+01 & 2.58E+03 ($>$) & 2.53E+03 & 3.93E+01 \\
AOA & 2.09E+03 ($>$) & 2.04E+03 & 2.40E+01 & 2.25E+03 ($>$) & 2.22E+03 & 5.36E+01 & 2.54E+03 ($>$) & 2.53E+03 & 3.32E+01 \\
SCA & 2.04E+03 ($>$) & 2.03E+03 & 4.13E+00 & 2.22E+03 ($>$) & 2.21E+03 & 4.53E+00 & 2.54E+03 ($>$) & 2.53E+03 & 5.13E+00 \\
COA & 2.01E+03 ($>$) & 2.00E+03 & 9.24E+00 & 2.21E+03 ($>$) & 2.20E+03 & 9.97E+00 & 2.53E+03 ($>$) & 2.53E+03 & 5.95E-10 \\
GWO & 2.03E+03 ($>$) & 2.00E+03 & 9.62E+00 & 2.22E+03 ($>$) & 2.20E+03 & 6.85E+00 & 2.54E+03 ($>$) & 2.53E+03 & 1.96E+01 \\
JADE & 2.06E+03 ($>$) & 2.02E+03 & 1.09E+01 & 2.23E+03 ($>$) & 2.22E+03 & 4.56E+00 & 2.60E+03 ($>$) & 2.53E+03 & 3.49E+01 \\
CMA-ES & 2.26E+03 ($>$) & 2.04E+03 & 1.24E+02 & 2.24E+03 ($>$) & 2.21E+03 & 4.16E+01 & 2.50E+03 ($<$) & 2.49E+03 & 4.84E+01 \\
L-SHADE & 2.00E+03 ($<$) & 2.00E+03 & 2.55E-01 & 2.21E+03 ($\approx$) & 2.20E+03 & 7.83E+00 & 2.53E+03 ($<$) & 2.53E+03 & 0.00E+00 \\
AL-SHADE & 2.00E+03 ($<$) & 2.00E+03 & 2.11E-01 & 2.20E+03 ($<$) & 2.20E+03 & 6.06E+00 & 2.53E+03 ($<$) & 2.53E+03 & 0.00E+00 \\
LSHADE-cnEpSin & 2.00E+03 ($<$) & 2.00E+03 & 1.58E-01 & 2.20E+03 ($<$) & 2.20E+03 & 8.11E+00 & 2.52E+03 ($<$) & 2.51E+03 & 4.38E+00 \\
LSHADE-SPACMA & 2.00E+03 ($<$) & 2.00E+03 & 2.15E-01 & 2.21E+03 ($<$) & 2.20E+03 & 8.59E+00 & 2.53E+03 ($<$) & 2.53E+03 & 0.00E+00 \\ \hline
No. & C10 & & & C11 & & & C12 & & \\ \hline
BWE & 2.50E+03 & 2.50E+03 & 3.95E-02 & 2.60E+03 & 2.60E+03 & 5.87E-07 & 2.86E+03 & 2.86E+03 & 1.23E+00 \\
GA & 2.55E+03 ($>$) & 2.50E+03 & 5.46E+01 & 2.89E+03 ($>$) & 2.71E+03 & 1.48E+02 & 2.88E+03 ($>$) & 2.85E+03 & 1.55E+01 \\
LEA & 2.50E+03 ($>$) & 2.50E+03 & 5.71E-02 & 2.61E+03 ($>$) & 2.60E+03 & 2.75E+01 & 2.86E+03 ($<$) & 2.86E+03 & 1.34E+00 \\
HEOA & 2.63E+03 ($>$) & 2.50E+03 & 2.78E+01 & 2.72E+03 ($>$) & 2.60E+03 & 1.36E+02 & 2.93E+03 ($>$) & 2.87E+03 & 6.70E+01 \\
AOA & 2.61E+03 ($>$) & 2.50E+03 & 1.12E+02 & 2.68E+03 ($>$) & 2.60E+03 & 1.25E+02 & 2.96E+03 ($>$) & 2.90E+03 & 5.06E+01 \\
SCA & 2.50E+03 ($>$) & 2.50E+03 & 1.53E-01 & 2.74E+03 ($>$) & 2.70E+03 & 1.72E+01 & 2.87E+03 ($>$) & 2.86E+03 & 1.05E+00 \\
COA & 2.54E+03 ($>$) & 2.41E+03 & 6.23E+01 & 2.74E+03 ($>$) & 2.60E+03 & 1.40E+02 & 2.87E+03 ($\approx$) & 2.86E+03 & 2.18E+00 \\
GWO & 2.56E+03 ($>$) & 2.50E+03 & 5.70E+01 & 2.75E+03 ($>$) & 2.60E+03 & 1.82E+02 & 2.86E+03 ($<$) & 2.86E+03 & 2.54E+00 \\
JADE & 2.51E+03 ($>$) & 2.50E+03 & 6.01E+00 & 2.85E+03 ($>$) & 2.60E+03 & 9.46E+01 & 2.90E+03 ($>$) & 2.89E+03 & 1.07E+01 \\
CMA-ES & 2.75E+03 ($>$) & 2.61E+03 & 3.10E+02 & 2.77E+03 ($\approx$) & 2.60E+03 & 1.44E+02 & 2.87E+03 ($\approx$) & 2.85E+03 & 2.64E+01 \\
L-SHADE & 2.50E+03 ($>$) & 2.50E+03 & 1.92E+01 & 2.61E+03 ($>$) & 2.60E+03 & 7.30E+01 & 2.86E+03 ($<$) & 2.86E+03 & 1.89E+00 \\
AL-SHADE & 2.52E+03 ($>$) & 2.50E+03 & 3.99E+01 & 2.61E+03 ($>$) & 2.60E+03 & 3.82E+01 & 2.86E+03 ($<$) & 2.86E+03 & 1.19E+00 \\
LSHADE-cnEpSin & 2.50E+03 ($\approx$) & 2.50E+03 & 4.52E-02 & 2.60E+03 ($<$) & 2.60E+03 & 0.00E+00 & 2.86E+03 ($<$) & 2.85E+03 & 3.33E+00 \\
LSHADE-SPACMA & 2.50E+03 ($>$) & 2.50E+03 & 5.16E-02 & 2.61E+03 ($>$) & 2.60E+03 & 2.75E+01 & 2.86E+03 ($<$) & 2.86E+03 & 1.04E+00 \\ \bottomrule
\end{tabular}
\end{table*}

\begin{table*}[!htbp]
\centering
\caption{Experimental results of BWE, well-known algorithm, and state-of-the-art DE variants on CEC2022 suite ($20D$).}
\label{table:cec2022_20D}
\sffamily
\scriptsize
\begin{tabular}{llllllllll}
\toprule
Metrics & Mean & Min & Std & Mean & Min & Std & Mean & Min & Std \\ \hline
No. & C1 & & & C2 & & & C3 & & \\ \hline
BWE & 3.00E+02 & 3.00E+02 & 1.69E-09 & 4.43E+02 & 4.00E+02 & 1.70E+01 & 6.00E+02 & 6.00E+02 & 1.32E-04 \\
GA & 3.14E+04 ($>$) & 1.06E+04 & 1.42E+04 & 4.28E+02 ($>$) & 4.02E+02 & 2.85E+01 & 6.15E+02 ($>$) & 6.08E+02 & 4.24E+00 \\
LEA & 3.00E+02 ($>$) & 3.00E+02 & 1.39E-06 & 4.33E+02 ($\approx$) & 4.00E+02 & 2.38E+01 & 6.01E+02 ($>$) & 6.00E+02 & 7.20E-01 \\
HEOA & 2.47E+04 ($>$) & 1.21E+04 & 1.04E+04 & 5.10E+02 ($>$) & 4.41E+02 & 4.44E+01 & 6.53E+02 ($>$) & 6.37E+02 & 8.41E+00 \\
AOA & 1.71E+04 ($>$) & 7.48E+03 & 6.05E+03 & 7.08E+02 ($>$) & 5.42E+02 & 1.27E+02 & 6.52E+02 ($>$) & 6.33E+02 & 7.86E+00 \\
SCA & 5.00E+03 ($>$) & 3.26E+03 & 1.51E+03 & 5.88E+02 ($>$) & 5.35E+02 & 3.18E+01 & 6.31E+02 ($>$) & 6.25E+02 & 3.80E+00 \\
COA & 3.00E+02 ($>$) & 3.00E+02 & 1.71E-04 & 4.40E+02 ($>$) & 4.00E+02 & 2.16E+01 & 6.13E+02 ($>$) & 6.01E+02 & 1.42E+01 \\
GWO & 6.02E+03 ($>$) & 7.02E+02 & 2.96E+03 & 4.79E+02 ($>$) & 4.45E+02 & 2.64E+01 & 6.02E+02 ($>$) & 6.00E+02 & 2.05E+00 \\
JADE & 2.75E+04 ($>$) & 1.52E+04 & 5.33E+03 & 1.41E+03 ($>$) & 4.49E+02 & 4.10E+02 & 6.57E+02 ($>$) & 6.00E+02 & 2.04E+01 \\
CMA-ES & 3.00E+02 ($<$) & 3.00E+02 & 3.95E-14 & 4.00E+02 ($<$) & 4.00E+02 & 7.28E-01 & 6.68E+02 ($>$) & 6.67E+02 & 6.11E-01 \\
L-SHADE & 1.63E+03 ($>$) & 3.00E+02 & 3.94E+03 & 4.48E+02 ($>$) & 4.45E+02 & 1.70E+00 & 6.00E+02 ($>$) & 6.00E+02 & 1.18E-02 \\
AL-SHADE & 3.00E+02 ($<$) & 3.00E+02 & 3.17E-14 & 4.48E+02 ($>$) & 4.45E+02 & 1.80E+00 & 6.00E+02 ($\approx$) & 6.00E+02 & 3.43E-02 \\
LSHADE-cnEpSin & 3.00E+02 ($<$) & 3.00E+02 & 0.00E+00 & 4.04E+02 ($<$) & 4.00E+02 & 2.15E+00 & 6.00E+02 ($<$) & 6.00E+02 & 2.99E-14 \\
LSHADE-SPACMA & 2.03E+03 ($>$) & 3.00E+02 & 3.58E+03 & 4.46E+02 ($>$) & 4.00E+02 & 1.24E+01 & 6.00E+02 ($>$) & 6.00E+02 & 2.99E-03 \\ \hline
No. & C4 & & & C5 & & & C6 & & \\ \hline
BWE & 8.28E+02 & 8.16E+02 & 6.99E+00 & 9.00E+02 & 9.00E+02 & 2.27E-02 & 3.06E+03 & 1.99E+03 & 7.64E+02 \\
GA & 8.39E+02 ($>$) & 8.21E+02 & 8.03E+00 & 1.00E+03 ($>$) & 9.02E+02 & 1.88E+02 & 3.91E+03 ($>$) & 1.85E+03 & 2.53E+03 \\
LEA & 8.83E+02 ($>$) & 8.39E+02 & 2.73E+01 & 1.03E+03 ($>$) & 9.01E+02 & 2.22E+02 & 8.98E+03 ($>$) & 2.08E+03 & 7.30E+03 \\
HEOA & 9.06E+02 ($>$) & 8.72E+02 & 1.91E+01 & 3.11E+03 ($>$) & 2.31E+03 & 4.16E+02 & 9.10E+03 ($>$) & 2.04E+03 & 1.03E+04 \\
AOA & 8.84E+02 ($>$) & 8.56E+02 & 1.49E+01 & 2.27E+03 ($>$) & 1.77E+03 & 2.81E+02 & 6.14E+03 ($>$) & 3.58E+03 & 2.44E+03 \\
SCA & 9.17E+02 ($>$) & 8.92E+02 & 9.58E+00 & 1.63E+03 ($>$) & 1.33E+03 & 2.93E+02 & 4.17E+07 ($>$) & 4.09E+06 & 3.01E+07 \\
COA & 8.75E+02 ($>$) & 8.29E+02 & 1.89E+01 & 1.69E+03 ($>$) & 9.09E+02 & 6.33E+02 & 5.84E+03 ($>$) & 1.99E+03 & 5.44E+03 \\
GWO & 8.43E+02 ($>$) & 8.23E+02 & 1.65E+01 & 1.03E+03 ($>$) & 9.02E+02 & 1.28E+02 & 7.89E+05 ($>$) & 1.98E+03 & 3.33E+06 \\
JADE & 9.43E+02 ($>$) & 8.50E+02 & 3.94E+01 & 3.60E+03 ($>$) & 1.05E+03 & 9.69E+02 & 2.55E+08 ($>$) & 2.79E+03 & 1.60E+08 \\
CMA-ES & 8.89E+02 ($>$) & 8.84E+02 & 3.48E+00 & 2.47E+03 ($>$) & 2.41E+03 & 1.76E+01 & 1.84E+03 ($<$) & 1.81E+03 & 1.47E+01 \\
L-SHADE & 8.11E+02 ($<$) & 8.05E+02 & 3.07E+00 & 9.01E+02 ($>$) & 9.00E+02 & 1.02E+00 & 1.85E+03 ($<$) & 1.80E+03 & 3.34E+01 \\
AL-SHADE & 8.09E+02 ($<$) & 8.04E+02 & 2.10E+00 & 9.00E+02 ($\approx$) & 9.00E+02 & 2.03E-01 & 1.84E+03 ($<$) & 1.80E+03 & 3.09E+01 \\
LSHADE-cnEpSin & 8.28E+02 ($\approx$) & 8.20E+02 & 3.47E+00 & 9.93E+02 ($>$) & 9.19E+02 & 5.47E+01 & 1.80E+03 ($<$) & 1.80E+03 & 4.88E-01 \\
LSHADE-SPACMA & 8.10E+02 ($<$) & 8.04E+02 & 4.35E+00 & 9.01E+02 ($>$) & 9.00E+02 & 1.27E+00 & 1.85E+03 ($<$) & 1.82E+03 & 2.68E+01 \\ \hline
No. & C7 & & & C8 & & & C9 & & \\ \hline
BWE & 2.04E+03 & 2.03E+03 & 6.60E+00 & 2.22E+03 & 2.22E+03 & 3.34E-01 & 2.48E+03 & 2.48E+03 & 1.04E-04 \\
GA & 2.04E+03 ($>$) & 2.03E+03 & 7.86E+00 & 2.23E+03 ($>$) & 2.22E+03 & 1.56E+01 & 2.56E+03 ($>$) & 2.51E+03 & 4.39E+01 \\
LEA & 2.06E+03 ($>$) & 2.03E+03 & 2.00E+01 & 2.22E+03 ($>$) & 2.22E+03 & 5.02E+00 & 2.48E+03 ($>$) & 2.48E+03 & 1.36E-03 \\
HEOA & 2.18E+03 ($>$) & 2.06E+03 & 8.16E+01 & 2.25E+03 ($>$) & 2.23E+03 & 4.64E+01 & 2.51E+03 ($>$) & 2.48E+03 & 1.95E+01 \\
AOA & 2.17E+03 ($>$) & 2.11E+03 & 5.08E+01 & 2.37E+03 ($>$) & 2.23E+03 & 1.38E+02 & 2.64E+03 ($>$) & 2.56E+03 & 4.96E+01 \\
SCA & 2.10E+03 ($>$) & 2.07E+03 & 1.32E+01 & 2.24E+03 ($>$) & 2.24E+03 & 3.68E+00 & 2.53E+03 ($>$) & 2.51E+03 & 1.22E+01 \\
COA & 2.04E+03 ($\approx$) & 2.02E+03 & 1.50E+01 & 2.25E+03 ($>$) & 2.22E+03 & 4.68E+01 & 2.48E+03 ($>$) & 2.48E+03 & 2.73E-05 \\
GWO & 2.06E+03 ($>$) & 2.02E+03 & 3.12E+01 & 2.24E+03 ($>$) & 2.22E+03 & 4.16E+01 & 2.50E+03 ($>$) & 2.48E+03 & 1.49E+01 \\
JADE & 2.14E+03 ($>$) & 2.02E+03 & 3.89E+01 & 2.26E+03 ($>$) & 2.22E+03 & 2.05E+01 & 2.72E+03 ($>$) & 2.48E+03 & 9.17E+01 \\
CMA-ES & 2.50E+03 ($>$) & 2.50E+03 & 1.62E+00 & 2.31E+03 ($>$) & 2.22E+03 & 9.69E+01 & 2.47E+03 ($<$) & 2.47E+03 & 1.35E-12 \\
L-SHADE & 2.02E+03 ($<$) & 2.01E+03 & 4.96E+00 & 2.22E+03 ($<$) & 2.21E+03 & 1.45E+00 & 2.48E+03 ($<$) & 2.48E+03 & 1.19E-13 \\
AL-SHADE & 2.02E+03 ($<$) & 2.00E+03 & 8.10E+00 & 2.22E+03 ($<$) & 2.22E+03 & 5.36E-01 & 2.48E+03 ($<$) & 2.48E+03 & 0.00E+00 \\
LSHADE-cnEpSin & 2.02E+03 ($<$) & 2.00E+03 & 5.94E+00 & 2.22E+03 ($<$) & 2.22E+03 & 1.09E+00 & 2.47E+03 ($<$) & 2.47E+03 & 1.15E-01 \\
LSHADE-SPACMA & 2.02E+03 ($<$) & 2.00E+03 & 5.64E+00 & 2.22E+03 ($<$) & 2.20E+03 & 3.63E+00 & 2.48E+03 ($<$) & 2.48E+03 & 2.67E-13 \\ \hline
No. & C10 & & & C11 & & & C12 & & \\ \hline
BWE & 2.50E+03 & 2.50E+03 & 2.42E+01 & 2.92E+03 & 2.90E+03 & 4.07E+01 & 2.95E+03 & 2.94E+03 & 5.61E+00 \\
GA & 2.55E+03 ($>$) & 2.50E+03 & 5.46E+01 & 2.89E+03 ($>$) & 2.71E+03 & 1.48E+02 & 2.88E+03 ($<$) & 2.85E+03 & 1.55E+01 \\
LEA & 2.50E+03 ($<$) & 2.50E+03 & 6.51E-02 & 2.93E+03 ($>$) & 2.60E+03 & 7.90E+01 & 2.96E+03 ($>$) & 2.94E+03 & 1.46E+01 \\
HEOA & 3.70E+03 ($>$) & 2.50E+03 & 5.64E+02 & 3.17E+03 ($>$) & 2.70E+03 & 2.63E+02 & 3.13E+03 ($>$) & 3.00E+03 & 1.29E+02 \\
AOA & 4.23E+03 ($>$) & 2.55E+03 & 6.06E+02 & 6.18E+03 ($>$) & 4.34E+03 & 8.77E+02 & 3.50E+03 ($>$) & 3.19E+03 & 1.44E+02 \\
SCA & 2.52E+03 ($>$) & 2.50E+03 & 4.18E+01 & 4.10E+03 ($>$) & 3.51E+03 & 3.06E+02 & 3.00E+03 ($>$) & 2.96E+03 & 1.42E+01 \\
COA & 2.64E+03 ($>$) & 2.45E+03 & 1.91E+02 & 2.89E+03 ($>$) & 2.60E+03 & 1.22E+02 & 2.98E+03 ($>$) & 2.94E+03 & 3.46E+01 \\
GWO & 3.10E+03 ($>$) & 2.50E+03 & 5.55E+02 & 3.31E+03 ($>$) & 2.90E+03 & 3.40E+02 & 2.96E+03 ($>$) & 2.94E+03 & 1.74E+01 \\
JADE & 2.61E+03 ($>$) & 2.50E+03 & 6.03E+01 & 5.41E+03 ($>$) & 2.90E+03 & 1.65E+03 & 3.24E+03 ($>$) & 2.95E+03 & 1.50E+02 \\
CMA-ES & 5.85E+03 ($>$) & 5.19E+03 & 2.31E+02 & 2.97E+03 ($\approx$) & 2.90E+03 & 4.79E+01 & 4.94E+03 ($>$) & 2.90E+03 & 9.11E+02 \\
L-SHADE & 2.49E+03 ($<$) & 2.40E+03 & 3.89E+01 & 2.91E+03 ($<$) & 2.60E+03 & 7.12E+01 & 2.94E+03 ($<$) & 2.93E+03 & 5.60E+00 \\
AL-SHADE & 2.51E+03 ($>$) & 2.40E+03 & 5.15E+01 & 2.91E+03 ($<$) & 2.90E+03 & 3.05E+01 & 2.94E+03 ($<$) & 2.93E+03 & 9.65E+00 \\
LSHADE-cnEpSin & 2.50E+03 ($<$) & 2.50E+03 & 4.92E-02 & 2.96E+03 ($\approx$) & 2.90E+03 & 4.90E+01 & 2.89E+03 ($<$) & 2.89E+03 & 2.85E-01 \\
LSHADE-SPACMA & 2.50E+03 ($<$) & 2.40E+03 & 5.01E+01 & 2.92E+03 ($>$) & 2.90E+03 & 4.30E+01 & 2.95E+03 ($\approx$) & 2.93E+03 & 1.85E+01 \\ \bottomrule
\end{tabular}
\end{table*}

\begin{table}[!htbp]
\centering
\caption{BWE versus competitive algorithms: Wilcoxon rank-sum test, Friedman ranking, and mean ranking ($10D$ and $20D$).}
\label{tab:cec2022_Statistical}
\sffamily
\scriptsize
\setlength{\tabcolsep}{3pt}
\begin{tabular}{l *{6}{l}}
\toprule
Algorithm          & \multicolumn{2}{c}{Wilcoxon (BWE vs.)} & \multicolumn{2}{c}{FM-Rank} & \multicolumn{2}{c}{M-Rank} \\
\cmidrule(lr){2-3} \cmidrule(lr){4-5} \cmidrule(lr){6-7}
                   & 10D     & 20D     & 10D      & 20D      & 10D & 20D \\
\midrule
BWE                & /       & /       & 4.1667   & 4.7500   & 4   & 5   \\
GA                 & 12/0/0  & 11/0/1  & 11.3333  & 6.5833   & 12  & 6   \\
LEA                & 11/0/1  & 10/1/1  & 5.6667   & 6.6667   & 6   & 7   \\
HEOA               & 12/0/0  & 12/0/0  & 11.4167  & 11.5000  & 13  & 12  \\
AOA                & 12/0/0  & 12/0/0  & 11.0833  & 12.0000  & 11  & 13  \\
SCA                & 12/0/0  & 12/0/0  & 9.4167   & 10.3333  & 9   & 11  \\
COA                & 10/2/0  & 8/1/3   & 7.5000   & 7.4167   & 7   & 8   \\
GWO                & 11/0/1  & 12/0/0  & 8.9167   & 9.1667   & 8   & 10  \\
JADE               & 12/0/0  & 12/0/0  & 12.0000  & 12.7500  & 14  & 14  \\
CMA-ES             & 6/2/4   & 7/1/4   & 9.4167   & 8.9167   & 10  & 9   \\
L-SHADE            & 3/1/8   & 4/0/8   & 3.9167   & 3.9167   & 3   & 3   \\
AL-SHADE           & 3/0/9   & 2/2/8   & 3.5000   & 3.5833   & 2   & 2   \\
LSHADE-cnEpSin     & 1/1/10  & 1/2/9   & 2.4167   & 2.7500   & 1   & 1   \\
LSHADE-SPACMA      & 5/0/7   & 5/1/6   & 4.2500   & 4.6667   & 5   & 4   \\
\bottomrule
\end{tabular}
\end{table}

\subsection{Computational cost analysis}
Computational cost is a key metric for evaluating metaheuristic algorithms.
We computed the computational cost of BWE and its competitors using the methodology defined in the literature \citep{liang2013problem}.

\begin{enumerate}[label=\arabic*., align=left, leftmargin=*]
    \renewcommand{\labelenumi}{\Roman{enumi}.}
    \item Firstly, we define $x=0.55$, bring $x$ into \cref{eq:cost1}, and run it independently 1,000,000 times. Also record the run time as $T_0$.
    \begin{equation}
    \label{eq:cost1}
        \begin{split}
            x &= x+x; \quad x=x/2; \quad x=x \times x; \quad x=\sqrt{x}; \\
            x &= \ln x; \quad x=e^x; \quad x=x/(x+2)
        \end{split}
    \end{equation}
    \item Next, F18 was evaluated on the CEC2017 for 200,000 times, and the run time was recorded as $T_1$.
    \item Then, the algorithm with tests was used to evaluate F18 on CEC2017 200,000 times (50 dimensions), and a running time of $T_2$ was recorded.
    \item Then again, step (III) was then repeated five times, and the average of the five $T_2$ times, $T_{\mathrm{mean}}$, was recorded.
    \item Finally, \cref{eq:cost2} is calculated to obtain $\hat{T}$ as the computational cost of the algorithm to be tested.
    \begin{equation}
    \label{eq:cost2}
        \hat{T}=((T_{\mathrm{mean}}-T_1 ))⁄T_0 
    \end{equation}
\end{enumerate}

As shown in the \cref{table:Cost}, BWE has a relatively standard computational cost.
However, BWE demonstrates an absolute advantage over conventional optimization algorithms in terms of optimization results.

\begin{table}[!htbp]
\caption{Computational costs for the BWE and its competitors.}
\label{table:Cost}
\sffamily
\footnotesize
\begin{tabular}{lllll}
\toprule
Algorithm      & $T_{mean}$  & $T_0$   & $T_1$     & $\hat{T}$     \\ \hline
BWE            & 1.8479 & 0.0534 & 1.0280 & 15.3590  \\
GA             & 1.3871 & 0.0598 & 1.0487 & 5.6530   \\
LEA            & 2.4169 & 0.0531 & 0.9791 & 27.0891  \\
HEOA           & 1.2099 & 0.0556 & 1.0311 & 3.2166   \\
AOA            & 1.6163 & 0.0556 & 1.0219 & 10.6885  \\
SCA            & 1.7818 & 0.0542 & 0.9993 & 14.4323  \\
COA            & 1.9637 & 0.0573 & 0.9529 & 17.6468  \\
GWO            & 2.0094 & 0.0635 & 1.0080 & 15.7723  \\
CMA-ES         & 5.4807 & 0.0587 & 0.9899 & 76.4586  \\
JADE           & 1.4465 & 0.0564 & 1.0011 & 7.8951   \\
L-SHADE        & 1.6389 & 0.0601 & 1.0166 & 10.3583  \\
AL-SHADE       & 1.5671 & 0.0601 & 0.9636 & 10.0452  \\
LSHADE-cnEpSin & 1.8978 & 0.0557 & 1.0635 & 14.9784  \\
LSHADE-SPACMA  & 6.4191 & 0.0539 & 0.9922 & 100.6288 \\
\bottomrule
\end{tabular}
\end{table}

\subsection{Sensitivity analysis}
A comprehensive sensitivity analysis of BWE's key hyperparameters is conducted on the CEC2017.

\subsubsection{Sensitivity analysis of $\alpha$ and $\beta$}
The step parameter $\tau$ is governed by the stochastic factor $\alpha$ and base factor $\beta$, constrained by $\alpha + \beta = 1.00$ to ensure terminal convergence ($\tau \to 1$ as $t \to T$). Thus, we analyze $\alpha \in [0.00, 0.40]$ with a 0.05 interval.

\begin{figure}
    \centering
    \includegraphics[width=0.9\columnwidth]{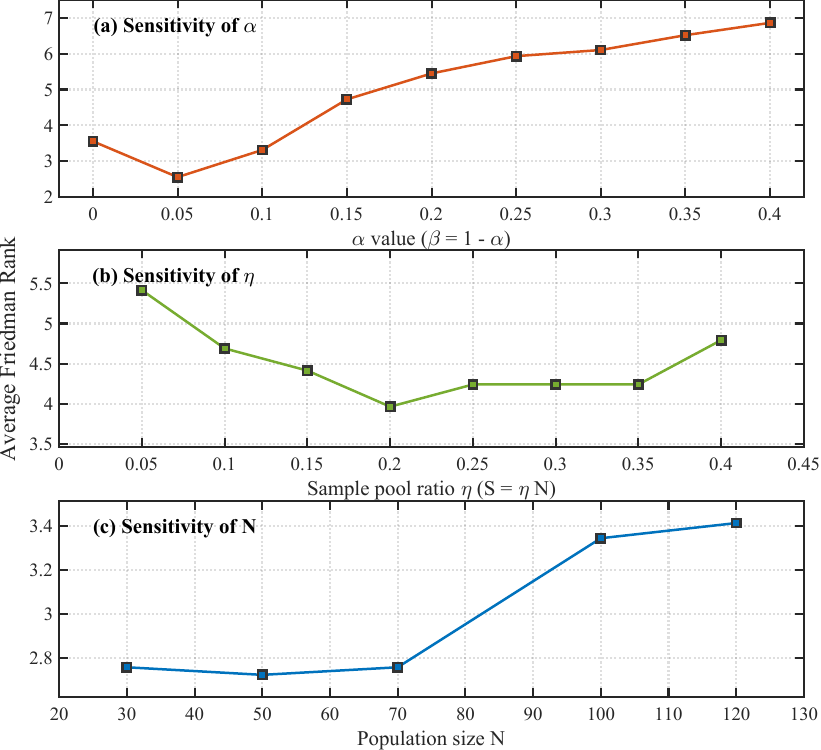}
    \caption{BWE parameter sensitivity analysis.}
    \label{fig:Sensitivity}
\end{figure}

In \cref{fig:Sensitivity}(a), $\alpha$ reveals a "V-shaped" trend in the Friedman mean ranking. The purely deterministic version ($\alpha=0$, Rank 3.55) underperforms, validating the role of stochastic perturbation in escaping local optima. Peak performance for $10D$ occurs at $\alpha=0.05$ (Rank 2.55), indicating that subtle randomness enhances precision in low-dimensional spaces. However, a smaller $\alpha$ may cause insufficient exploration as problem complexity increases. Given the multi-dimensional benchmarks ($10D$ to $100D$), $\alpha=0.2$ is selected as a robust configuration, providing a higher stochastic buffer for rugged, high-dimensional search landscapes.

\subsubsection{Sensitivity analysis of $\eta$}
The sample pool ratio $\eta$ determines the random walk sampling density, directly affecting the selection of intermediate control points and the algorithm's time complexity. We evaluate $\eta \in [0.05, 0.40]$ (step 0.05) to balance search precision and efficiency (\cref{fig:Sensitivity}(b)).

The ranking exhibits a clear "V-shaped" curve peaking at $\eta=0.20$ (Rank 3.9655), suggesting that utilizing 20\% of the population as guidance nodes sufficiently represents topological features. A low $\eta$ ($\leq 0.15$) causes sampling sparsity and a loss of geometric information, while an oversized $\eta$ ($\geq 0.25$) leads to performance deterioration (e.g., Rank 4.7931 at $\eta=0.40$) due to the information dilution effect from low-quality solutions. Thus, $\eta=0.20$ is empirically justified, balancing high optimization precision with low computational overhead for distance metrics.

\subsubsection{Sensitivity analysis of $N$}
Population size $N$ dictates the balance between exploration breadth per generation and total evolutionary iterations under a fixed budget. We test $N \in \{30, 50, 70, 100, 120\}$ with a constant $MaxFEs = 100,000$ (\cref{fig:Sensitivity}(c)).

The Friedman rank follows a "V-shaped" trend, peaking at $N=50$ (Rank 2.7241). Performance remains remarkably stable for $N \in [30, 70]$, but degrades noticeably for $N \geq 100$ (e.g., Rank 3.4138 at $N=120$). With a fixed $MaxFEs$, an excessively large $N$ reduces the number of generations, preventing the Bézier-guided paths from undergoing sufficient refinement for high-precision convergence. Consequently, $N=50$ is selected as the standard configuration, as it provides a robust balance between global exploration and local exploitation efficiency.

\section{Constrained engineering problems}
\label{sec:engineering}
In this section, we compare the performance of all algorithms on five real-world engineering problems to assess the BWE's capability to solve practical optimization challenges.
For all experiments, the population size is set to 50, the maximum number of function evaluations is 100,000, and each algorithm is run independently 25 times.

\subsection{The three-bar truss design problem}
\begin{figure}
    \centering
    \includegraphics[width=0.7\columnwidth]{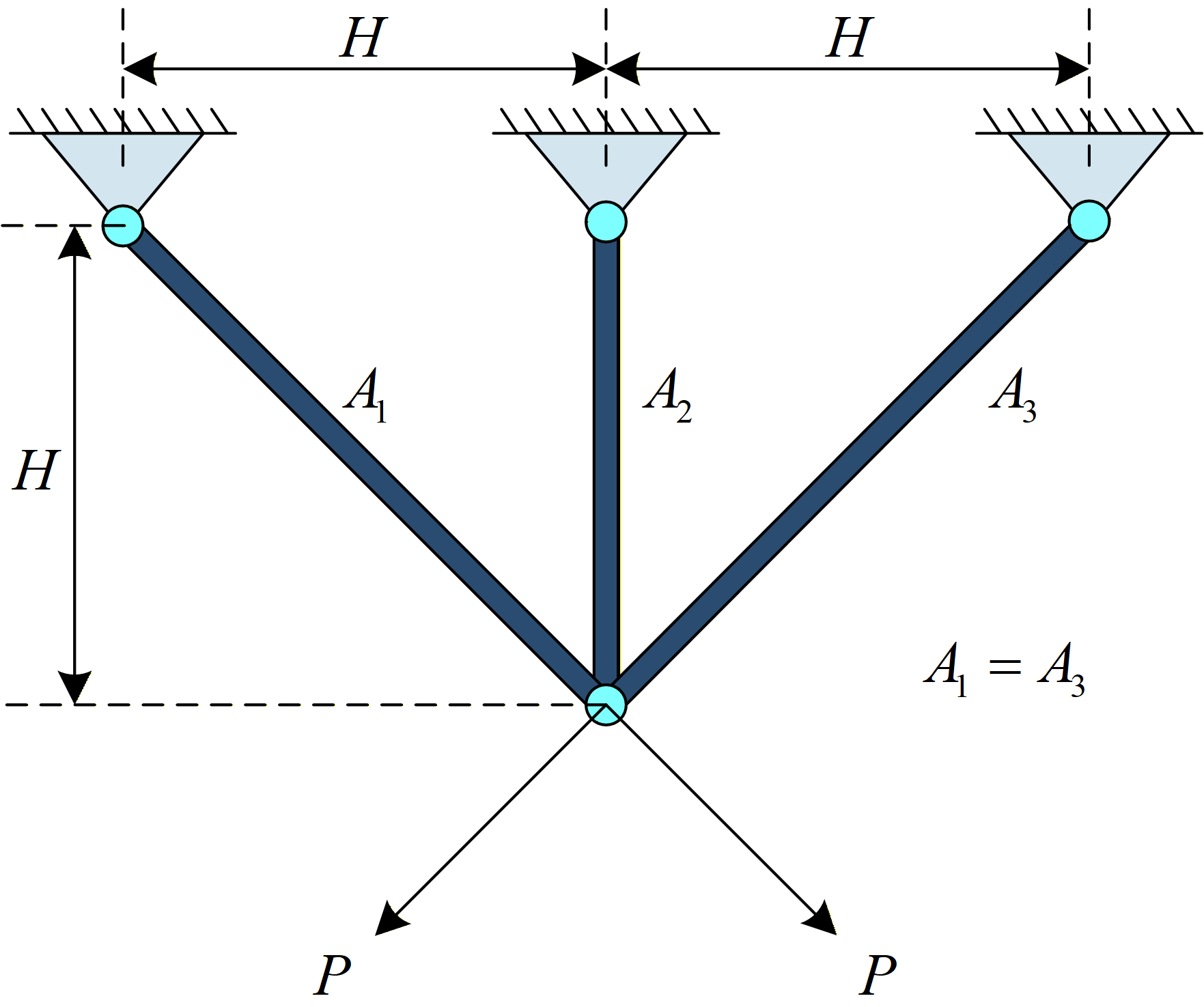}
    \caption{Diagram of three-bar truss problem.}
    \label{fig:threebartruss}
\end{figure}

The three-bar truss design problem is a classical benchmark in structural optimization \citep{ouyang2025multi}. 
The objective is to minimize the total weight (or volume, assuming unit density) of the truss by optimizing the cross-sectional areas of its members. 
The design variables are the cross-sectional areas $A_1$ (for the two symmetric inclined bars) and $A_2$ (for the vertical bar). 
The mathematical formulation is given in \cref{eq:three_bar_truss}, subject to stress constraints on the members 
(refer to \cref{fig:threebartruss} for the schematic of the structure).

Let $\mathbf{x} = [x_1, x_2]^\top = [A_1, A_2]^\top$. 

Minimize:
\begin{equation}
f(\mathbf{x}) = (2\sqrt{2}\, x_1 + x_2) \times l
\label{eq:three_bar_truss}
\end{equation}

Subject to:
\[\begin{gathered}
g_1(\mathbf{x})=\frac{\sqrt{2x_1}+x_2}{\sqrt{2x_1^2}+2x_1x_2}P-\sigma\leq0, \end{gathered}\]
\[\begin{gathered}
g_2(\mathbf{x})=\frac{x_2}{\sqrt{2x_1^2}+2x_1x_2}P-\sigma\leq0,
\end{gathered}\]
\[\begin{gathered}
g_3(\mathbf{x})=\frac{1}{\sqrt{2x_2}+x_1}P-\sigma\leq0.
\end{gathered}\]

The design variables are bounded by
\[ 0 \leq x_1, x_2 \leq 1, \]
where $l = 100\,\mathrm{cm}$, $P = 2\,\mathrm{kN/cm^2}$, and $\sigma = 2\,\mathrm{kN/cm^2}$.

\begin{table}[!htbp]
\caption{The best optimization results of three-bar truss problem.}
\label{tab:threebar}
\centering
\begin{tabular*}{\linewidth}{@{\extracolsep{\fill}} llll @{\extracolsep{\fill}}}
\toprule
Algorithm & $x_1$ & $x_2$ & $f$ \\ \hline
BWE  & 0.788601 & 0.408458 & 263.8958488 \\
GA   & 0.793260 & 0.395615 & 263.9293587 \\
LEA  & 0.789214 & 0.406726 & 263.8960654 \\
HEOA & 0.780316 & 0.432692 & 263.9758930 \\
AOA  & 0.791815 & 0.399872 & 263.9462475 \\
SCA  & 0.788113 & 0.409861 & 263.8979911 \\
COA  & 0.788646 & 0.408330 & 263.8958505 \\
GWO  & 0.788735 & 0.408079 & 263.8958489 \\ 
\bottomrule
\end{tabular*}
\end{table}

\cref{tab:threebar} presents the best results achieved by all algorithms for the three-bar truss problem. As observed, the optimization results across all algorithms are relatively close for this problem. However, BWE still outperforms all other algorithms, securing the top rank. COA ranks second, followed by GWO in third place, while HEOA and AOA demonstrate comparatively poorer performance.

\subsection{The gear train design problem}
\begin{figure}
    \centering
    \includegraphics[width=0.8\columnwidth]{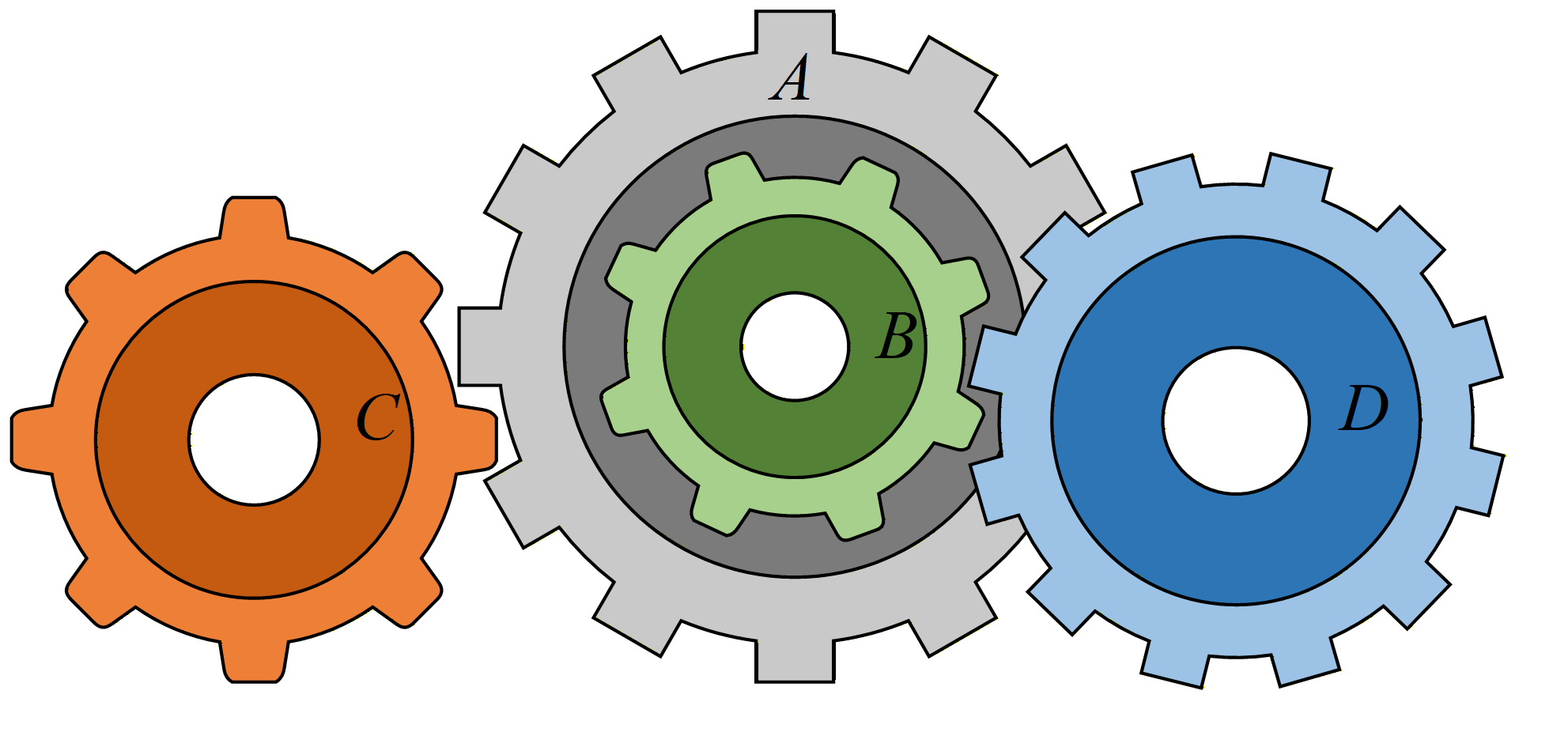}
    \caption{Diagram of gear train problem.}
    \label{fig:geartrain}
\end{figure}

The gear train design problem is a well-known discrete optimization benchmark in mechanical engineering \citep{gao2025escape}. 
The objective is to minimize the error in the transmission ratio relative to the target value of $1/6.931$ by selecting appropriate integer tooth numbers for the gears. A schematic of the typical four-gear train configuration is provided in \cref{fig:geartrain}. 
The gear meshing conditions are implicitly satisfied by integer tooth counts and problem formulation.

The design variables represent the number of teeth on each gear: 
$\mathbf{x} = [x_1, x_2, x_3, x_4]^\top = [n_1, n_2, n_3, n_4]^\top$, 
where $n_1$ is the input gear, $n_2$ the driving gear on the intermediate shaft, $n_3$ the driven gear on the intermediate shaft, and $n_4$ the output gear.

Minimize:
\begin{equation}
f(\mathbf{x}) = \left( \frac{1}{6.931} - \frac{x_2 x_3}{x_1 x_4} \right)^2
\label{eq:gear_train}
\end{equation}

In this work, the integer constraints are handled by a simple discretization scheme based on the $\text{round}(\cdot)$ operator. The variables are restricted to the integer range
\[ x_1, x_2, x_3, x_4 \in \{12, 13, \dots, 60\}. \]

The optimization results obtained by BWE and the competing algorithms for this problem are summarized in \cref{tab:gear}. 
The proposed BWE, along with LEA, SCA, COA, and GWO, achieved the best performance and jointly ranked first with identical optimal objective values. 
HEOA and GA followed closely, sharing second place. 
In comparison, AOA exhibited noticeably inferior performance, placing near the bottom of the ranking.

\begin{table}[!htbp]
\caption{The best optimization results of gear train problem.}
\label{tab:gear}
\centering
\begin{tabular*}{\linewidth}{@{\extracolsep{\fill}} llllll @{\extracolsep{\fill}}}
\toprule
Algorithm & $x_1$ & $x_2$ & $x_3$ & $x_4$ & $f$ \\ \hline
BWE  & 49 & 16 & 19 & 43 & 2.70086E-12 \\
GA   & 57 & 31 & 13 & 49 & 9.93988E-11 \\
LEA  & 49 & 16 & 19 & 43 & 2.70086E-12 \\
HEOA & 51 & 26 & 15 & 53 & 2.30782E-11 \\
AOA  & 48 & 17 & 22 & 54 & 1.16612E-10 \\
SCA  & 49 & 19 & 16 & 43 & 2.70086E-12 \\
COA  & 43 & 19 & 16 & 49 & 2.70086E-12 \\
GWO  & 49 & 16 & 19 & 43 & 2.70086E-12 \\ 
\bottomrule
\end{tabular*}
\end{table}

\subsection{The cantilever beam design problem}
\begin{figure}
    \centering
    \includegraphics[width=0.8\columnwidth]{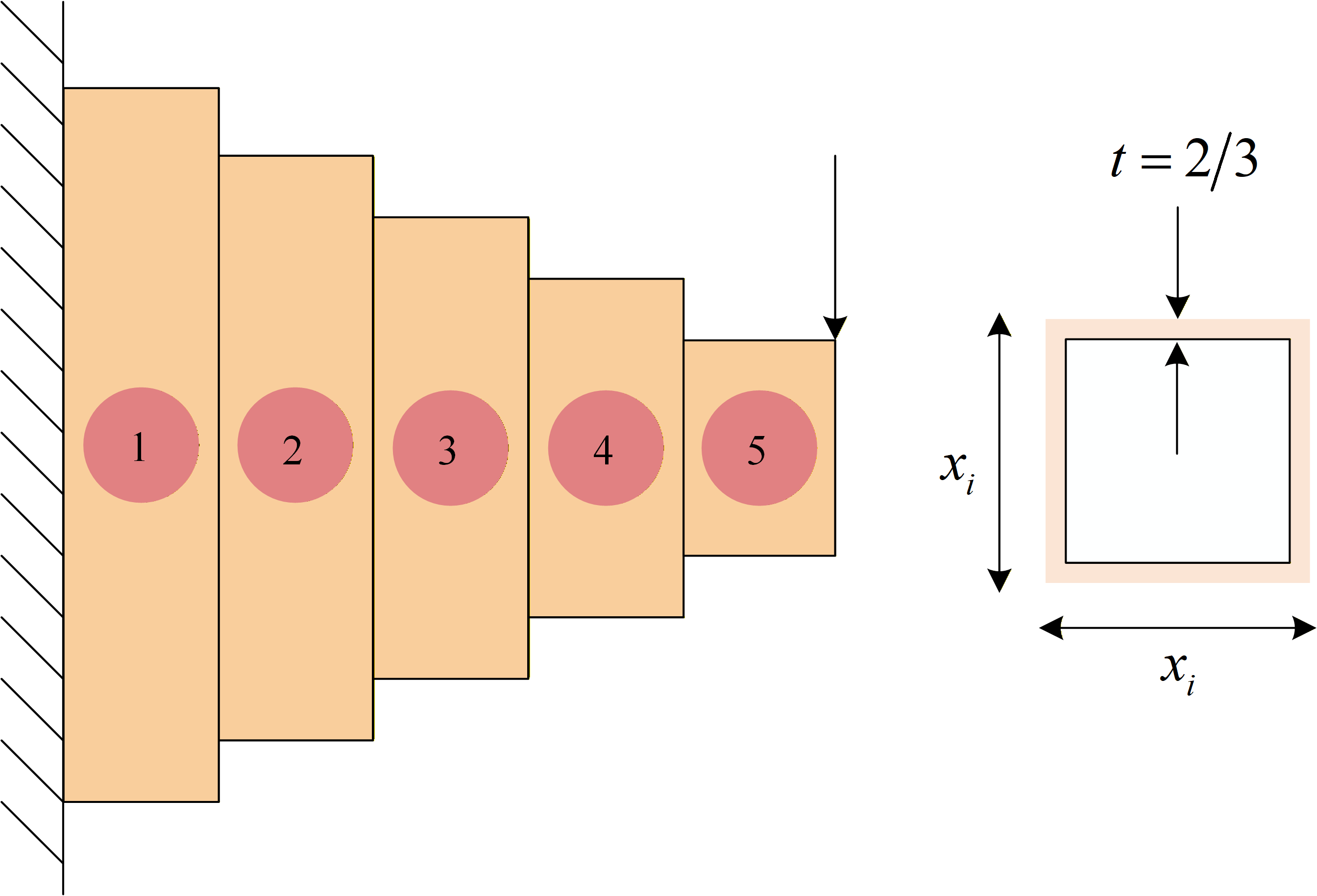}
    \caption{Diagram of cantilever beam problem.}
    \label{fig:cantileverbeam}
\end{figure}

The cantilever beam design problem is a classical constrained optimization benchmark in structural engineering \citep{ouyang2025beaver}. 
The objective is to minimize the total mass (or volume, assuming constant material density) of a five-element stepped cantilever beam subjected to a tip load, by optimizing the cross-sectional widths of each segment. 
The wall thickness of all elements is fixed at $2/3$ (dimensionless). A schematic illustration of the typical cantilever beam configuration is shown in \cref{fig:cantileverbeam}.

The design variables are the widths of the five beam segments: 
$\mathbf{x} = [x_1, x_2, x_3, x_4, x_5]^\top$, 
where $x_1$ corresponds to the segment closest to the fixed support and $x_5$ to the tip segment. The mathematical model for this problem is formulated in \cref{eq:cantilever_beam}.

Minimize:
\begin{equation}
f(\mathbf{x}) = 0.0624 \left( x_1 + x_2 + x_3 + x_4 + x_5 \right)
\label{eq:cantilever_beam}
\end{equation}

Subject to the deflection constraint:
\[g(\mathbf{x})=\frac{61}{x_1^3}+\frac{37}{x_2^3}+\frac{19}{x_3^3}+\frac{7}{x_4^3}+\frac{1}{x_5^3}-1\leq0.\]

The design variables are bounded by
\[ 0.01 \leq x_1, x_2, x_3, x_4, x_5 \leq 100. \]

The optimization results obtained by BWE and the competing algorithms for this problem are summarized in \cref{tab:cantilever}. 
The proposed BWE achieved the best performance, attaining the lowest objective value and ranking first. 
COA followed closely in second place, with only a marginal difference in the objective function value. 
LEA and GWO also produced competitive results, though slightly behind the top performers.

\begin{table*}[!htbp]
\caption{The best optimization results of cantilever beam problem.}
\label{tab:cantilever}
\centering
\begin{tabular*}{\linewidth}{@{\extracolsep{\fill}} lllllllll @{\extracolsep{\fill}}}
\toprule
 & BWE & GA & LEA & HEOA & AOA & SCA & COA & GWO \\ \hline
$x_1$ & 6.017240 & 5.798262 & 6.004703 & 5.101970 & 6.921504 & 5.711693 & 6.012693 & 6.009764 \\
$x_2$ & 5.308314 & 5.353929 & 5.308371 & 5.048000 & 5.838029 & 5.333728 & 5.312056 & 5.310970 \\
$x_3$ & 4.496258 & 4.912094 & 4.495530 & 5.037346 & 4.229940 & 4.773796 & 4.494843 & 4.494879 \\
$x_4$ & 3.501381 & 3.700574 & 3.518641 & 5.130901 & 4.257413 & 3.551357 & 3.501450 & 3.504029 \\
$x_5$ & 2.150475 & 2.100672 & 2.146678 & 2.757903 & 1.808930 & 2.207816 & 2.152629 & 2.154075 \\
$f$   & 1.3399568 & 1.3644091 & 1.3399728 & 1.4399499 & 1.4386829 & 1.3464915 & 1.3399571 & 1.3399600 \\
\bottomrule
\end{tabular*}
\end{table*}

\subsection{The corrugated bulkhead design problem}
\begin{figure}
    \centering
    \includegraphics[width=0.7\columnwidth]{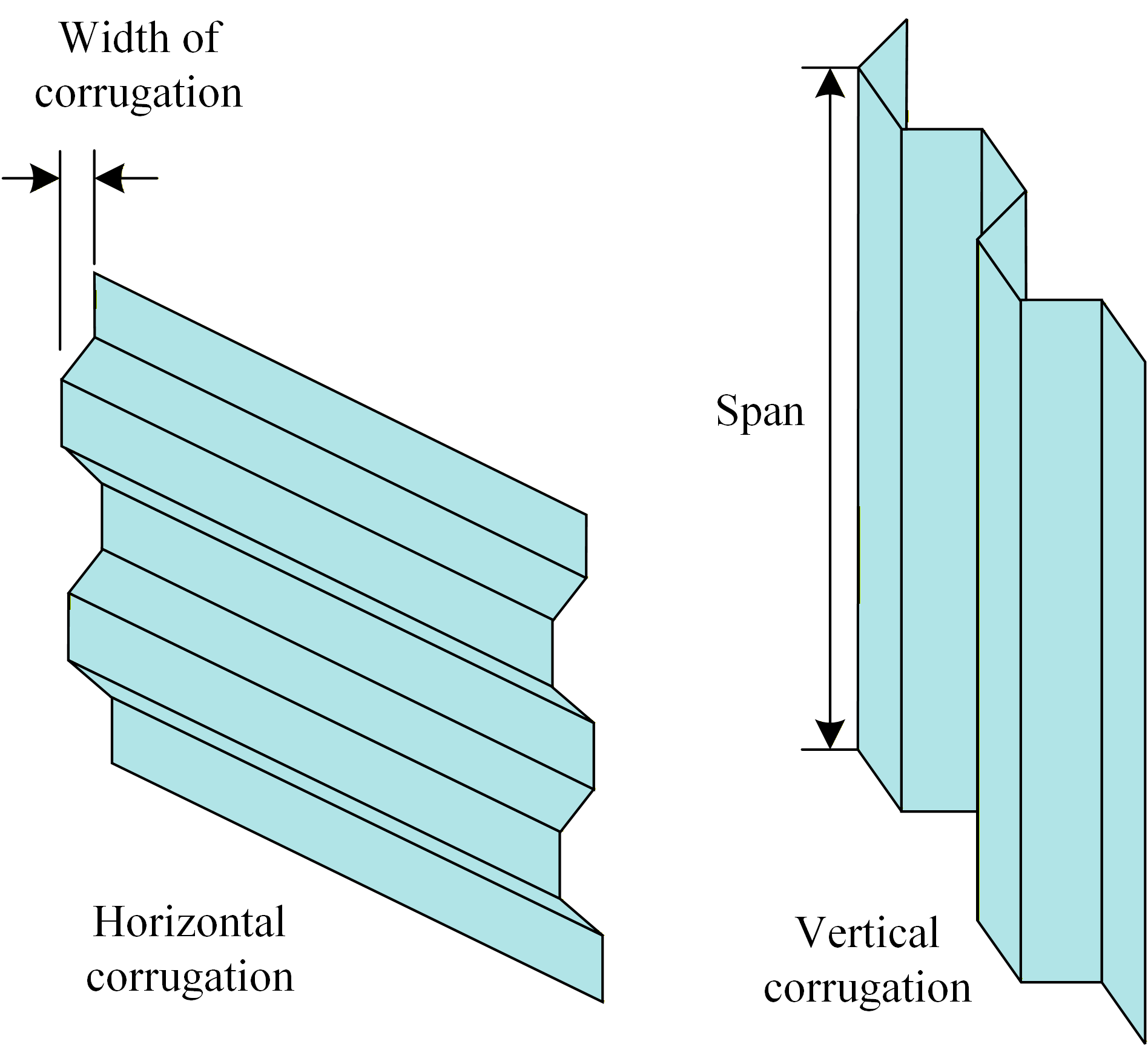}
    \caption{Diagram of corrugated bulkhead problem.}
    \label{fig:corrugatedbulkhead}
\end{figure}

The corrugated bulkhead design problem is a realistic constrained optimization benchmark drawn from ship structural engineering, particularly for chemical tankers \citep{wang2025multi}. 
The primary objective is to minimize the structural mass of the corrugated bulkhead while satisfying a set of strength, buckling, and geometric feasibility constraints.
A schematic of the typical corrugated bulkhead geometry and loading configuration is illustrated in \cref{fig:corrugatedbulkhead}.

The design variables represent key geometric parameters of the corrugation profile:
$\mathbf{x} = [x_1, x_2, x_3, x_4]^\top$, 
where $x_1$ is the corrugation unit width, $x_2$ is the corrugation height, $x_3$ is the effective (projected) length of the bulkhead panel, $x_4$ is the plate thickness.

Minimize:
\begin{equation}
f(\mathbf{x}) = \frac{5.885\, x_4 (x_1 + x_3)}{x_1 + \sqrt{|x_3^2 - x_2^2|}}
\label{eq:corrugated_bulkhead}
\end{equation}

Subject to:
\begin{align*}
g_1(\mathbf{x}) ={}& -x_4 x_2 \left(0.4 x_1 + \frac{x_3}{6}\right) \notag \\
&+ 8.94 \left( x_1 + \sqrt{|x_3^2 - x_2^2|} \right) \le 0, \\
g_2(\mathbf{x}) ={}& -x_4 x_2^2 \left(0.2 x_1 + \frac{x_3}{12}\right) \notag \\
&+ 2.2 \left[ 8.94 \left( x_1 + \sqrt{|x_3^2 - x_2^2|} \right) \right]^{4/3} \le 0, \\
g_3(\mathbf{x}) ={}& -x_4 + 0.0156x_1 + 0.15 \le 0, \\
g_4(\mathbf{x}) ={}& -x_4 + 0.0156x_3 + 0.15 \le 0, \\
g_5(\mathbf{x}) ={}& -x_4 + 1.05 \le 0, \\
g_6(\mathbf{x}) ={}& -x_3 + x_2 \le 0.
\end{align*}

The design variables are bounded by
\[ 0 \leq x_1, x_2, x_3 \leq 100, \qquad 0 \leq x_4 \leq 5. \]

\begin{table*}[!htbp]
\caption{The best optimization results of corrugated bulkhead problem.}
\label{tab:corrugated}
\centering
\begin{tabular*}{\linewidth}{@{\extracolsep{\fill}} lllllllll @{\extracolsep{\fill}}}
\toprule
 & BWE & GA & LEA & HEOA & AOA & SCA & COA & GWO \\ \hline
$x_1$ & 57.692052   & 27.279600    & 57.652027   & 30.259014    & 47.282658   & 51.078194   & 57.692128   & 57.691533   \\
$x_2$ & 34.147626   & 34.923619    & 34.149888   & 35.988426    & 34.719467   & 34.064346   & 34.147616   & 34.147206   \\
$x_3$ & 57.692226   & 53.313824    & 57.683409   & 59.047327    & 57.314328   & 57.061057   & 57.692087   & 57.688726   \\
$x_4$ & 1.050000    & 1.050779     & 1.050082    & 1.072336     & 1.051371    & 1.051287    & 1.050000    & 1.050010    \\
$f$   & 6.84296410  & 7.37653647   & 6.84405725  & 7.31249030   & 6.96755452  & 6.90757547  & 6.84296414  & 6.84308539  \\
\bottomrule
\end{tabular*}
\end{table*}

\cref{tab:corrugated} reports the best solutions obtained by different algorithms for the corrugated bulkhead design problem.
The proposed BWE achieves the minimum objective value, slightly outperforming COA, GWO, and LEA, whose results are very close to the best solution.
This indicates that several algorithms are able to reach high-quality designs for this problem.
However, GA and HEOA perform relatively poorly, producing noticeably larger objective function values than the other methods.

\subsection{The rolling element bearing design problem}
The rolling element bearing design problem is a complex constrained optimization benchmark in mechanical engineering \citep{gao2025freedom}. 
The primary objective is to maximize the fatigue life of the bearing, which is directly proportional to its dynamic load-carrying capacity.

\begin{figure}
    \centering
    \includegraphics[width=0.8\columnwidth]{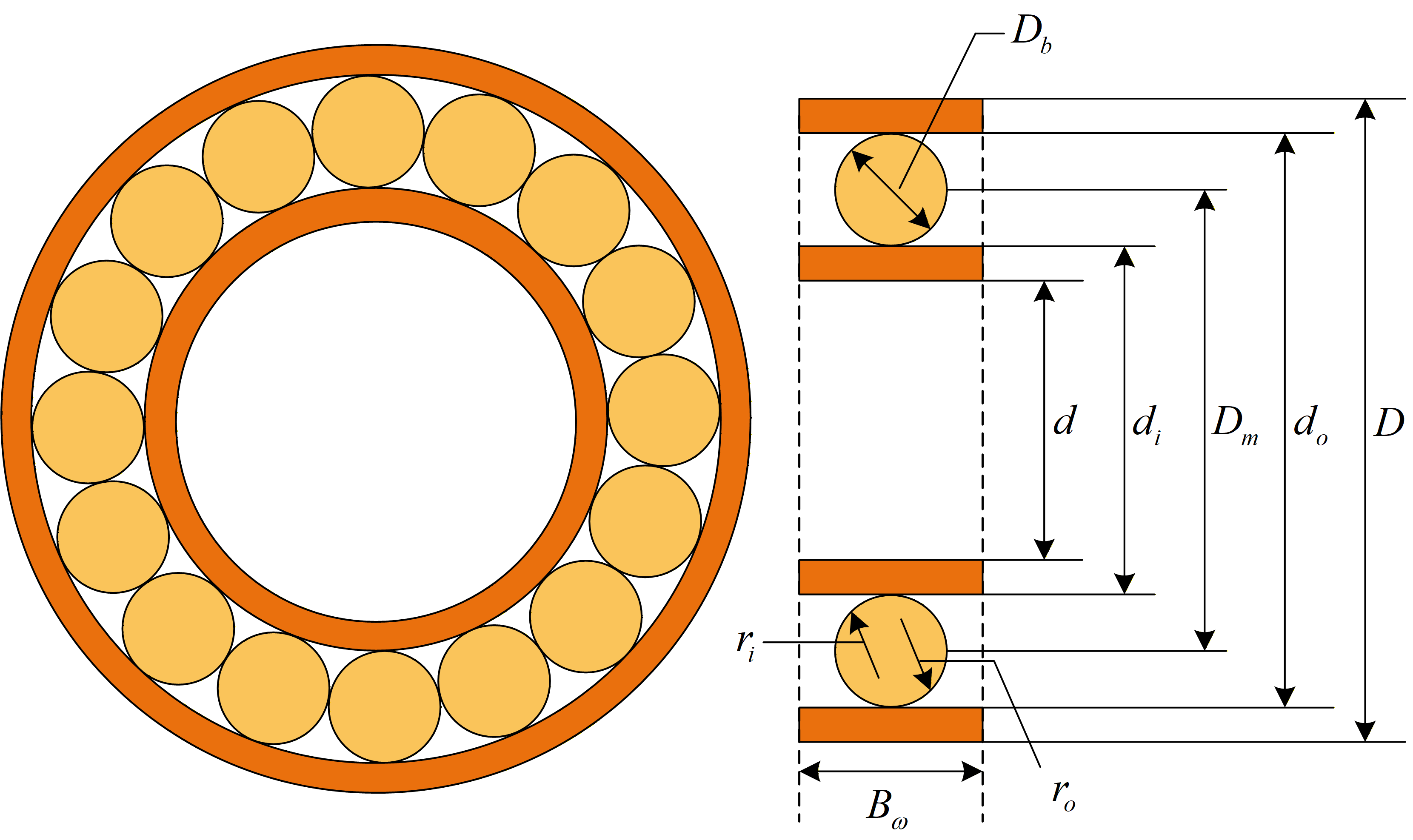}
    \caption{Diagram of rolling element bearing problem.}
    \label{fig:rolling}
\end{figure}

The problem involves ten continuous design variables and nine inequality constraints. 
A schematic illustration of the typical rolling element bearing configuration is provided in \cref{fig:rolling}, and the complete mathematical formulation is given in \cref{eq:rolling}.

The design variables are defined as follows:
\begin{equation*}
\begin{split}
\mathbf{x}
&= \begin{bmatrix} x_1, x_2, x_3, x_4, x_5, x_6, x_7, x_8, x_9, x_{10} \end{bmatrix} \\
&= \begin{bmatrix} D_m, D_b, Z, f_i, f_o, K_{Dmin}, K_{Dmax}, \varepsilon, e, \zeta \end{bmatrix}.
\end{split}
\end{equation*}
where $D_m$ denotes the pitch diameter of the bearing (mm), $D_b$ denotes the diameter of the rolling element (mm), $Z$ is the number of rolling elements, $f_i$ and $f_o$ are the inner and outer raceway curvature factors, respectively, $K_{Dmin}$ and $K_{Dmax}$ are the minimum and maximum roller diameter factors, respectively, $\varepsilon$ is the inner ring shoulder diameter factor, $e$ is the outer ring shoulder diameter factor, and $\zeta$ is the lubricant film thickness factor.

Maximize:
\begin{equation}
f(\mathbf{x}) = 
\begin{cases} 
f_c \cdot Z^{2/3} \cdot D_b^{1.8},  D_b \leq 25.4\, \\
3.647 \cdot f_c \cdot Z^{2/3} \cdot D_b^{1.4}, D_b > 25.4\
\end{cases}
\label{eq:rolling}
\end{equation}

Subject to:
\[g_{1}(\mathbf{x})=-{\frac{\varphi_{0}}{2\arcsin(D_{b}/D_{m})}}+Z-1\leq0, \]
\[g_{2}(\mathbf{x})=-2D_{b}+K_{Dmin}(D-d)\leq0, \]
\[g_{3}(\mathbf{x})=-K_{Dmax}(D-d)+2D_{b}\leq0, \]
\[g_{4}(\mathbf{x})=-D_{m}+(0.5-e)(D+d)\leq0, \]
\[g_{5}(\mathbf{x})=D_{m}-(0.5+e)(D+d)\leq0, \]
\[g_{6}(\mathbf{x})=-D_{m}+0.5(D+d)\leq0, \]
\[g_{7}(\mathbf{x})=-0.5(D-D_{m}-D_{b})+\varepsilon D_{b}\leq0, \]
\[g_{8}(\mathbf{x})=\zeta B_{\omega}-D_{b}\leq0, \]
\[g_{9}(\mathbf{x})=0.515-f_{i}\leq0, \]
\[g_{10}(\mathbf{x})=0.515-f_{o}\leq0.\]
where
\begin{equation*}
\begin{aligned}
f_c = &37.91 \Biggl\{
  1 + \Biggl[ 1.04 \left( \frac{1-\gamma}{1+\gamma} \right)^{1.72}\left( \frac{f_i(2f_o-1)}{f_o(2f_i-1)} \right)^{0.41}\\
&\Biggr]^{10/3}
  \Biggr\}^{-0.3} \times \left[ \frac{\gamma^{0.3} (1-\gamma)^{1.39}}{f_o (1+\gamma)^{1/3}} \right]
\times \left[ \frac{2f_i}{2f_i-1} \right]^{0.41}.
\end{aligned}
\end{equation*}

\begin{equation*}
\begin{aligned}
\varphi_{0} = &2\pi - 2\arccos \Biggl\{\\
&\frac{ \bigl[(D-d)/2 - 3(T/4)\bigr]^2
      + \bigl(D/2 - T/4 - D_b\bigr)^2 }
     { 2 \bigl[(D-d)/2 - 3(T/4)\bigr]
       \bigl(D/2 - T/4 - D_b\bigr) } \\
&- \frac{ (d/2 + T/4)^2 }
        { 2 \bigl[(D-d)/2 - 3(T/4)\bigr]
          \bigl(D/2 - T/4 - D_b\bigr) }
\Biggr\}.
\end{aligned}
\end{equation*}

\begin{align*}
T &= D - d - 2D_b, \quad
B_\omega = 30, \quad
D = 160, \\
d &= 90, \quad
r_i = r_o = 11.033.
\end{align*}

Variable ranges:
\begin{equation*}
\begin{aligned}
&0.5(D+d) \leq x_{1} \leq 0.6(D+d), \\
&0.15(D-d) \leq x_{2} \leq 0.45(D-d), \\
&4 \leq x_{3} \leq 50, \\
&0.515 \leq x_{4}, x_{5} \leq 0.6, \\
&0.4 \leq x_{6} \leq 0.5, \\
&0.6 \leq x_{7} \leq 0.7, \\
&0.3 \leq x_{8} \leq 0.4.
\end{aligned}
\end{equation*}

\begin{table*}[!htbp]
\centering
\caption{The best optimization results of rolling element bearing problem.}
\label{tab:rolling}

\begin{tabular*}{\linewidth}{@{\extracolsep{\fill}} lllllllll @{\extracolsep{\fill}}}
\toprule
    & BWE & GA & LEA & HEOA & AOA & SCA & COA & GWO \\ \hline
$x_1$  & 125.719055 & 127.500369 & 125.718921 & 126.914512 & 125 & 125 & 125.719018 & 125.714146 \\
$x_2$  & 21.425590 & 16.599469 & 21.425535 & 18.681463 & 21.222296 & 21.256095 & 21.425580 & 21.423891 \\
$x_3$  & 11.214661 & 12.425448 & 11.098732 & 11.718520 & 11.372401 & 11.179570 & 11.043250 & 11.362095 \\
$x_4$  & 0.515 & 0.589968 & 0.515 & 0.515 & 0.515 & 0.515 & 0.515 & 0.515000 \\
$x_5$  & 0.515 & 0.515131 & 0.518947 & 0.572258 & 0.6 & 0.540596 & 0.515041 & 0.525075 \\
$x_6$  & 0.400152 & 0.455753 & 0.498022 & 0.496351 & 0.4 & 0.401520 & 0.400970 & 0.430843 \\
$x_7$  & 0.678865 & 0.631829 & 0.652130 & 0.686000 & 0.7 & 0.7 & 0.651657 & 0.699138 \\
$x_8$  & 0.3 & 0.312122 & 0.3 & 0.348728 & 0.314598 & 0.3 & 0.300001 & 0.300118 \\
$x_9$  & 0.045108 & 0.029144 & 0.092487 & 0.097498 & 0.037182 & 0.069254 & 0.1 & 0.080427 \\
$x_{10}$ & 0.662617 & 0.612456 & 0.601526 & 0.610284 & 0.600000 & 0.689805 & 0.696629 & 0.632523 \\
$f$   & 85549.238 & 28943.61252 & 85548.120 & 70906.392 & 84084.876 & 84335.923 & 85549.160 & 85534.694 \\ \bottomrule
\end{tabular*}
\end{table*}

\cref{tab:rolling} presents the optimization results for the rolling element bearing problem achieved by BWE and competitive algorithms. The results demonstrate that BWE and COA performed exceptionally well, ranking first and second respectively. Meanwhile, although LEA and GWO did not surpass BWE, they still yielded satisfactory optimization outcomes. The optimization performance of GA and HEOA was less than ideal.

\section{Conclusion and future works}
\label{sec:Conclusion}
In this paper, a novel evolutionary algorithm named Bézier Walk Evolution (BWE) is proposed for global optimization.
BWE draws inspiration from computational geometry and stochastic processes, establishing an intrinsic analogy between evolutionary search and dynamic curve fitting.
BWE introduces three key contributions: (1) a hierarchical trajectory generation strategy based on variable-order Bézier curves that naturally decouples exploration and exploitation; (2) a distance-aware random walk mechanism leveraging population topology to guide control point selection, ensuring informative rather than blind search trajectories; and (3) an approximate tortuosity perturbation scheme designed to prevent stagnation in local optima.

Extensive experiments on the CEC2017 and CEC2022 benchmark suites across dimensions ranging from $10D$ to $100D$ demonstrate BWE's remarkable scalability and robustness.
For high-dimensional multimodal problems, BWE achieves superior convergence accuracy compared to classical algorithms (e.g., GA, SCA) and exhibits highly competitive performance against state-of-the-art evolutionary strategies (e.g., LSHADE-SPACMA, CMA-ES).
Exploration-exploitation analysis and trajectory visualization confirm its ability to preserve population diversity in early stages while ensuring rapid convergence in later stages.
Furthermore, BWE was successfully applied to five constrained engineering design problems, including the three-bar truss and rolling element bearing designs, consistently locating feasible solutions with minimal objective values.

Despite these promising results, BWE still exhibits certain limitations compared to mature SOTA algorithms, particularly regarding its ability to robustly escape specific deceptive local optima.
Future research will focus on the following directions:

\begin{itemize}[align=left, leftmargin=*]
    \item \textbf{Algorithmic Extensions:} Developing multi-objective, discrete, and binary variants of BWE for combinatorial optimization problems.
    
    \item \textbf{Mechanism Enhancement:} Improving performance through better population initialization, alternative trajectory search strategies, and more effective guidance mechanisms.
    
    \item \textbf{Hybridization and Complementarity:} Hybridizing BWE with other metaheuristics or local search operators to address weaknesses in specific landscapes.
\end{itemize}

\printcredits
\section*{Declaration of competing interest}
The authors declare that they have no known competing financial interests or personal relationships that could have appeared to influence the work reported in this paper.

\section*{Data availability}
No data was used for the research described in the article.

\bibliographystyle{cas-model2-names}

\bibliography{cas-refs}

\end{document}